\documentclass[jair,twoside,11pt,theapa]{article}
\makeatletter

\usepackage{jair}
\input epsf 
\usepackage[]{fontenc}
\usepackage[latin9]{inputenc}
\usepackage{verbatim}
\usepackage{float}
\usepackage{graphicx}
\usepackage{amssymb}
\usepackage{theapa}

\providecommand{\citep}{\cite}
\providecommand{\citealp}{\citeR}
\providecommand{\citet}{\citeA}

\usepackage{datetime}

\jairheading{32}{2008}{289--353}{09/07}{05/08}
\ShortHeadings{Optimal and Approximate Q-Value Functions for Dec-POMDPs}
{Oliehoek, Spaan \& Vlassis}
\firstpageno{289}

\usepackage{amsmath}
\usepackage{amssymb}
\usepackage{amsthm}

\usepackage[hang,tabbotcap]{subfigure}

\usepackage{colortbl}
\usepackage{multirow}
\usepackage{hhline}
\usepackage{algorithmic}
\usepackage{psfrag}
\usepackage{longtable}
\usepackage{verbatim}
\usepackage{icomma} 

\numberwithin{equation}{section}

\theoremstyle{plain}
\newtheorem{theorem}{Theorem}[section]
\newtheorem{lemma}{Lemma}[section]

\newtheorem{proposition}{Proposition}[section]

\theoremstyle{definition}
\newtheorem{definition}{Definition}[section]

\theoremstyle{plain}

\newtheorem*{propositionRep}{Proposition}

\newcommand{\proofup}{\begin{proof}\upshape}

\providecommand{\fig}{Figure~}
\providecommand{\figs}{Figures~}
\providecommand{\tab}{Table~}

\providecommand{\argmax}{\operatornamewithlimits{arg\,max}}
\providecommand{\defas}{\ \operatorname{\equiv}\ }
\providecommand{\inprod}[2]{{#1} \cdot {#2}}

\providecommand{\assign}{\leftarrow}

\providecommand{\real}{\mathbb {R} }

\providecommand{\Po}{\Pr_{o}}

\providecommand{\QF}{F} 
\providecommand{\kQF}{F_{k}} 
\providecommand{\kQFk}[1]{F_{#1}} 

\providecommand{\RF}{g} 

\providecommand{\QK}{K} 
\providecommand{\kQK}{K_k} 
\providecommand{\kQKk}[1]{K_{#1}} 

\providecommand{\kQ}{Q_k} 
\providecommand{\kQk}[1]{Q_{#1}} 

\providecommand{\jq}{q} 
\providecommand{\jqT}[1]{\jq^{#1}}
\providecommand{\jqTTG}[1]{\jq^{\ttg=#1}}
\providecommand{\qATTG}[2]{\jq_{#1}^{\ttg=#2}}
\providecommand{\qA}[1]{\jq_{#1}}
\providecommand{\qAT}[2]{\jq_{#1}^{#2}}
\providecommand{\jn}{\jpolBG} 

\providecommand{\jnT}[1]{\jn^{#1}}
\providecommand{\nA}[1]{\jn_{#1}}
\providecommand{\nAT}[2]{\jn_{#1}^{#2}}

\providecommand{\cat}[2]{ \langle {#1} \circ {#2} \rangle}

\providecommand{\ats}{\hat{\ts}}

\providecommand{\aEmpty}{a_{\emptyset}}

\renewcommand{\Pr}{P}
\providecommand{\nrA}{n} 
\providecommand{\nrS}{|\mathcal{S}|} 
\providecommand{\eps}{\epsilon} 

\providecommand{\PrS}{\mathcal{P}} 

\providecommand{\ttg}{\tau}         
\providecommand{\ts}{t}             

\providecommand{\Asingle}{}

\providecommand{\aone}{a} 	
\providecommand{\atwo}{\bar{a}} 	
\providecommand{\oone}{o} 	
\providecommand{\otwo}{\bar{o}} 	

\providecommand{\aoo}{a_1} 	
\providecommand{\aot}{\bar{a}_1} 	
\providecommand{\ato}{a_2} 	
\providecommand{\att}{\bar{a}_2} 	
\providecommand{\ooo}{o_1} 	
\providecommand{\oot}{\bar{o}_1} 	
\providecommand{\oto}{o_2} 	
\providecommand{\ott}{\bar{o}_2} 	

\providecommand{\aA}[1]{a_{#1}} 	
\providecommand{\aAS}[1]{\mathcal{A}_{#1}} 	
\providecommand{\aAT}[2]{\aA{#1}^{#2}} 	
\providecommand{\aAI}[2]{a_{{#1}{#2}}} 	
\providecommand{\aAIT}[3]{\aAI{#1}{#2}^{#3}} 	
\providecommand{\oA}[1]{o_{#1}} 	
\providecommand{\oAS}[1]{\mathcal{O}_{#1}} 	
\providecommand{\oAT}[2]{\oA{#1}^{#2}} 	
\providecommand{\oAI}[2]{o_{{#1}{#2}}} 	
\providecommand{\oAIT}[3]{\oAI{#1}{#2}^{#3}} 	

\providecommand{\jo}{o}    
\providecommand{\joS}{\mathcal{O}}    
\providecommand{\joT}[1]{\jo^{#1}}  
\providecommand{\ja}{a}    
\providecommand{\jaS}{\mathcal{A}}    
\providecommand{\jaT}[1]{\ja^{#1}}  
\providecommand{\jaIT}[2]{\ja^{#2}_{#1}}  

\providecommand{\comsymb}{\Sigma} 
\providecommand{\domsymb}{d} 

\providecommand{\msgAB}{\Sigma}		
\providecommand{\msgCost}{C_{\msgAB}}	
\providecommand{\dPol}[1]{\pol{#1}^{\domsymb}}
\providecommand{\cPol}[1]{\pol{#1}^{\comsymb}}
\providecommand{\dRew}{R^{\domsymb}}
\providecommand{\cRew}{R^{\comsymb}}
    
\providecommand{\jca}{\ja^{\comsymb}} 
\providecommand{\caA}[1]{\aA{#1}^{\comsymb}} 	
\providecommand{\caAS}[1]{\aAS{#1}^{\comsymb}} 	
\providecommand{\caAT}[2]{\aA{#1}^{\comsymb,#2}} 	
\providecommand{\caAI}[2]{\caA{#1,#2}}	 	
\providecommand{\caAIT}[3]{\aA{#1,#2}^{\comsymb,#3}}
\providecommand{\jda}{\ja^{d}} 
\providecommand{\daA}[1]{\aA{#1}^{\domsymb}} 	
\providecommand{\daAS}[1]{\aAS{#1}^{\domsymb}}	
\providecommand{\daAT}[2]{\aA{#1}^{\domsymb,#2}} 	
\providecommand{\daAI}[2]{\caA{#1,#2}}	 	
\providecommand{\daAIT}[3]{\aA{#1,#2}^{\domsymb,#3}}
\providecommand{\jco}{\jo^{\comsymb}} 
\providecommand{\coA}[1]{\oA{#1}^{\comsymb}} 	
\providecommand{\coAS}[1]{\oAS{#1}^{\comsymb}} 	
\providecommand{\coAT}[2]{\oA{#1}^{\comsymb,#2}} 	
\providecommand{\coAI}[2]{\coA{#1,#2}}	 	
\providecommand{\coAIT}[3]{\oA{#1,#2}^{\comsymb,#3}}
\providecommand{\jdo}{\jo^{\domsymb}} 
\providecommand{\doA}[1]{\oA{#1}^{\domsymb}} 	
\providecommand{\doAS}[1]{\oAS{#1}^{\domsymb}}	
\providecommand{\doAT}[2]{\oA{#1}^{\domsymb,#2}} 	
\providecommand{\doAI}[2]{\coA{#1,#2}}	 	
\providecommand{\doAIT}[3]{\oA{#1,#2}^{\domsymb,#3}}

\providecommand{\jc}{\sigma} 
\providecommand{\cA}[1]{\sigma_{#1}} 	
\providecommand{\cAT}[2]{\cA{#1}^{#2}} 	
\providecommand{\cAI}[2]{\sigma_{{#1}{#2}}} 	
\providecommand{\cAIT}[3]{\cAI{#1}{#2}^{#3}} 	
\providecommand{\cHist}{\vec{\jc}}
\providecommand{\cHistA}[1]{\vec{\sigma}_{#1}}

\providecommand{\cNull}{\sigma_{\emptyset}}  

\providecommand{\oaHist}{\vec{\btheta}}
\providecommand{\oaHistS}{\vec{\bTheta}}
\providecommand{\oaHistT}[1]{\vec{\btheta}{}^{\,#1}}
\providecommand{\oaHistTS}[1]{\vec{\bTheta}{}^{#1}}
\providecommand{\oaHistA}[1]{\vec{\theta}_{#1}}
\providecommand{\oaHistAS}[1]{\vec{\Theta}_{#1}}
\providecommand{\oaHistAT}[2]{\vec{\theta}_{#1}^{\,#2}}
\providecommand{\oaHistATS}[2]{\vec{\Theta}_{#1}^{#2}}
\providecommand{\oHist}{\vec{ \jo } }
\providecommand{\oHistS}{\vec{ \joS } }
\providecommand{\oHistT}[1]{  \vec{\jo}^{\,#1} } 
\providecommand{\oHistTS}[1]{  \oHistS^{#1}  }
\providecommand{\oHistA}[1]{\vec{o}_{#1}}
\providecommand{\oHistAS}[1]{\vec{\mathcal{O}}_{#1}}
\providecommand{\oHistAT}[2]{\vec{o}_{#1}^{\,#2}}
\providecommand{\oHistATS}[2]{\vec{\mathcal{O}}_{#1}^{#2}}

\providecommand{\aHist}{\vec{ \ja }}
\providecommand{\aHistS}{\vec{ \jaS} }
\providecommand{\aHistT}[1]{\vec{ \ja}^{\,#1}}
\providecommand{\aHistTS}[1]{ \aHistS^{#1}}
\providecommand{\aHistA}[1]{\vec{a}_{#1}}
\providecommand{\aHistAS}[1]{\vec{\mathcal{A}}_{#1}}
\providecommand{\aHistAT}[2]{\vec{a}_{#1}^{\,#2}}
\providecommand{\aHistATS}[2]{\vec{\mathcal{A}}_{#1}^{#2}}

\providecommand{\oaHistTP}[1]{\vec{\btheta'}^{\,#1}} 
\providecommand{\oaHistATP}[2]{\vec{\theta'}_{#1}^{\,#2}}  

\providecommand{\CoaHistTS}[2]{\oaHistTS{#1}_{#2}}

\providecommand{\sHistT}[1]{\vec{s}^{\,#1}} 

\providecommand{\oNullA}[1]{o_{#1,\emptyset}}
\providecommand{\joNull}{\jo_{\emptyset}}
\providecommand{\oHistEmpty}{\oHist_\emptyset}
\providecommand{\oaHistEmpty}{\oaHist_\emptyset}
\def\bpi{\pi} 
\def\bPi{\Pi} 
\def\btheta{\theta}
\def\bTheta{{\Theta}}

\providecommand{\jpol}{\bpi}	
\providecommand{\jpolT}[1]{\bpi^{#1}}	
\providecommand{\jpolS}{\bPi}	
\providecommand{\jppol}{\bpi} 	
\providecommand{\jmpol}{\mu}	
\providecommand{\jspol}{\varsigma} 	
\providecommand{\polA}[1]{\pi_{#1}}	
\providecommand{\polAS}[1]{\Pi_{#1}}	
\providecommand{\pol}[1]{\polA{#1}}	
\providecommand{\polS}[1]{\polAS{#1}}	
\providecommand{\ppol}[1]{\pi_{#1}} 	
\providecommand{\mpol}[1]{\mu_{#1}} 	
\providecommand{\mjpol}{\mu} 	
\providecommand{\spol}[1]{\varsigma_{#1}} 	

\providecommand{\polAI}[2]{\jpol_{#1,#2}} 

\providecommand{\drA}[1]{\delta_{#1}}
\providecommand{\drAT}[2]{\delta_{#1}^{#2}}
\providecommand{\jdrT}[1]{\delta^{#1}}

\providecommand{\partJPol}{\varphi} 
\providecommand{\partJPolS}{\Phi} 
\providecommand{\partJPolT}[1]{\partJPol^{#1}} 
\providecommand{\partPolAT}[2]{\partJPol^{#2}_{#1}}

\providecommand{\pF}{\psi}	
\providecommand{\pFT}[1]{\psi^{#1}}	

\providecommand{\pP}{\partJPol}	
\providecommand{\pJPol}{\pP}	
\providecommand{\pPolA}[1]{\pP_{#1}}
\providecommand{\pJPolT}[1]{\pP^{#1}}
\providecommand{\pPolAT}[2]{\pPolA{#1}^{#2}}	

\providecommand{\stSymb}{q}
\providecommand{\stSymbS}{\mathcal{Q}}
\providecommand{\stFunc}{Q} 
\providecommand{\stFuncJPol}{Q_\jpol} 
\providecommand{\stJPol}{\stSymb} 
\providecommand{\stJPolT}[1]{\stSymb^{\ttg=#1}} 
\providecommand{\stPolAT}[2]{\stSymb_{#1}^{\ttg=#2}}
\providecommand{\stJPolTS}[1]{\stSymbS^{\ttg=#1}} 
\providecommand{\stPolATS}[2]{\stSymbS_{#1}^{\ttg=#2}}

\providecommand{\jpolttg}{\jpol^{\ttg}}	
\providecommand{\jpolttgT}[1]{\jpol^{\ttg=#1}}	

\providecommand{\V}{V}	

\providecommand{\Vjpol}[1]{\V_{#1}}  
\providecommand{\VjpolT}[2]{\V_{#1}^{#2}}  

\providecommand{\Vttg}{\V^{\ttg}}	
\providecommand{\VttgT}[1]{\V^{\ttg=#1}}	

\providecommand{\QH}{\widehat{Q}}
\providecommand{\VH}{\widehat{V}}
\providecommand{\VHttg}{\VH^{\ttg}}
\providecommand{\VHttgT}[1]{\VH^{\ttg=#1}}	

\providecommand{\vVec}{v}   
\providecommand{\vVecS}{\mathcal{V}}   
\providecommand{\vVecT}[1]{v^{#1}}   
\providecommand{\vVecTS}[1]{\mathcal{V}^{#1}}   

\providecommand{\avVec}[1]{v_{#1}}   
\providecommand{\avVecS}[1]{\mathcal{V_{#1}}}   
\providecommand{\avVecT}[2]{v_{#1}^{#2}}   
\providecommand{\avVecTS}[2]{\mathcal{V}_{#1}^{#2}}   

\providecommand{\vecPOMDP}{\vVec}
\providecommand{\vecS}{\mathcal{V}}
\providecommand{\vecA}[1]{\vecPOMDP_{#1}}
\providecommand{\vecAI}[2]{\vecPOMDP_{#1,#2}}
\providecommand{\vecAS}[1]{\vecS_{#1}}

\providecommand{\vecIT}[2]{\vecPOMDP_{#1}^{#2}} 
\providecommand{\vecI}[1]{\vecPOMDP_{#1}} 

\providecommand{\jtype}{\theta}
\providecommand{\jtypeS}{\Theta}
\providecommand{\typeA}[1]{\jtype_{#1}}
\providecommand{\typeAS}[1]{\jtypeS_{#1}}
\providecommand{\utF}{u}		
\providecommand{\utA}[1]{\utF_{#1}}	
\providecommand{\polBG}[1]{\jpolBG_{#1}}
\providecommand{\jpolBG}{\beta} 
\providecommand{\sgPolA}[1]{\alpha_{#1}} 
\providecommand{\sgJPol}{\alpha}

\providecommand{\FSCndS}{\ndS}
\providecommand{\FSCnd}{\nd}
\providecommand{\FSCndStart}{\FSCnd^{0}}
\providecommand{\FSCas}{\alpha}	
\providecommand{\FSCos}{\eta}	
\providecommand{\FSCndAS}[1]{\FSCndS_{#1}}
\providecommand{\FSCjnd}{\mathbf{\FSCnd}}
\providecommand{\FSCndA}[1]{\FSCnd_{#1}}
\providecommand{\FSCndStartA}[1]{\FSCndA{#1}^{0}}
\providecommand{\FSCjndStart}{\mathbf{\FSCnd}^{0}}
\providecommand{\FSCasA}[1]{\FSCas_{#1}}	
\providecommand{\FSCjas}{\mathbf{\FSCas}}	
\providecommand{\FSCosA}[1]{\FSCos_{#1}}	
\providecommand{\FSCjos}{\mathbf{\FSCos}}	
\providecommand{\FSCjosjnd}{\FSCjos^\FSCjnd}	

\providecommand{\nd}{n} 		
\providecommand{\ndO}{\nd^O} 		
\providecommand{\ndS}{\mathcal{N}} 	
\providecommand{\nroot}{\nd_{root}}	
\providecommand{\path}{\sigma} 		
\providecommand{\natPcomp}{\nu} 	
\providecommand{\natPCP}{\nu'} 		
\providecommand{\seqA}[1]{\sigma_{#1}} 	
\providecommand{\seqRoot}{\sigma_{root}}
\providecommand{\rw}{\rho} 		
\providecommand{\rwA}[1]{\rw_{#1}}	
\providecommand{\rwAP}[2]{\rw_{#1}^{#2}}	
\providecommand{\rwNO}{\rw_{NO}}	
\providecommand{\IS}{I}			
\providecommand{\ISA}[1]{\IS_{#1}}	
\providecommand{\ISAT}[2]{\IS_{#1}^{#2}}
\providecommand{\ISS}{\mathcal{I}}		
\providecommand{\ISAS}[1]{\ISS_{#1}}	
\providecommand{\ISATS}[2]{\ISS_{#1}^{#2}}
\providecommand{\ISF}{K}		
\providecommand{\POM}{\mathbf{R}}	
\providecommand{\POMA}[1]{\mathbf{R}_{#1}}
\providecommand{\POME}{r}		
\providecommand{\POMEA}[1]{r_{#1}}	
\providecommand{\POMone}{\mathbf{R}}	
\providecommand{\POMEone}{r}		

\providecommand{\QBG}{\textrm{Q}_{\textrm{BG}}} 
\providecommand{\kQBG}{k\textrm{-Q}_{\textrm{BG}}} 
\providecommand{\kQBGQ}{k\textrm{-}Q_{\textrm{B}}} 
\providecommand{\kQBGv}[1]{k\textrm{-Q}_{{#1}\textrm{-BG}}} 
\providecommand{\kQBGQv}[1]{k\textrm{-}Q_{{#1}\textrm{-B}}} %
\providecommand{\kQBGk}[1]{{#1}\textrm{-Q}_{\textrm{BG}}} 
\providecommand{\kQBGQk}[1]{{#1}\textrm{-}Q_{\textrm{B}}} 
\providecommand{\QMDP}{\textrm{Q}_{\textrm{MDP}}} 
\providecommand{\QPOMDP}{\textrm{Q}_{\textrm{POMDP}}} 
\providecommand{\QBGQ}{{Q}_{\textrm{B}}} 
\providecommand{\QMDPQ}{{Q}_{\textrm{M}}} 
\providecommand{\QPOMDPQ}{{Q}_{\textrm{P}}} 
\providecommand{\QPOMDPV}{{V}_{\textrm{P}}} 

\providecommand{\jpolMDP}{\jpol_{\textrm{M}}}
\providecommand{\jpolPOMDP}{\jpol_{\textrm{P}}}

\providecommand{\QBGQH}{{\QH}_{\textrm{B}}} 
\providecommand{\QMDPQH}{{\QH}_{\textrm{M}}} 
\providecommand{\QPOMDPQH}{{\QH}_{\textrm{P}}} 

\providecommand{\MAA}{{\textrm{MAA}^{*}}} 
\providecommand{\GMAA}{{\textrm{GMAA}^{*}}} 
\providecommand{\kGMAA}{{k\textrm{-GMAA}^{*}}} 
\providecommand{\PSO}{\texttt{Next}} 
\providecommand{\PSOset}{\partJPolS_{\PSO}} 
\providecommand{\maxlb}{{\underbar{v}}^{\star}}
\providecommand{\maxjpol}{{\jpol}^{\star}}
\providecommand{\selectO}{\texttt{Select}}
\providecommand{\polPool}{\textrm{P}}

\providecommand{\modA}[1]{m_{#1}}	
\providecommand{\modAS}[1]{\modS_{#1}}	
\providecommand{\modS}{M}	
\providecommand{\IPhistA}[1]{\oHistA{#1}} 
\providecommand{\IPfuncA}[1]{\pol{#1}} 
\providecommand{\IPenvA}[1]{O_{#1}} 
\providecommand{\IPtypeA}[1]{\typeA{#1}}
\providecommand{\IPframeA}[1]{\hat{\IPtypeA{#1}}}

\providecommand{\sP}{p}	
\providecommand{\lsA}[1]{\hat{s}_{#1}}  
\providecommand{\factor}[1]{\mathcal{F}_{#1}} 

\providecommand{\Aug}[1]{\bar{#1}}
\providecommand{\sAug}{\Aug{s}}

\providecommand{\sL}{s_l}
\providecommand{\sR}{s_r}

\providecommand{\aOL}{a_{\textrm{\tiny OL}}}
\providecommand{\aOR}{a_{\textrm{\tiny OR}}}
\providecommand{\aLi}{a_{\textrm{\tiny Li}}}

\providecommand{\oHL}{o_{\textrm{\tiny HL}}}
\providecommand{\oHR}{o_{\textrm{\tiny HR}}}

\newlength{\thickertableline}
\definecolor{tabledark}{rgb}{0.4, 0.4, 0.60}
\definecolor{tablemid}{rgb}{0.80, 0.80, 0.95}
\definecolor{tablelight}{rgb}{0.90, 0.90, 1.0}

\definecolor{examtablelight}{rgb}{0.95,1.0,1.0}
\definecolor{examtabledark}{rgb}{0.2,0.4,0.2}

\def\<#1>{%
    \expandafter\ifx\csname<#1>\endcsname\relax
        \errmessage{abbreviation <#1> undefined!}
    \else
        \csname<#1>\endcsname
    \fi
}
\def\abbr#1#2{%
    \expandafter\def\csname<#1>\endcsname{#2}%
}
\abbr{dectig}{Dec-Tiger}
\abbr{sdectig}{Skewed Dec-Tiger}

\providecommand{\tabularnewline}{\\}
\floatstyle{ruled}
\newfloat{algorithm}{tbp}{loa}
\floatname{algorithm}{Algorithm}

\makeatother

\begin{document}
\title{Optimal and Approximate Q-value Functions \\ for Decentralized POMDPs}

\author{%
\name Frans A. Oliehoek \email f.a.oliehoek@uva.nl \\
\addr Intelligent Systems Lab Amsterdam, University of Amsterdam\\
Amsterdam, The Netherlands 
\AND 
\name Matthijs T.J. Spaan \email mtjspaan@isr.ist.utl.pt \\
\addr Institute for Systems and Robotics, Instituto Superior T{\'e}cnico\\
Lisbon, Portugal
\AND
\name Nikos Vlassis \email vlassis@dpem.tuc.gr \\
\addr Department of Production Engineering and Management, Technical University of Crete\\
Chania, Greece
}

\maketitle

\begin{abstract}
Decision-theoretic planning is a popular approach to sequential decision
making problems, because it treats uncertainty in sensing and acting
in a principled way. In single-agent frameworks like MDPs and POMDPs,
planning can be carried out by resorting to Q-value functions: an
optimal Q-value function $Q^{*}$ is computed in a recursive manner
by dynamic programming, and then an optimal policy is extracted from
$Q^{*}$. In this paper we study whether similar Q-value functions
can be defined for decentralized POMDP models (Dec-POMDPs), and how
policies can be extracted from such value functions. We define two
forms of the optimal Q-value function for Dec-POMDPs: one that gives
a normative description as the Q-value function of an optimal pure
joint policy and another one that is sequentially rational and thus
gives a recipe for computation. This computation, however, is infeasible
for all but the smallest problems. Therefore, we analyze various approximate
Q-value functions that allow for efficient computation. We describe
how they relate, and we prove that they all provide an upper bound
to the optimal Q-value function $Q^{*}$. Finally, unifying some previous
approaches for solving Dec-POMDPs, we describe a family of algorithms
for extracting policies from such Q-value functions, and perform an
experimental evaluation on existing test problems, including a new
firefighting benchmark problem.
\end{abstract}

\section{Introduction}

\label{sec:Introduction}

One of the main goals in artificial intelligence (AI) is the development
of intelligent agents, which perceive their environment through sensors
and influence the environment through their actuators. In this setting,
an essential problem is how an agent should decide which action to
perform in a certain situation. In this work, we focus on \emph{planning}:
constructing a plan that specifies which action to take in each situation
the agent might encounter over time. In particular, we will focus
on planning in a cooperative multiagent system (MAS): an environment
in which multiple agents coexist and interact in order to perform
a joint task. We will adopt a decision-theoretic approach, which allows
us to tackle uncertainty in sensing and acting in a principled way. 

Decision-theoretic planning has roots in control theory and in operations
research. In control theory, one or more controllers control a stochastic
system with a specific output as goal. Operations research considers
tasks related to scheduling, logistics and work flow and tries to
optimize the concerning systems. Decision-theoretic planning problems
can be formalized as \emph{Markov decision processes} (MDPs), which
have have been frequently employed in both control theory as well
as operations research, but also have been adopted by AI for planning
in stochastic environments. In all these fields the goal is to find
a (conditional) plan, or \emph{policy}, that is optimal with respect
to the desired behavior. Traditionally, the main focus has been on
systems with only one agent or controller, but in the last decade
interest in systems with multiple agents or decentralized control
has grown.

A different, but also related field is that of game theory. Game theory
considers agents, called \emph{players}, interacting in a dynamic,
potentially stochastic process, the game. The goal here is to find
optimal \emph{strategies} for the agents, that specify how they should
play and therefore correspond to policies. In contrast to decision-theoretic
planning, game theory has always considered multiple agents, and as
a consequence several ideas and concepts from game theory are now
being applied in decentralized decision-theoretic planning. In this
work we apply game-theoretic models to decision-theoretic planning
for multiple agents.

\subsection{Decision-Theoretic Planning}

In the last decades, the \emph{Markov decision process (MDP)} framework
has gained in popularity in the AI community as a model for planning
under uncertainty \citep{Boutilier99decisiontheoretic,guestrin03JAIR}.
MDPs can be used to formalize a discrete time planning task of a single
agent in a stochastically changing environment, on the condition that
the agent can observe the state of the environment. Every time step
the state changes stochastically, but the agent chooses an action
that selects a particular transition function. Taking an action from
a particular state at time step $\ts$ induces a probability distribution
over states at time step $\ts+1$. 

The agent's objective can be formulated in several ways. The first
type of objective of an agent is reaching a specific goal state, for
example in a maze in which the agent's goal is to reach the exit.
A different formulation is given by associating a certain cost with
the execution of a particular action in a particular state, in which
case the goal will be to minimize the expected total cost. Alternatively,
one can associate rewards with actions performed in a certain state,
the goal being to maximize the total reward.

When the agent knows the probabilities of the state transitions, i.e.,
when it knows the model, it can contemplate the expected transitions
over time and construct a plan that is most likely to reach a specific
goal state, minimizes the expected costs or maximizes the expected
reward. This stands in some contrast to reinforcement learning (RL)
\citep{SuttonBarto98}, where the agent does not have a model of the
environment, but has to learn good behavior by repeatedly interacting
with the environment. Reinforcement learning can be seen as the combined
task of learning the model of the environment \emph{and} planning,
although in practice often it is not necessary to explicitly recover
the environment model. In this article we focus only on planning,
but consider two factors that complicate computing successful plans:
the inability of the agent to observe the state of the environment
as well as the presence of multiple agents.

In the real world an agent might not be able to determine what the
state of the environment exactly is, because the agent's sensors are
noisy and/or limited. When sensors are noisy, an agent can receive
faulty or inaccurate observations with some probability. When sensors
are limited the agent is unable to observe the differences between
states that cannot be detected by the sensor, e.g., the presence or
absence of an object outside a laser range-finder's field of view.
When the same sensor reading might require different action choices,
this phenomenon is referred to as \emph{perceptual aliasing\@.} In
order to deal with the introduced sensor uncertainty, a \emph{partially
observable Markov decision process (POMDP)} extends the MDP model
by incorporating observations and their probability of occurrence
conditional on the state of the environment \citep{Kaelbling98AI}.

The other complicating factor we consider is the presence of multiple
agents. Instead of planning for a single agent we now plan for a team
of cooperative agents. We assume that communication within the team
is not possible.%
\footnote{As it turns out, the framework we consider can also model communication
with a particular cost that is subject to minimization \citep{Pynadath02_com_MTDP,GoldmanZ04Categorization}.
The non-communicative setting can be interpreted as the special case
with infinite cost.%
} A major problem in this setting is how the agents will have to coordinate
their actions. Especially, as the agents are not assumed to observe
the state---each agent only knows its own observations received and
actions taken---there is no common signal they can condition their
actions on. Note that this problem is in addition to the problem of
partial observability, and not a substitution of it; even if the agents
could freely and instantaneously communicate their individual observations,
the joint observations would not disambiguate the true state.

One option is to consider each agent separately, and have each such
agent maintain an explicit model of the other agents. This is the
approach as chosen in the Interactive POMDP (I-POMDP) framework \citep{Gmytrasiewicz05}.
A problem in this approach, however, is that the other agents also
model the considered agent, leading to an infinite recursion of beliefs
regarding the behavior of agents. We will adopt the \emph{decentralized
partially observable Markov decision process (Dec-POMDP)} model for
this class of problems \citep{Bernstein02Complexity}. A Dec-POMDP
is a generalization to multiple agents of a POMDP and can be used
to model a team of cooperative agents that are situated in a stochastic,
partially observable environment.

The single-agent MDP setting has received much attention, and many
results are known. In particular it is known that an optimal plan,
or policy, can be extracted from the optimal action-value, or \emph{Q-value,}
function $Q^{*}(s,a)$, and that the latter can be calculated efficiently.
For POMDPs, similar results are available, although finding an optimal
solution is harder (PSPACE-complete for finite-horizon problems, \citealp{Papadimitriou87}).

On the other hand, for Dec-POMDPs relatively little is known except
that they are provably intractable (NEXP-complete, \citealp{Bernstein02Complexity}).
In particular, an outstanding issue is whether Q-value functions can
be defined for Dec-POMDPs just as in (PO)MDPs, and whether policies
can be extracted from such Q-value functions. Currently most algorithms
for planning in Dec-POMDPs are based on some version of policy search
\citep{Nair03_JESP,Hansen04,Szer05MAA,Varakantham07}, and a proper
theory for Q-value functions in Dec-POMDPs is still lacking. Given
the wide range of applications of value functions in single-agent
decision-theoretic planning, we expect that such a theory for Dec-POMDPs
can have great benefits, both in terms of providing insight as well
as guiding the design of solution algorithms.

\subsection{Contributions}

In this paper we develop theory for Q-value functions in Dec-POMDPs,
showing that an optimal Q-function $Q^{*}$ \emph{can} be defined
for a Dec-POMDP. We define two forms of the optimal Q-value function
for Dec-POMDPs: one that gives a normative description as the Q-value
function of an optimal  pure joint policy and another one that is
sequentially rational and thus gives a recipe for computation. We
also show that given $Q^{*}$, an optimal policy can be computed by
\emph{forward-sweep policy computation,} solving a sequence of Bayesian
games forward through time (i.e., from the first to the last time
step), thereby extending the solution technique of \citet{Emery-Montemerlo04}
to the exact setting.

Computation of $Q^{*}$ is infeasible for all but the smallest problems.
Therefore, we analyze three different approximate Q-value functions
$\QMDP,$ $\QPOMDP$ and $\QBG$ that can be more efficiently computed
and which constitute upper bounds to $Q^{*}$. We also describe a
generalized form of $\QBG$ that includes $\QPOMDP$, $\QBG$ and
$Q^{*}$. This is used to prove a hierarchy of upper bounds: $Q^{*}\leq\QBG\leq\QPOMDP\leq\QMDP$.

Next, we show how these approximate Q-value functions can be used
to compute optimal or sub-optimal policies. We describe a generic
policy search algorithm, which we dub Generalized $\MAA$ ($\GMAA$)
as it is a generalization of the $\MAA$ algorithm by \citet{Szer05MAA},
that can be used for extracting a policy from an approximate Q-value
function. By varying the implementation of a sub-routine of this algorithm,
this algorithm unifies $\MAA$ and forward-sweep policy computation
and thus the approach of \citet{Emery-Montemerlo04}. 

\sloppypar Finally, in an experimental evaluation we examine the
differences between $\QMDP$, $\QPOMDP$, $\QBG$ and $Q^{*}$ for
several problems. We also experimentally verify the potential benefit
of tighter heuristics, by testing different settings of $\GMAA$ on
some well known test problems and on a new benchmark problem involving
firefighting agents. 

This article is based on previous work by \citet{Oliehoek07aamas}---abbreviated
OV here---containing several new contributions: (1) Contrary to the
OV work, the current work includes a section on the sequential rational
description of $Q^{*}$ and suggests a way to compute $Q^{*}$ in
practice (OV only provided a normative description of $Q^{*}$). (2)
The current work provides a formal proof of the hierarchy of upper
bounds to $Q^{*}$ (which was only qualitatively argued in the OV
paper). (3) The current article additionally contains a proof that
the solutions for the Bayesian games with identical payoffs given
by equation \eqref{eq:BG_solution} constitute Pareto optimal Nash
equilibria of the game (which was not proven in the OV paper). (4)
This article contains a more extensive experimental evaluation of
the derived bounds of $Q^{*}$, and introduces a new benchmark problem
(firefighting). (5) Finally, the current article provides a more complete
introduction to Dec-POMDPs and existing solution methods, as well
as Bayesian games, hence it can serve as a self-contained introduction
to Dec-POMDPs.

\subsection{Applications}

Although the field of multiagent systems in a stochastic, partially
observable environment seems quite specialized and thus narrow, the
application area is actually very broad. The real world is practically
always partially observable due to sensor noise and perceptual aliasing.
Also, in most of these domains communication is not free, but consumes
resources and thus has a particular cost. Therefore models as Dec-POMDPs,
which do consider partially observable environments are relevant for
essentially all teams of embodied agents. 

Example applications of this type are given by \citet{Emery-Montemerlo05phd},
who considered multi-robot navigation in which a team of agents with
noisy sensors has to act to find/capture a goal. \citet{Becker04TransIndepJAIR}
use a multi-robot space exploration example. Here, the agents are
Mars rovers and have to decide on how to proceed their mission: whether
to collect particular samples at specific sites or not. The rewards
of particular samples can be sub- or super-additive, making this task
non-trivial.  An overview of application areas in cooperative robotics
is presented by \citet{Arai02}, among which is robotic soccer, as
applied in RoboCup \citep{Kitano97}. Another application that is
investigated within this project is crisis management: RoboCup Rescue
\citep{Kitano99RCR} models a situation where rescue teams have to
perform a search and rescue task in a crisis situation. This task
also has been modeled as a partially observable system \citep{Nair02TeamFormForReform,Nair02teamRoboCupRescue,nair03_RMTDP,Oliehoek06aamas,Paquet05_onlinePOMDP}.

There are also many other types of applications. \citet{Nair05ndPOMDPs,Lesser03distSensorNetsBook}
give applications for distributed sensor networks (typically used
for surveillance). An example of load balancing among queues is presented
by \citet{Cogill04approxDP}. Here agents represent queues and can
only observe queue sizes of themselves and immediate neighbors. They
have to decide whether to accept new jobs or pass them to another
queue. Another frequently considered application domain is communication
networks. \citet{Peshkin01PhD} treated a packet routing application
in which agents are routers and have to minimize the average transfer
time of packets. They are connected to immediate neighbors and have
to decide at each time step to which neighbor to send each packet.
Other approaches to communication networks using decentralized, stochastic,
partially observable systems are given by \citet{Ooi96,Tao01,Altman02}.

\subsection{Overview of Article}

The rest of this article is organized as follows. In Section~\ref{sec:Decentralized-POMDPs}
we will first formally introduce the Dec-POMDP model and provide background
on its components. Some existing solution methods are treated in Section~\ref{sec:Existing-solution-methods}.
Then, in Section~\ref{sec:Q-value-functions-forDec-POMDPs} we show
how a Dec-POMDP can be modeled as a series of Bayesian games and how
this constitutes a theory of Q-value functions for BGs. We also treat
two forms of optimal Q-value functions, $Q^{*}$, here. Approximate
Q-value functions are described in Section~\ref{sec:Approximate-Q-value-functions}
and one of their applications is discussed in Section~\ref{sec:generaliz-value-based-policy-search}.
Section~\ref{sec:Experiments} presents the results of the experimental
evaluation. Finally, Section~\ref{sec:Conclusion-and-discussion}
concludes.

\section{Decentralized POMDPs}

\label{sec:Decentralized-POMDPs}

In this section we define the Dec-POMDP model and discuss some of
its properties. Intuitively, a Dec-POMDP models a number of agents
that inhabit a particular environment, which is considered at discrete
\emph{time steps}, also referred to as \emph{stages} \citep{Boutilier99decisiontheoretic}
or \emph{(decision) epochs} \citep{Puterman94}. The number of time
steps the agents will interact with their environment is called the
\emph{horizon} of the decision problem, and will be denoted by $h$.
In this paper the horizon is assumed to be finite. At each stage $\ts=0,1,2,\dots,h-1$
every agent takes an action and the combination of these actions influences
the environment, causing a state transition. At the next time step,
each agent first receives an observation of the environment, after
which it has to take an action again. The probabilities of state
transitions and observations are specified by the Dec-POMDP model,
as are rewards received for particular actions in particular states.
The transition- and observation probabilities specify the dynamics
of the environment, while the rewards specify what behavior is desirable.
Hence, the reward model defines the agents' goal or task: the agents
have to come up with a plan that maximizes the expected long term
reward signal.  In this work we  assume that planning takes place
off-line, after which the computed plans are distributed to the agents,
who then merely execute the plans on-line. That is, computation of
the plan is centralized, while execution is decentralized. In the
centralized planning phase, the entire model as detailed below is
available. During execution each agent only knows the joint policy
as found by the planning phase and its individual history of actions
and observations.

\subsection{Formal Model}

\label{sub:Formal-Model}

In this section we more formally treat the basic components of a Dec-POMDP.
We start by giving a mathematical definition of these components.

\begin{definition}

A \emph{decentralized partially observable Markov decision process
(Dec-POMDP)} is defined as a tuple $\left\langle \nrA,\mathcal{S},\mathcal{A},T,R,\mathcal{O},O,h,b^{0}\right\rangle $
where:

\begin{itemize}
\item \emph{$\nrA$} is the number of agents. 
\item $\mathcal{S}$ is a finite set of states.
\item $\mathcal{A}$ is the set of joint actions.
\item $T$ is the transition function.
\item $R$ is the immediate reward function.
\item $\mathcal{O}$ is the set of joint observations.
\item $O$ is the observation function.
\item $h$ is the horizon of the problem. 
\item $b^{0}\in\PrS(\mathcal{S})$, is the initial state distribution at
time $\ts=0$.%
\footnote{$\PrS(\cdot)$ denotes the set of probability distributions over $(\cdot)$.%
}
\end{itemize}
\end{definition} 

The Dec-POMDP model extends single-agent (PO)MDP models by considering
\emph{joint} actions and observations. In particular, we define $\mathcal{A}=\times_{i}\aAS i$
as the set of \emph{joint actions}. Here, $\aAS i$ is the set of
actions available to agent $i$. Every time step, one joint action
$\ja=\left\langle \aA1,...,\aA\nrA\right\rangle $ is taken. In a
Dec-POMDP, agents only know their own individual action; they do not
observe each other's actions. We will assume that any action $\aA i\in\aAS i$
can be selected at any time. So the set $\aAS i$ does not depend
on the stage or state of the environment. In general, we will denote
the stage using superscripts, so $\jaT{\ts}$ denotes the joint action
taken at stage~$\ts$, $\aAT i\ts$ is the individual action of agent
$i$ taken at stage $\ts$.  Also, we write $\aA{\neq i}=\left\langle \aA1,\dots,\aA{i-1},\aA{i+1},\dots,\aA{\nrA}\right\rangle $
for a profile of actions for all agents but $i$.

\begin{figure}[t]
\noindent \begin{centering}
{ \providecommand{\s}{\footnotesize}

\psfrag{decis}{\s actions}
\psfrag{obs}{\s observations}
\psfrag{states}{\s states}

\psfrag{s0}[br][bc]{\s $s^0$}
\psfrag{s1}[br][bc]{\s $s^1$}
\psfrag{sh-1}[br][bc]{\s $s^{h-1}$}

\psfrag{o0}[br][bc]{\s $\jo^0$}
\psfrag{o1}[br][bc]{\s $\jo^1$}
\psfrag{oh-1}[br][bc]{\s $\jo^{h-1}$}

\psfrag{o0_1}[bc][bc]{\s $\oAT{1}{0}$}
\psfrag{o0_n}[br][bc]{\s $\oAT{\nrA}{0}$}
\psfrag{o1_1}[bc][bc]{\s $\oAT{1}{1}$}
\psfrag{o1_n}[br][bc]{\s $\oAT{\nrA}{1}$}
\psfrag{oh-1_1}[bc][bc]{\s $\oAT{1}{h-1}$}
\psfrag{oh-1_n}[br][bc]{\s $\oAT{\nrA}{h-1}$}

\psfrag{a0}{\s $\ja^0$}
\psfrag{a1}{\s $\ja^1$}
\psfrag{ah-2}{\s $\ja^{h-2}$}
\psfrag{ah-1}{\s $\ja^{h-1}$}

\psfrag{vd}[Bl][Bc]{$\vdots$}
\psfrag{...}[Bl][Bl]{$\dots$}

\psfrag{a0_1}[Br][Bc]{\s $\aAT{1}{0}$}
\psfrag{a0_n}[Br][Bc]{\s $\aAT{\nrA}{0}$}
\psfrag{a1_1}[Br][Bc]{\s $\aAT{1}{1}$}
\psfrag{a1_n}[Br][Bc]{\s $\aAT{\nrA}{1}$}
\psfrag{ah-1_1}[Br][Bc]{\s $\aAT{1}{h-1}$}
\psfrag{ah-1_n}[Br][Bc]{\s $\aAT{\nrA}{h-1}$}

\psfrag{ah-2}{\s $\ja^{h-2}$}
\psfrag{ah-1}{\s $\ja^{h-1}$}

\psfrag{ts}{\s $\ts$}
\psfrag{t0}[br][bc]{\s $0$}
\psfrag{t1}[br][bc]{\s $1$}
\psfrag{th-1}[br][bc]{\s $h-1$}
\psfrag{ttg}{\s $\ttg$}
\psfrag{ttg0}[cl][cl]{\s $h$}
\psfrag{ttg1}[br][bc]{\s $h-1$}
\psfrag{ttgh-1}[cl][cl]{\s $1$}\includegraphics[width=0.95\columnwidth]{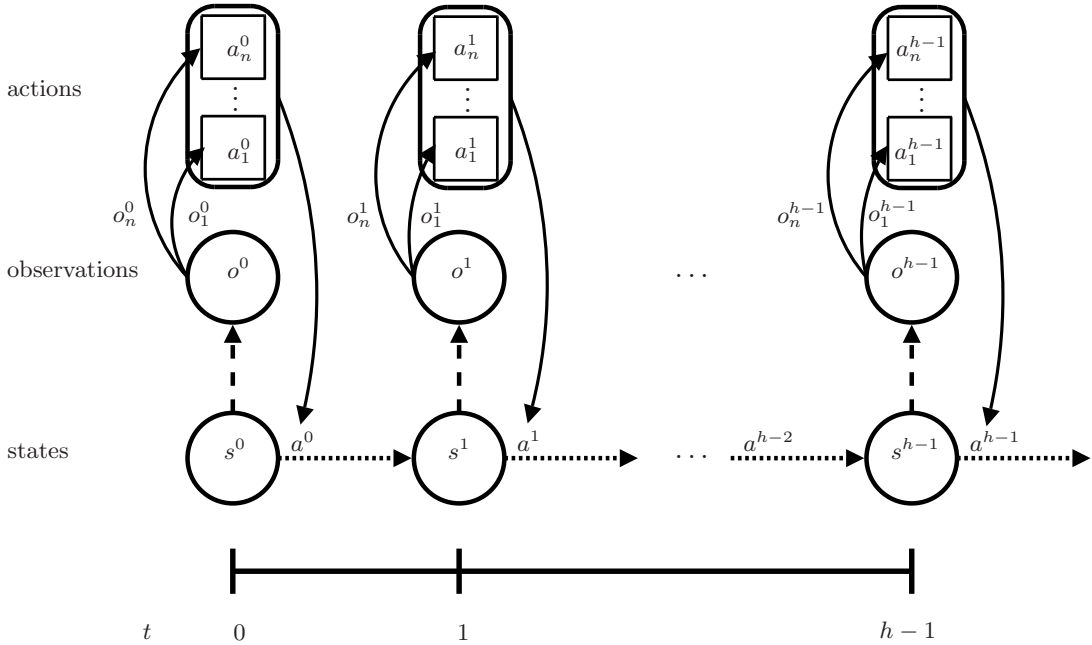}}
\par\end{centering}

\caption{An illustration of the dynamics of a Dec-POMDP. At every stage the
environment is in a particular state. This state emits a joint observation,
of which each agent observes its individual observation. Then each
agent selects an action forming the joint action.}
\label{fig:DecPOMDP-dynamics}
\end{figure}

Similarly to the set of joint actions, $\mathcal{O}=\times_{i}\mathcal{O}_{i}$
is the set of joint observations, where $\mathcal{O}_{i}$ is a set
of observations available to agent $i$. Every time step the environment
emits one joint observation $\jo=\left\langle \oA1,...,\oA\nrA\right\rangle $,
from which each agent $i$ only observes its own component $\oA i$,
as illustrated by \fig\ref{fig:DecPOMDP-dynamics}. Notation with
respect to time and indices for observations is analogous to the notation
for actions. In this paper, we will assume that the action- and observation
sets are finite. Infinite action- and observation sets are very difficult
to deal with even in the single-agent case, and to the authors' knowledge
no research has been performed on this topic in the partially observable,
multiagent case.

Actions and observations are the interface between the agents and
their environment. The Dec-POMDP framework describes this environment
by its \emph{states} and \emph{transitions.} This means that rather
than considering a complex, typically domain-dependent model of the
environment that explains how this environment works, a descriptive
stance is taken: A Dec-POMDP specifies an environment model simply
as the set of states $\mathcal{S}=\left\{ s_{1},...,s_{\nrS}\right\} $
the environment can be in, together with the probabilities of state
transitions that are dependent on executed joint actions. In particular,
the transition from some state to a next state depends stochastically
on the past states and actions. This probabilistic dependence models
\emph{outcome uncertainty}: the fact that the outcome of an action
cannot be predicted with full certainty.

An important characteristic of Dec-POMDPs is that the states possess
the \emph{Markov property}. That is, the probability of a particular
next state depends on the current state and joint action, but not
on the whole history:

\begin{equation}
\Pr(s^{\ts+1}|s^{\ts},\ja^{\ts},s^{\ts-1},\ja^{\ts-1},...,s^{0},\ja^{0})=\Pr(s^{\ts+1}|s^{\ts},\ja^{\ts}).\label{eq:T_markov}\end{equation}
Also, we will assume that the transition probabilities are \emph{stationary},
meaning that they are independent of the stage $\ts$.

In a way similar to how the transition model $T$ describes the stochastic
influence of actions on the environment, the observation model $O$
describes how the state of the environment is perceived by the agents.
Formally, $O$ is the observation function, a mapping from joint actions
and successor states to probability distributions over joint observations:
$O:\mathcal{A}\times\mathcal{S}\rightarrow\mathcal{P}(\mathcal{O})$.
I.e., it specifies\begin{equation}
\Pr(\joT{\ts}|\jaT{\ts-1},s^{\ts}).\label{eq:O_P(o|a,s)}\end{equation}
This implies that the observation model also satisfies the Markov
property (there is no dependence on the history). Also the observation
model is assumed stationary: there is no dependence on the stage $\ts$.

Literature has identified different categories of observability \citep{Pynadath02_com_MTDP,GoldmanZ04Categorization}.
When the observation function is such that the individual observation
for all the agents will always uniquely identify the true state, the
problem is considered \emph{fully-} or \emph{individually observable.}
In such a case, a Dec-POMDP effectively reduces to a multiagent MDP
(MMDP) as described by \citet{Boutilier96mmdp}. The other extreme
is when the problem is \emph{non-observable,} meaning that none of
the agents observes any useful information. This is modeled by the
fact that agents always receive a null-observation, $\forall_{i}\;\oAS i=\left\{ \oNullA i\right\} $.
Under non-observability agents can only employ an open-loop plan.
Between these two extremes there are \emph{partially observable} problems.
One more special case has been identified, namely the case where not
the individual, but the joint observation identifies the true state.
This case is referred to as \emph{jointly-} or \emph{collectively
observable}. A jointly observable Dec-POMDP is referred to as a \emph{Dec-MDP}.

The reward function $R(s,\ja)$ is used to specify the goal of the
agents and is a function of states and joint actions. In particular,
a desirable sequence of joint actions should correspond to a high
`long-term' reward, formalized as the return.

\begin{definition} Let the \emph{return} or \emph{cumulative reward}
of a Dec-POMDP be defined as total of the rewards received during
an execution:\begin{equation}
r(0)+r(1)+\dots+r(h-1),\label{eq:r(0)+...+r(h-1)}\end{equation}
where $r(\ts)$ is the reward received at time step $\ts$. \end{definition} 

When, at stage $\ts$, the state is $s^{\ts}$ and the taken joint
action is $\jaT\ts$, we have that $r(\ts)=R(s^{\ts},\ja)$. Therefore,
given the sequence of states and taken joint actions, it is straightforward
to determine the return by substitution of $r(\ts)$ by $R(s^{\ts},\ja)$
in \eqref{eq:r(0)+...+r(h-1)}.

In this paper we consider as optimality criterion the \emph{expected
cumulative reward}, where the expectation refers to the expectation
over sequences of states and executed joint actions. The planning
problem is to find a conditional plan, or \emph{policy,} for each
agent to maximize the optimality criterion. In the Dec-POMDP case
this amounts to finding a tuple of policies, called a \emph{joint
policy} that maximizes the expected cumulative reward.

Note that, in a Dec-POMDP, the agents are assumed not to observe the
immediate rewards: observing the immediate rewards could convey information
regarding the true state which is not present in the received observations,
which is undesirable as all information available to the agents should
be modeled in the observations. When planning for Dec-POMDPs the only
thing that matters is the \emph{expectation} of the cumulative future
reward which is available in the off-line planning phase, not the
actual reward obtained. Indeed, it is not even assumed that the actual
reward can be observed at the end of the episode.

Summarizing, in this work we consider Dec-POMDPs with finite actions
and observation sets and a finite planning horizon. Furthermore,
we consider the general Dec-POMDP setting, without any simplifying
assumptions on the observation, transition, or reward models.

\subsection{Example: Decentralized Tiger Problem}

\label{sub:The-decentralized-tiger}

Here we will describe the decentralized tiger problem introduced by
\citet{Nair03_JESP}. This test problem has been frequently used \citep{Nair03_JESP,Emery-Montemerlo04,Emery-Montemerlo05,Szer05MAA}
and is a modification of the (single-agent) tiger problem \citep{Kaelbling98AI}.
It concerns two agents that are standing in a hallway with two doors.
Behind one of the doors is a tiger, behind the other a treasure. Therefore
there are two states: the tiger is behind the left door ($\sL$) or
behind the right door ($\sR$). Both agents have 3 actions at their
disposal: open the left door ($\aOL$), open the right door ($\aOR$)
and listen ($\aLi$). But they cannot observe each other's actions.
In fact, they can only receive 2 observations. Either they hear a
sound left ($\oHL$) or right ($\oHR$).

At $t=0$ the state is $\sL$ or $\sR$ with probability $0.5$. As
long as no agent opens a door the state doesn't change, when a door
is opened, the state resets to $\sL$ or $\sR$ with probability $0.5$.
The full transition, observation and reward model are listed by \citet{Nair03_JESP}.
The observation probabilities are independent, and identical for both
agents. For instance, when the state is $\sL$ and both perform action
$\aLi$, both agents have a 85\% chance of observing $\oHL$, and
the probability of both hearing the tiger left is $0.85\cdot0.85=0.72$.

When the agents open the door for the treasure they receive a positive
reward, while they receive a penalty for opening the wrong door. When
opening the wrong door jointly, the penalty is less severe. Opening
the correct door jointly leads to a higher reward.

Note that, when the wrong door is opened by one or both agents, they
are attacked by the tiger and receive a penalty. However, neither
of the agents observe this attack nor the penalty and the episode
continues. Arguably, a more natural representation would be to have
the episode end after a door is opened or to let the agents observe
whether they encountered the tiger or treasure, however this is not
considered in this test problem.

\subsection{Histories}

\label{sub:Model:Histories}

As mentioned, the goal of planning in a Dec-POMDP is to find a (near-)
optimal tuple of policies, and these policies specify for each agent
how to act in a specific situation. Therefore, before we define a
policy, we first need to define exactly what these specific situations
are. In essence such situations are those parts of the history of
the process that the agents can observe. 

Let us first consider what the history of the process is. A Dec-POMDP
with horizon~$h$ specifies $h$ time steps or stages $\ts=0,...,h-1$.
At each of these stages, there is a state~$s^{\ts}$, joint observation
$\jo^{\ts}$ and joint action $\ja^{\ts}$. Therefore, when the agents
will have to select their $k$-th actions (at $\ts=k-1$), the history
of the process is a sequence of states, joint observations and joint
actions, which has the following form:\[
\left(s^{0},\jo^{0},\ja^{0},s^{1},\jo^{1},\ja^{1},...,s^{k-1},\jo^{k-1}\right).\]
Here $s^{0}$ is the initial state, drawn according to the initial
state distribution $b^{0}$. The initial joint observation $\jo^{0}$
is assumed to be the empty joint observation: $\jo^{0}=\joNull=\left\langle \oNullA1,...,\oNullA\nrA\right\rangle $.

From this history of the process, the states remain unobserved and
agent $i$ can only observe its own actions and observations. Therefore
an agent will have to base its decision regarding which action to
select on the sequence of actions and observations observed up to
that point.

\begin{definition} We define \emph{the action-observation history
for agent $i$,} $\oaHistA i$, as the sequence of actions taken by
and observations received by agent $i$. At a specific time step $\ts$,
this is:

\[
\oaHistAT i\ts=\left(\oAT i0,\aAT i0,\oAT i1\dots,\aAT i{\ts-1},\oAT i\ts\right).\]
The \emph{joint action-observation history}, $\oaHist,$ is the action-observation
history for all agents:\[
\oaHistT\ts=\langle\oaHistAT1\ts,\dots,\oaHistAT\nrA\ts\rangle.\]
Agent $i$'s set of possible action-observation histories at time
$\ts$ is $\oaHistATS i\ts=\times_{t}(\mathcal{O}_{i}\times\mathcal{A}_{i})$.
The set of all possible action-observation histories for agent $i$
is $\oaHistAS i=\cup_{\ts=0}^{h-1}\oaHistATS i\ts$.%
\footnote{Note that in a particular Dec-POMDP, it may be the case that not all
of these histories can actually be realized, because of the probabilities
specified by the transition and observation model.%
} Finally the set of all possible \emph{joint} action-observation histories
is given by $\oaHistS=\cup_{\ts=0}^{h-1}(\oaHistATS1\ts\times...\times\oaHistATS\nrA\ts)$.
At $\ts=0$, the action-observation history is empty, denoted by $\oaHistT0=\oaHistEmpty$.\end{definition}

We will also use a notion of history only using the observations of
an agent.

\begin{definition} Formally, we define \emph{the observation history
for agent $i$,} $\oHistA i$, as the sequence of observations an
agent has received. At a specific time step $t$, this is:\[
\oHistAT i\ts=\left(o_{i}^{0},o_{i}^{1},\dots,o_{i}^{t}\right).\]
The \emph{joint observation history,} $\oHist,$ is the observation
history for all agents:\[
\oHistT\ts=\langle\oHistAT1\ts,\dots,\oHistAT\nrA\ts\rangle.\]
The set of observation histories for agent $i$ at time $t$ is denoted
$\oHistATS i\ts=\times_{t}\mathcal{O}_{i}$. Similar to the notation
for action-observation histories, we also use $\oHistAS i$ and $\oHistS$
and the empty observation history is denoted $\oHistEmpty$.\end{definition}

Similarly we can define the action history as follows. 

\begin{definition} The \emph{action history for agent $i$,} $\aHistA i$,
is the sequence of actions an agent has performed. At a specific time
step $\ts$, we write:\[
\aHistAT i\ts=\left(\aAT i0,\aAT i1,\dots,\aAT i{\ts-1}\right).\]
\end{definition}

Notation for joint action histories and sets are analogous to those
for observation histories. Also write $\oHistA{\neq i},\oaHistA{\neq i},$
etc. to denote a tuple of observation-, action-observation histories,
etc.\ for all agents except $i$. Finally we note that, clearly,
an (joint) action-observation history consists of an (joint) action-
and an (joint) observation history: $\oaHistT\ts=\left\langle \oHistT\ts,\aHistT\ts\right\rangle $.

\subsection{Policies}

\label{sub:Model:Policies}

As discussed in the previous section, the action-observation history
of an agent specifies all the information the agent has when it has
to decide upon an action. For the moment we assume that an individual
policy $\polA i$ for agent $i$ is a deterministic mapping from action-observation
sequences to actions. 

The number of possible action-observation histories is usually very
large as this set grows exponentially with the horizon of the problem.
At time step $\ts$, there are $\left(\left|\mathcal{A}_{i}\right|\cdot\left|\mathcal{O}_{i}\right|\right)^{\ts}$
action-observation histories for agent $i$. As a consequence there
are a total of \[
\sum_{\ts=0}^{h-1}\left(\left|\mathcal{A}_{i}\right|\cdot\left|\mathcal{O}_{i}\right|\right)^{\ts}=\frac{\left(\left|\mathcal{A}_{i}\right|\cdot\left|\mathcal{O}_{i}\right|\right)^{h}-1}{\left(\left|\mathcal{A}_{i}\right|\cdot\left|\mathcal{O}_{i}\right|\right)-1}\]
of such sequences for agent $i$. Therefore the number of policies
for agent $i$ becomes:\begin{equation}
\left|\mathcal{A}_{i}\right|^{\frac{\left(\left|\mathcal{A}_{i}\right|\cdot\left|\mathcal{O}_{i}\right|\right)^{h}-1}{\left(\left|\mathcal{A}_{i}\right|\cdot\left|\mathcal{O}_{i}\right|\right)-1}},\label{eq:Model:policies:nr_ao_pols}\end{equation}
which is doubly exponential in the horizon $h$.

\subsubsection{Pure and Stochastic Policies}

\begin{figure}
\newcommand{\thissize}{\small}

\psfrag{a1}[ct][cc]{\thissize $\aOL$}
\psfrag{a2}[ct][cc]{\thissize $\aLi$}
\psfrag{a3}[ct][cc]{\thissize $\aOR$}
\psfrag{o1}[cc][cc]{\thissize $\oHL$}
\psfrag{o2}[cc][cc]{\thissize $\oHR$} 
\psfrag{aoh}[lt][lc]{\thissize act.-obs. history}\includegraphics[width=1\columnwidth]{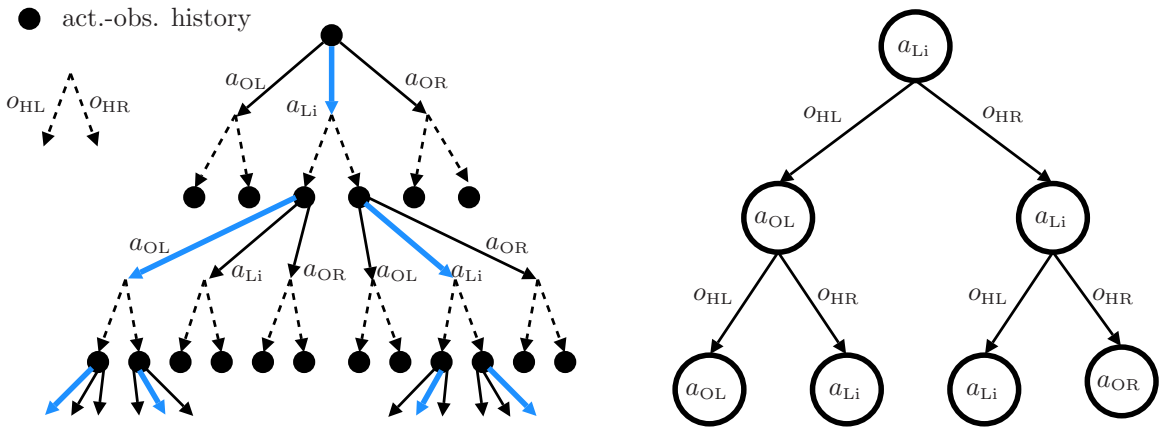}

\caption{A deterministic policy can be represented as a tree. Left: a tree
of action-observation histories $\oaHistA i$ for one of the agents
from the  \<dectig>   problem. An arbitrary deterministic policy
$\polA i$ is highlighted. Clearly shown is that $\polA i$ only reaches
a subset of of histories $\oaHistA i$. ($\oaHistA i$ that are not
reached are not further expanded.) Right: The same policy can be shown
in a simplified policy tree.}

\label{fig:DecPOMDPs:deterministic-pol}
\end{figure}

It is possible to reduce the number of policies under consideration
by realizing that a lot of policies specify the same behavior. This
is illustrated by the left side of \fig\ref{fig:DecPOMDPs:deterministic-pol},
which clearly shows that under a deterministic policy only a subset
of possible action-observation histories are reached. Policies that
only differ with respect to an action-observation history that is
not reached in the first place, manifest the same behavior. The consequence
is that in order to specify a deterministic policy, the observation
history suffices: when an agent takes its action deterministically,
he will be able to infer what action he took from only the observation
history as illustrated by the right side of \fig\ref{fig:DecPOMDPs:deterministic-pol}.

\begin{definition}A \emph{pure} or \emph{deterministic policy}, $\ppol i$,
for agent $i$ in a Dec-POMDP is a mapping from observation histories
to actions, $\ppol i:\oHistAS i\rightarrow\mathcal{A}_{i}$. The set
of pure policies of agent $i$ is denoted $\polAS i$.\end{definition}

\begin{sloppypar}  Note that also for pure policies we sometimes
write $\ppol i(\oaHistA i)$. In this case we mean the action that
$\ppol i$ specifies for the observation history contained in $\oaHistA i$.
For instance, let $\oaHistA i=\left\langle \oHistA i,\aHistA i\right\rangle $,
then $\ppol i(\oaHistA i)=\ppol i(\oHistA i)$. We use $\jpol=\left\langle \pol1,...,\pol\nrA\right\rangle $
to denote a \emph{joint policy}, a profile specifying a policy for
each agent\emph{.} We say that a pure joint policy is an \emph{induced}
or \emph{implicit} mapping from joint observation histories to joint
actions $\jpol:\oHistS\rightarrow\mathcal{A}$. That is, the mapping
is induced by individual policies $\pol i$ that make up the joint
policy. Also we use $\jpol_{\neq i}=\left\langle \pol1,\dots,\pol{i-1},\pol{i+1},\dots,\pol\nrA\right\rangle $,
to denote a profile of policies for all agents but $i$. \end{sloppypar}

Apart from pure policies, it is also possible to have the agents execute
\emph{randomized policies}, i.e., policies that do not always specify
the same action for the same situation, but in which there is an element
of chance that decides which action is performed. There are two types
of randomized policies: mixed policies and stochastic policies.

\begin{definition} A \emph{mixed policy, $\mpol i$,} for an agent
$i$ is a set of pure policies, $\mathcal{M}\subseteq\polS i$, along
with a probability distribution over this set. Thus a mixed policy
$\mpol i\in\mathcal{P}(\mathcal{M})$ is an element of the set of
probability distributions over $\mathcal{M}$. \end{definition}

\begin{definition}A \emph{stochastic} or \emph{behavioral policy},
$\spol i$, for agent $i$ is a mapping from action-observation histories
to probability distributions over actions, $\spol i:\oaHistAS i\rightarrow\PrS(\aAS i)$.\end{definition}

When considering stochastic policies, keeping track of only the observations
is insufficient, as in general all action-observation histories can
be realized. That is why stochastic policies are defined as a mapping
from the full space of action-observation histories to probability
distributions over actions. Note that we use $\pol i$ and $\polS i$
to denote a policy (space) in general, so also for randomized policies.
We will only use $\ppol i$, $\mpol i$ and $\spol i$ when there
is a need to discriminate between different types of policies. 

A common way to represent the temporal structure in a policy is to
split it in \emph{decision rules} $\drA i$ that specify the policy
for each stage. An individual policy is then represented as a sequence
of decision rules $\polA i=(\drAT i0,\dots,\drAT i{h-1})$. In case
of a deterministic policy, the form of the decision rule for stage
$\ts$ is a mapping from length-$\ts$ observation histories to actions
$\drAT i\ts:\oHistATS i\ts\rightarrow\aAS i$.

\subsubsection{Special Cases with Simpler Policies.}

There are some special cases of Dec-POMDPs in which the policy can
be specified in a simpler way. Here we will treat three such cases:
in case the state $s$ is observable, in the single-agent case and
the case that combines the previous two: a single agent in an environment
of which it can observe the state.

The last case, a single agent in a fully observable environment,
corresponds to the regular MDP setting. Because the agent can observe
the state, which is Markovian, the agent does not need to remember
any history, but can simply specify the decision rules $\drA{\Asingle}$
of its policy $\polA{\Asingle}=\left(\drAT{\Asingle}0,\dots,\drAT{\Asingle}{h-1}\right)$
as mappings from states to actions: $\forall\ts\;\drAT{\Asingle}\ts:\mathcal{S}\rightarrow\aAS{\Asingle}$.
The complexity of the policy representation reduces even further in
the infinite-horizon case, where an optimal policy $\polA\Asingle^{*}$
is known to be \emph{stationary}. As such, there is only one decision
rule $\drA{\Asingle}$, that is used for all stages.

The same is true for multiple agents that can observe the state, i.e.,
a fully observable Dec-POMDP as defined in Section \ref{sub:Formal-Model}.
This is essentially the same setting as the \emph{multiagent Markov
decision process (MMDP)} introduced by \citet{Boutilier96mmdp}. In
this case, the decision rules for agent $i$'s policy are mappings
from states to actions $\forall\ts\;\drAT{i}\ts:\mathcal{S}\rightarrow\aAS{i}$,
although in this case some care needs to be taken to make sure no
coordination errors occur when searching for these individual policies.

In a POMDP, a Dec-POMDP with a single agent, the agent cannot observe
the state, so it is not possible to specify a policy as a mapping
from states to actions. However, it turns out that maintaining a probability
distribution over states, called \emph{belief}, $b\in\mathcal{P}(\mathcal{S})$,
is a Markovian signal:\[
P(s^{\ts+1}|a^{\ts},o^{\ts},a^{\ts-1},o^{\ts-1},\dots,a^{0},o^{0})=P(s^{\ts+1}|b^{\ts},a^{\ts}),\]
where the belief $b^{\ts}$ is defined as \[
\forall_{s^{\ts}}\quad b^{\ts}(s^{\ts})\defas\Pr(s^{\ts}|o^{\ts},a^{\ts-1},o^{\ts-1},\dots,a^{0},o^{0})=\Pr(s^{\ts}|b^{\ts-1},a^{\ts-1},o^{\ts}).\]
As a result, a single agent in a partially observable environment
can specify its policy as a series of mappings from the set of beliefs
to actions $\forall\ts\;\drAT{\Asingle}\ts:\PrS(\mathcal{S})\rightarrow\aAS{\Asingle}$.

Unfortunately, in the general case we consider, no such space-saving
simplifications of the policy are possible. Even though the transition
and observation model can be used to compute a \emph{joint} belief,
this computation requires knowledge of the joint actions and observations.
During execution, the agents simply have no access to this information
and thus can not compute a joint belief.

\subsubsection{The Quality of Joint Policies}

Clearly, policies differ in how much reward they can expect to accumulate,
which will serve as a criterion of a joint policy's quality. Formally,
we consider the expected cumulative reward of a joint policy, also
referred to as its \emph{value}.

\begin{definition} The \emph{value} $V(\jpol)$ of a joint policy
$\jpol$ is defined as\begin{equation}
\V(\jpol)\defas E\Bigl[\sum_{\ts=0}^{h-1}R(s^{\ts},\jaT\ts)\Big|\jpol,b^{0}\Bigr],\label{eq:def:V(jpol):Expect}\end{equation}
where the expectation is over states, observations and---in the case
of a randomized $\jpol$---actions.\end{definition}

In particular we can calculate this expectation as \begin{equation}
\V(\jpol)=\sum_{\ts=0}^{h-1}\sum_{\oaHistT\ts\in\oaHistTS\ts}\sum_{s^{\ts}\in\mathcal{S}}\Pr(s^{\ts},\oaHistT\ts|\jpol,b^{0})\sum_{\jaT\ts\in\mathcal{A}}R(s^{\ts},\jaT\ts)\Pr_{\jpol}(\ja^{\ts}|\oaHistT{\ts}),\label{eq:V(pol):iterative}\end{equation}
where $\Pr_{\jpol}(\ja^{\ts}|\oaHistT{\ts})$ is the probability of
$\ja$ as specified by $\jpol$, and where $\Pr(s^{\ts},\oaHistT\ts|\jpol,b^{0})$
is recursively defined as\begin{equation}
\Pr(s^{\ts},\oaHistT\ts|\jpol,b^{0})=\sum_{s^{\ts-1}\in\mathcal{S}}\Pr(s^{\ts},\oaHistT{\ts}|s^{\ts-1},\oaHistT{\ts-1},\jpol)\Pr(s^{\ts-1},\oaHistT{\ts-1}|\jpol,b^{0}),\label{eq:P(s,oaHist|jpol)}\end{equation}
with \begin{equation}
\Pr(s^{\ts},\oaHistT{\ts}|s^{\ts-1},\oaHistT{\ts-1},\jpol)=\Pr(\jo^{\ts}|\ja^{\ts-1},s^{\ts})\Pr(s^{\ts}|s^{\ts-1},\ja^{\ts-1})\Pr_{\jpol}(\ja^{\ts-1}|\oaHistT{\ts-1})\label{eq:P(s,oah|s,oah)}\end{equation}
a term that is completely specified by the transition and observation
model and the joint policy. For stage $0$ we have that $\Pr(s^{0},\oaHistEmpty|\jpol,b^{0})=b^{0}(s^{0})$. 

Because of the recursive nature of $\Pr(s^{\ts},\oaHistT\ts|\jpol,b^{0})$
it is more intuitive to  specify the value recursively:\begin{equation}
V_{\jpol}(s^{\ts},\oaHistT\ts)=\sum_{\ja^{\ts}\in\mathcal{A}}\Pr_{\jpol}(\ja^{\ts}|\oaHistT{\ts})\left[R(s^{\ts},\jaT\ts)+\sum_{s^{\ts+1}\in\mathcal{S}}\sum_{\joT{\ts+1}\in\mathcal{O}}\Pr(s^{\ts+1},\joT{\ts+1}|s^{\ts},\ja^{\ts})V_{\jpol}(s^{\ts+1},\oaHistT{\ts+1})\right],\label{eq:V_jpol(s,oaHist)}\end{equation}
with $\oaHistT{\ts+1}=(\oaHistT{\ts},\ja^{\ts},\joT{\ts+1})$. The
value of joint policy $\jpol$ is then given by \begin{equation}
V(\jpol)=\sum_{s^{0}\in\mathcal{S}}V_{\jpol}(s^{0},\oaHistEmpty)b^{0}(s^{0}).\label{eq:V(pol):recursive}\end{equation}

For the special case of evaluating a pure joint policy $\jpol$, eq.
\eqref{eq:V(pol):iterative} can be written as:\begin{equation}
V(\jpol)=\sum_{\ts=0}^{h-1}\sum_{\oaHistT\ts\in\CoaHistTS{\ts}{}}\Pr(\oaHistT\ts|\jpol,b^{0})R(\oaHistT\ts,\jpol(\oaHistT\ts)),\label{eq:V(pol)}\end{equation}
 where \begin{equation}
R(\oaHistT\ts,\ja^{\ts})=\sum_{s^{\ts}\in\mathcal{S}}R(s^{\ts},\ja^{\ts})\Pr(s^{\ts}|\oaHistT\ts,b^{0})\label{eq:exp_imm_R_R(oaHist,ja)}\end{equation}
denotes the expected immediate reward. In this case, the recursive
formulation \eqref{eq:V_jpol(s,oaHist)} reduces to\begin{equation}
\VjpolT{\jpol}{\ts}(s^{\ts},\oHistT\ts)=R\left(s^{\ts},\pi(\oHistT\ts)\right)+\sum_{s^{\ts+1}\in\mathcal{S}}\sum_{\jo^{\ts+1}\in\mathcal{O}}\Pr(s^{\ts+1},\jo^{\ts+1}|s^{\ts},\pi(\oHistT\ts))\VjpolT{\jpol}{\ts+1}(s^{\ts+1},\oHistT{\ts+1}).\label{eq:V_pure_jpol}\end{equation}

Note that, when performing the computation of the value for a joint
policy recursively, intermediate results should be cached. A particular
$(s^{\ts+1},\oHistT{\ts+1})$-pair (or $(s^{\ts+1},\oaHistT{\ts+1})$-pair
for a stochastic joint policy) can be reached from $\left|\mathcal{S}\right|$
states $s^{\ts}$ of the previous stage. The value $\VjpolT{\jpol}{\ts+1}(s^{\ts+1},\oHistT{\ts+1})$
is the same, however, and should be computed only once.

\subsubsection{Existence of an Optimal Pure Joint Policy}

Although randomized policies may be useful, we can restrict our attention
to pure policies without sacrificing optimality, as shown by the following.

\begin{proposition} \label{prop:1optimalPureJP} A Dec-POMDP has
at least one optimal pure joint policy.

\proofup  See appendix \ref{sub:1optimalPureJP}.\end{proof}\end{proposition}

\section{Overview of Dec-POMDP Solution Methods}

\label{sec:Existing-solution-methods}

In order to provide some background on solving Dec-POMDPs, this section
gives an overview of some recently proposed methods. We will limit
this review to a number of finite-horizon methods for general Dec-POMDPs
that are related to our own approach. 

We will not review the work performed on infinite-horizon Dec-POMDPs,
such as the work by \citet{Peshkin00,Bernstein05,Szer05BestFirstSearch,Amato06msdm,Amato07uai}\textbf{.
}In this setting policies are usually represented by finite state
controllers (FSCs). Since an infinite-horizon Dec-POMDP is undecidable
\citep{Bernstein02Complexity}, this line of work, focuses on finding
$\epsilon$-approximate solutions \citep{Bernstein05_phd} or (near-)
optimal policies for given a particular controller size.

There also is a substantial amount of work on methods exploiting particular
independence assumptions. In particular, transition and observation
independent Dec-MDPs \citep{Becker04TransIndepJAIR,Wu+Durfee06:AAMAS}
and Dec-POMDPs \citep{KimNair06Exploiting_NDPOMDPs,Varakantham07}
have received quite some attention. These models assume that each
agent~$i$ has an individual state space $\mathcal{S}_{i}$ and that
the actions of one agent do not influence the transitions between
the local states of another agent. Although such models are easier
to solve, the independence assumptions severely restrict their applicability.
Other special cases that have been considered are, for instance, goal
oriented Dec-POMDPs \citep{GoldmanZ04Categorization}, event-driven
Dec-MDPs \citep{Becker:04:aamas}, Dec-MDPs with time and resource
constraints \citep{Beynier:05:AAMAS,Beynier:06:AAAI,Marecki:07:AAMAS},
Dec-MDPs with local interactions \citep{Spaan08aamas} and factored
Dec-POMDPs with additive rewards \citep{Oliehoek08aamas}.

A final body of related work which is beyond the scope of this article
are models and techniques for explicit communication in Dec-POMDP
settings \citep{Ooi96,Pynadath02_com_MTDP,Goldman03OptimizingInformation,NairRoth04COM_JESP,Becker05MyopicCom,Roth05AAMAS,Oliehoek07msdm,Roth07aamas,Goldman:07:AAMASjournal:learningComm}.
The Dec-POMDP model itself can model communication actions as regular
actions, in which case the semantics of the communication actions
becomes part of the optimization problem \citep{Xuan01,Goldman03OptimizingInformation,Spaan06aamas}.
In contrast, most approaches mentioned typically assume that communication
happens outside the Dec-POMDP model and with pre-defined semantics.
A typical assumption is that at every time step the agents communicate
their individual observations before selecting an action. \citet{Pynadath02_com_MTDP}
showed that, under assumptions of instantaneous and cost-free communication,
sharing individual observations in such a way is optimal.

\subsection{Brute Force Policy Evaluation }

\label{sub:Brute-force-policy}

Because there exists an optimal pure joint policy for a finite-horizon
Dec-POMDP, it is in theory possible to enumerate all different pure
joint policies, evaluate them using equations \eqref{eq:V(pol):recursive}
and \eqref{eq:V_pure_jpol} and choose the best one. The number of
pure joint policies to be evaluated is:

\begin{equation}
O\left(|\mathcal{A}_{*}|^{\frac{\nrA(|\mathcal{O}_{*}|^{h}-1)}{|\mathcal{O}_{*}|-1}}\right),\label{eq:nr_joint_pols}\end{equation}
 where $|\mathcal{A}_{*}|$ and $|\mathcal{O}_{*}|$ denote the largest
individual action and observation sets. The cost of evaluating each
policy is $O\left(|\mathcal{S}|\cdot|\mathcal{O}|^{h}\right)$. The
resulting total cost of brute-force policy evaluation is

\begin{equation}
O\left(|\mathcal{A}_{*}|^{\frac{\nrA(|\mathcal{O}_{*}|^{h}-1)}{|\mathcal{O}_{*}|-1}}\times|\mathcal{S}|\times|\mathcal{O}_{*}|^{\nrA h}\right),\label{eq:brute_force_dec_pomdp_complex}\end{equation}
which is doubly exponential in the horizon $h$.

\subsection{Alternating Maximization}

\label{sub:JESP}

\citet{Nair03_JESP} introduced \emph{Joint Equilibrium based Search
for Policies (JESP).} This method guarantees to find a locally optimal
joint policy, more specifically, a \emph{Nash equilibrium}: a tuple
of policies such that for each agent $i$ its policy $\polA i$ is
a best response for the policies employed by the other agents $\polA{\neq i}$.
It relies on a process we refer to as \emph{alternating maximization}.
This is a procedure that computes a policy $\polA i$ for an agent
$i$ that maximizes the joint reward, while keeping the policies of
the other agents fixed. Next, another agent is chosen to maximize
the joint reward by finding its best-response to the fixed policies
of the other agents. This process is repeated until the joint policy
converges to a Nash equilibrium, which is a local optimum. The main
idea of fixing some agents and having others improve their policy
was presented before by \citet{Chades02}, but they used a heuristic
approach for memory-less agents. The process of alternating maximization
is also referred to as \emph{hill-climbing} or \emph{coordinate ascent.}

\citet{Nair03_JESP} describe two variants of JESP, the first of which,
Exhaustive-JESP, implements the above idea in a very straightforward
fashion: Starting from a random joint policy, the first agent is chosen.
This agent then selects its best-response policy by evaluating the
joint reward obtained for all of its individual policies when the
other agents follow their fixed policy. 

The second variant, DP-JESP, uses a dynamic programming approach to
compute the best-response policy for a selected agent $i$. In essence,
fixing the policies of all other agents allows for a reformulation
of the problem as an augmented POMDP. In this augmented POMDP a state
$\sAug=\langle s,\oHistAT{\neq i}{}\rangle$ consists of a nominal
state~$s$ and the observation histories of the other agents $\oHistAT{\neq i}{}$.
Given the fixed deterministic policies of other agents $\polA{\neq i}$,
such an augmented state $\sAug$ is a Markovian state, and all transition
and observation probabilities can easily be derived from $\polA{\neq i}$.

Like most methods proposed for Dec-POMDPs, JESP exploits the knowledge
of the initial belief $b^{0}$ by only considering reachable beliefs
$b(\sAug)$ in the solution of the POMDP. However, in some cases the
initial belief might not be available. As demonstrated by \citet{Varakantham06AAMAS},
JESP can be extended to plan for the entire space of initial beliefs,
overcoming this problem.

\subsection{$\MAA$}

\label{sub:MAA*}

\citet{Szer05MAA} introduced a heuristically guided policy search
method called \emph{multiagent A{*} ($\MAA$)}. It performs a guided
A{*}-like search over partially specified joint policies, pruning
joint policies that are guaranteed to be worse than the best (fully
specified) joint policy found so far by an admissible heuristic.

In particular $\MAA$ considers joint policies that are partially
specified with respect to time: a partial joint policy $\partJPolT\ts=(\jdrT0,\jdrT1,\dots,\jdrT{\ts-1})$
specifies the joint decision rules for the first $\ts$ stages. For
such a partial joint policy $\partJPolT\ts$ a heuristic value $\VH(\partJPolT\ts)$
is calculated by taking $\V^{0\dots\ts-1}(\partJPolT\ts)$, the actual
expected reward $\partJPolT\ts$ achieves over the first $\ts$ stages,
and adding $\VH^{\ts\dots h-1}$, a heuristic value for the remaining
$h-\ts$ stages. Clearly when $\VH^{\ts\dots h-1}$ is an \emph{admissible
heuristic---}a guaranteed overestimation---so is $\VH(\partJPolT\ts)$.

$\MAA$ starts by placing the completely unspecified joint policy
$\partJPolT0$ in an open list. Then, it proceeds by selecting partial
joint policies $\partJPolT\ts=(\jdrT0,\jdrT1,\dots,\jdrT{\ts-1})$
from the list and `expanding' them: generating all $\partJPolT{\ts+1}=(\jdrT0,\jdrT1,\dots,\jdrT{\ts-1},\jdrT\ts)$
by appending all possible joint decision rules $\jdrT\ts$ for next
time step ($\ts)$. The left side of \fig\eqref{fig:maa-vs-dp} illustrates
the expansion process. After expansion, all created children are heuristically
valuated and placed in the open list, any partial joint policies $\partJPolT{\ts+1}$
with $\VH(\partJPolT{\ts+1})$ less than the expected value $V(\jpol)$
of some earlier found (fully specified) joint policy $\jpol$, can
be pruned. The search ends when the list becomes empty, at which point
we have found an optimal fully specified joint policy.

\subsection{Dynamic Programming for Dec-POMDPs}

\label{sub:Dynamic-programming-for}%
\begin{figure}
\begin{centering}
\newcommand{\thissize}{\normalsize}

{
\psfrag{a1}[cc][cc]{\thissize $\aone$}
\psfrag{a2}[cc][cc]{\thissize $\atwo$}
\psfrag{o1}[cc][cc]{\thissize $\oone$}
\psfrag{o2}[cc][cc]{\thissize $\otwo$}
\psfrag{old}{\thissize `old' component}
\psfrag{new}{\thissize `new' component}\includegraphics[width=0.8\columnwidth]{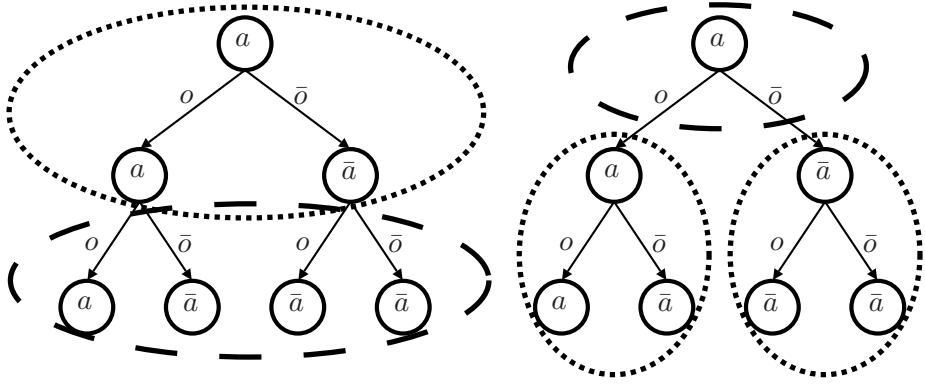}}
\par\end{centering}

\caption{Difference between policy construction in $\MAA$ (left) and dynamic
programming (right) for an agent with actions $\aone,\atwo$ and observations
$\oone,\otwo$. The dashed components are newly generated, dotted
components result from the previous iteration. $\MAA$ `expands' a
partial policy from the leaves, while dynamic programming backs up
a set of `sub-tree policies' forming new ones. 
}

\label{fig:maa-vs-dp}
\end{figure}

$\MAA$ incrementally builds policies from the first stage $\ts=0$
to the last $\ts=h-1$. Prior to this work, \citet{Hansen04} introduced
dynamic programming (DP) for Dec-POMDPs, which constructs policies
the other way around: starting with a set of `1-step policies' (actions)
that can be executed at the last stage, they construct a set of 2-step
policies to be executed at $h-2$, etc.

It should be stressed that the policies maintained are quite different
from those used by $\MAA$. In particular a partial policy in $\MAA$
has the form $\partJPolT\ts=(\jdrT0,\jdrT1,\dots,\jdrT{\ts-1})$.
The policies maintained by DP do not have such a correspondence to
decision rules. We define the \emph{time-to-go} $\ttg$ at stage $\ts$
as \begin{equation}
\ttg=h-\ts.\label{eq:def:ttg}\end{equation}
Now $\stPolAT ik$ denotes a $k$-steps-to-go \emph{sub-tree policy}
for agent $i$. That is, $\stPolAT ik$ is a policy tree that has
the same form as a full policy for the horizon-$k$ problem. Within
the original horizon-$h$ problem $\stPolAT ik$ is a candidate for
execution starting at stage $\ts=h-k$. The set of $k$-steps-to-go
sub-tree policies maintained for agent $i$ is denoted $\stPolATS ik.$
Dynamic programming for Dec-POMDPs is based on backup operations:
constructing $\stPolATS i{k+1}$ a set of sub-tree policies $\stPolAT i{k+1}$
from a set $\stPolATS i{k}$. For instance, the right side of \fig\ref{fig:maa-vs-dp}
shows how $\stPolAT i3$, a 3-steps-to-go sub-tree policy, is constructed
from two $\stPolAT i2\in\stPolATS i2$. Also illustrated is the difference
between this process and $\MAA$ expansion (on the left side).

Dynamic programming consecutively constructs $\stPolATS i1,\stPolATS i2,\dots,\stPolATS ih$
for all agents $i$. However, the size of the set $\stPolATS i{k+1}$
is given by \[
|\stPolATS i{k+1}|=\left|\aAS i\right||\stPolATS i{k}|^{\left|\oAS i\right|},\]
and as a result the sizes of the maintained sets grow doubly exponential
with $k$. To counter this source of intractability, \citet{Hansen04}
propose to eliminate dominated sub-tree policies. The expected reward
of a particular sub-tree policy $\stPolAT ik$ depends on the probability
over states when $\stPolAT ik$ is started (at stage $\ts=h-k$) as
well as the probability with which the other agents $j\neq i$ select
their sub-tree policies $\stPolAT{j}k\in\stPolATS{j}k$. If we let
$\stPolAT{\neq i}k$ denote a sub-tree profile for all agents but
$i$, and $\stPolATS{\neq i}k$ the set of such profiles, we can say
that $\stPolAT ik$ is dominated if it is not maximizing at any point
in the \emph{multiagent belief} space: the simplex over $S\times\stPolATS{\neq i}k$.
\citeauthor{Hansen04} test for dominance over the entire multiagent belief
space by linear programming. Removal of a dominated sub-tree policy
$\stPolAT ik$ of an agent~$i$ may cause a sub-tree policy $\stPolAT jk$
of an other agent $j$ to become dominated. Therefore \citeauthor{Hansen04}
propose to iterate over agents until no further pruning is possible,
a procedure known as \emph{iterated elimination of dominated policies}
\citep{OsborneRubinstein94}.

Finally, when the last backup step is completed the optimal policy
can be found by evaluating all joint policies $\jpol\in\stPolATS1h\times\dots\times\stPolATS\nrA h$
for the initial belief $b^{0}$.

\subsection{Extensions on DP for Dec-POMDPs}

\label{sub:DP_extensions}

In the last few years several extensions to the dynamic programming
algorithm for Dec-POMDPs have been proposed. The first of these extensions
is due to \citet{Szer06_PBDP}. Rather than testing for dominance
over the entire multiagent belief space, \citeauthor{Szer06_PBDP}
propose to  perform point-based dynamic programming (PBDP). In order
to prune the set of sub-tree policies $\stPolATS ik$, the set of
all the belief points $\mathcal{B}_{i,\textrm{reachable}}\subset\PrS(S\times\stPolATS{\neq i}k)$
that can possibly be reached by deterministic joint policies are generated.
Only the sub-tree policies $\stPolAT ik$ that maximize the value
at some $b_{i}\in\mathcal{B}_{i,\textrm{reachable}}$ are kept. The
proposed algorithm is optimal, but intractable because it needs to
generate all the multiagent belief points that are reachable through
all joint policies. To overcome this bottleneck, \citeauthor{Szer06_PBDP}
propose to randomly sample one or more joint policies and use those
to generate $\mathcal{B}_{i,\textrm{reachable}}$.

\citet{SeukenZ07IJCAI} also proposed a point-based extension of the
DP algorithm, called memory-bounded dynamic programming (MBDP). Rather
than using a randomly selected policy to generate the belief points,
they propose to use heuristic policies. A more important difference,
however, lies in the pruning step. Rather than pruning dominated sub-tree
policies $\stPolAT ik$, MBDP prunes all sub-tree policies except
a few in each iteration. More specifically, for each agent \emph{maxTrees
}sub-tree policies are retained, which is a parameter of the planning
method. As a result, MBDP has only linear space and time complexity
with respect to the horizon. The MBDP algorithm still depends on the
exhaustive generation of the sets $\stPolATS i{k+1}$ which now contain
$\left|\aAS i\right|maxTrees^{\left|\oAS i\right|}$ sub-tree policies.
Moreover, in each iteration all $\left(\left|\aAS*\right|maxTrees^{\left|\oAS*\right|}\right)^{\nrA}$
joint sub-tree policies have to be evaluated for each of the sampled
belief points. To counter this growth, \citet{Seuken07IMBDP} proposed
an extension that limits the considered observations during the backup
step to the \emph{maxObs }most likely observations.

Finally, a further extension of the DP for Dec-POMDPs algorithm is
given by \citet{Amato07msdm}. Their approach, bounded DP (BDP), establishes
a bound not on the used memory, but on the quality of approximation.
In particular, BDP uses $\epsilon$-pruning in each iteration. That
is, a $\stPolAT ik$ that is maximizing in some region of the multiagent
belief space, but improves the value in this region by at most $\epsilon$,
is also pruned. Because iterated elimination using $\epsilon$- pruning
can still lead to an unbounded reduction in value, \citeauthor{Amato07msdm}
propose to perform one iteration of $\epsilon$-pruning, followed
by iterated elimination using normal pruning.

\subsection{Other Approaches for Finite-Horizon Dec-POMDPs}

\label{sub:other_DecPOMDP_methods}

There are a few other approaches for finite-horizon Dec-POMDPs, which
we will only briefly describe here. \citet{Aras07icaps} proposed
a mixed integer linear programming formulation for the optimal solution
of finite-horizon Dec-POMDPs. Their approach is based on representing
the set of possible policies for each agent in \emph{sequence form
}\citep{Romanovskii62,Koller94,Koller97}. In sequence form, a single
policy for an agent $i$ is represented as a subset of the set of
`sequences' (roughly corresponding to action-observation histories)
for the agent. As such the problem can be interpreted as a combinatorial
optimization problem, which \citeauthor{Aras07icaps} propose to solve
with a mixed integer linear program.

\citet{Oliehoek07idc} also recognize that finding a solution for
Dec-POMDPs in essence is a combinatorial optimization problem and
propose to apply the Cross-Entropy method \citep{Boer05CEtutor},
a method for combinatorial optimization that recently has become popular
because of its ability to find near-optimal solutions in large optimization
problems. The resulting algorithm performs a sampling-based policy
search for approximately solving Dec-POMDPs. It operates by sampling
pure policies from an appropriately parameterized stochastic policy,
and then evaluates these policies either exactly or approximately
in order to define the next stochastic policy to sample from, and
so on until convergence.

Finally, \citet{Emery-Montemerlo04,Emery-Montemerlo05} proposed to
approximate Dec-POMDPs through series of Bayesian games. Since our
work in this article is based on the same representation, we defer
a detailed explanation to the next section. We do mention here that
while \citeauthor{Emery-Montemerlo04} assume that the algorithm is
run on-line (interleaving planning and execution), no such assumption
is necessary. Rather we will apply the same framework during a off-line
planning phase, just like the other algorithms covered in this overview.

\section{Optimal Q-value Functions}

\label{sec:Q-value-functions-forDec-POMDPs}

In this section we will show how a Dec-POMDP can be modeled as a series
of \emph{Bayesian games (BGs)}. A BG is a game-theoretic model that
can deal with uncertainty \citep{OsborneRubinstein94}. Bayesian games
are similar to the more well-known \emph{normal form}, or \emph{matrix
games,} but allow to model agents that have some private information.
This section will introduce Bayesian games and show how a Dec-POMDP
can be modeled as a series of Bayesian games (BGs). This idea of using
a series of BGs to find policies for a Dec-POMDP has been proposed
in an approximate setting by \citet{Emery-Montemerlo04}. In particular,
they showed that using series of BGs and an approximate payoff function,
they were able to obtain approximate solutions on the \<dectig> problem\textbf{,}
comparable to results for JESP (see Section \ref{sub:JESP}).

The main result of this section is that an optimal Dec-POMDP policy
can be computed from the solution of a sequence of Bayesian games,
if the payoff function of those games coincides with the Q-value function
of an optimal policy $\jpol^{*}$, i.e., with the optimal Q-value
function $Q^{*}$. Thus, we extend the results of \citet{Emery-Montemerlo04}
to include the optimal setting.  Also, we conjecture that this form
of $Q^{*}$ can not be computed without already knowing an optimal
policy $\jpol^{*}$. By transferring the game-theoretic concept of
sequential rationality to Dec-POMDPs, we find a description of $Q^{*}$
that is computable without knowing $\jpol^{*}$ up front.

\subsection{Game-Theoretic Background}

Before we can explain how Dec-POMDPs can be modeled using Bayesian
games, we will first introduce them together with some other necessary
game theoretic background.

\subsubsection{Strategic Form Games and Nash Equilibria}

At the basis of the concept of a Bayesian game lies a simpler form
of game: the \emph{strategic-} or \emph{normal form game}. A strategic
game consists of a set of agents or players, each of which has a set
of actions (or strategies). The combination of selected actions specifies
a particular outcome. When a strategic game consists of two agents,
it can be visualized as a matrix as shown in \fig \ref{fig:strategic-games}.
The first game shown is called `Chicken' and involves two teenagers
who are driving head on. Both have the option to drive on or chicken
out. Each teenager's payoff is maximal ($+2$) when he drives on and
his opponent chickens out. However, if both drive on, a collision
follows giving both a payoff of $-1$. The second game is the meeting
location problem. Both agents want to meet in location A or B. They
have no preference over which location, as long as both pick the same
location. This game is fully cooperative, which is modeled by the
fact that the agents receive identical payoffs.

\begin{figure}
\begin{centering}
\begin{tabular}{c|cc}
 & D & C\tabularnewline
\hline 
D & $-1,-1$ & $+2,0$\tabularnewline
C & $0,+2$ & $+1,+1$\tabularnewline
\end{tabular}\hspace{5em}\begin{tabular}{c|cc}
 & A & B\tabularnewline
\hline 
A & $+2$ & $0$\tabularnewline
B & $0$ & $+2$\tabularnewline
\end{tabular}
\par\end{centering}

\caption{Left: The game `Chicken'. Both players have the option to (D)rive
on or (C)hicken out. Right: The meeting location problem. Because
the game has identical payoffs, each entry contains just one number.}

\label{fig:strategic-games}
\end{figure}

\begin{definition} Formally, a \emph{strategic game} is a tuple $\left\langle \nrA,\mathcal{A},\utF\right\rangle $,
where $\nrA$ is the number of agents, $\mathcal{A}=\times_{i}\aAS i$
is the set of joint actions, and $\utF=\left\langle \utA{1},\dots,\utA\nrA\right\rangle $
with $\utA i:\mathcal{A}\rightarrow\real$ is the payoff function
of agent~$i$.\end{definition}

Game theory tries to specify for each agent how to play. That is,
a game-theoretic solution should suggest a policy for each agent.
In a strategic game we write $\sgPolA i$ to denote a policy for agent
$i$ and $\sgJPol$ for a joint policy. A policy for agent $i$ is
simply one of its actions $\sgPolA i=\aA i\in\aAS i$ (i.e., a pure
policy), or a probability distribution over its actions $\sgPolA i\in\PrS(\aAS i)$
(i.e., a mixed policy). Also, the policy suggested to each agent should
be rational given the policies suggested to the other agent; it would
be undesirable to suggest a particular policy to an agent, if it can
get a better payoff by switching to another policy. Rather, the suggested
policies should form an equilibrium, meaning that it is not profitable
for an agent to unilaterally deviate from its suggested policy. This
notion is formalized by the concept of Nash equilibrium.

{ 
\renewcommand{\aA}[1]{\sgPolA{#1}}
\renewcommand{\ja}{\sgJPol}

\begin{definition}A pure policy profile $\ja=\left\langle \aA1,\dots,\aA i,\dots,\aA\nrA\right\rangle $
specifying a pure policy for each agent is a \emph{Nash Equilibrium
(NE)} if and only if\emph{\begin{equation}
\utA i(\left\langle \aA1,\dots,\aA i,\dots,\aA\nrA\right\rangle )\geq\utA i(\left\langle \aA1,\dots,\aA i',\dots,\aA\nrA\right\rangle ),\quad\forall_{i:1\leq i\leq\nrA},\ \forall_{\aA i'\in\aAS i}.\label{eq:def:NashEquilibrium}\end{equation}
}\end{definition}

}

This definition can be easily extended to incorporate mixed policies
by defining\[
\utA i(\left\langle \sgPolA1,\dots,\sgPolA\nrA\right\rangle )=\sum_{\left\langle \aA1,\dots,,\aA\nrA\right\rangle }\utA i(\left\langle \aA1,\dots,\aA\nrA\right\rangle )\prod_{i=1}^{\nrA}\Pr_{\sgPolA i}(\aA i).\]
\citet{Nash50} proved that when allowing mixed policies, every (finite)
strategic game contains at least one NE, making it a proper solution
for a game. However, it is unclear how such a NE should be found.
In particular, there may be multiple NEs in a game, making it unclear
which one to select. In order to make some discrimination between
Nash equilibria, we can consider NEs such that there is no other NE
that is better for everyone.

{ 
\renewcommand{\aA}[1]{\sgPolA{#1}}
\renewcommand{\ja}{\sgJPol}\begin{definition}A Nash Equilibrium $\ja=\left\langle \aA1,\dots,\aA i,\dots,\aA\nrA\right\rangle $
is referred to as \emph{Pareto Optimal (PO)} when there is no other
NE \emph{$\ja'$} that specifies at least the same payoff for all
agents and a higher payoff for at least one agent:\emph{\[
\nexists_{\ja'}\quad\left(\forall_{i}\;\utA i(\ja')\geq\utA i(\ja)\;\wedge\;\exists_{i}\;\utA i(\ja')>\utA i(\ja)\right).\]
} \end{definition}In the case when multiple Pareto optimal Nash equilibria
exist, the agents can agree beforehand on a particular ordering, to
ensure the same NE is chosen.

}

\subsubsection{Bayesian Games}

A Bayesian game \citep{OsborneRubinstein94} is an augmented normal
form game in which the players hold some private information. This
private information defines the \emph{type} of the agent, i.e., a
particular type $\typeA i\in\typeAS i$ of an agent $i$ corresponds
to that agent knowing some particular information. The payoff the
agents receive now no longer only depends on their actions, but also
on their private information. Formally, a BG is defined as follows:

\begin{definition}  A \emph{Bayesian game (BG)} is a tuple $\left\langle \nrA,\mathcal{A},\jtypeS,\Pr(\jtypeS),\left\langle \utA1,...\utA\nrA\right\rangle \right\rangle $,
where $\nrA$ is the number of agents, $\mathcal{A}$ is the set of
joint actions, $\jtypeS=\times_{i}\typeAS i$ is the set of joint
types over which a probability function  $\Pr(\jtypeS)$ is specified,
and $\utA i:\jtypeS\times\mathcal{A}\rightarrow\real$ is the payoff
function of agent~$i$.\end{definition}

In a normal form game the agents select an action. Now, in a BG the
agents can condition their action on their private information. This
means that in BGs the agents use a different type of policies. For
a BG, we denote a joint policy $\jpolBG=\left\langle \polBG1,...,\polBG\nrA\right\rangle $,
where the individual policies are mappings from types to actions:
$\polBG i:\typeAS i\rightarrow\aAS i$. In the case of identical payoffs
for the agents, the solution of a BG is given by the following theorem:

\begin{theorem}

For a BG with identical payoffs, i.e., $\forall_{i,j}\forall_{\jtype}\forall_{\ja}\;\utA i(\jtype,\ja)=\utA j(\jtype,\ja)$,
the solution is given by:\begin{equation}
\jpolBG^{*}=\argmax_{\jpolBG}\sum_{\jtype\in\jtypeS}\Pr(\jtype)\utF(\jtype,\jpolBG(\jtype)),\label{eq:BG_solution}\end{equation}
 where $\jpolBG(\jtype)=\left\langle \polBG1(\typeA1),...,\polBG\nrA(\typeA\nrA)\right\rangle $
is the joint action specified by $\jpolBG$ for joint type $\jtype$.
This solution constitutes a Pareto optimal Nash equilibrium.

\begin{proof} \upshape The proof consists of two parts: the first
shows that $\jpolBG^{*}$ is a Nash equilibrium, the second shows
it is Pareto optimal.

\paragraph*{Nash equilibrium proof.}

It is clear that $\jpolBG^{*}$ satisfying \ref{eq:BG_solution} is
a Nash equilibrium by rewriting from the perspective of an arbitrary
agent $i$ as follows:

\begin{eqnarray*}
\polBG i^{*} & = & \argmax_{\polBG i}\Biggl[\max_{\polBG{\neq i}}\sum_{\jtype\in\jtypeS}\Pr(\jtype)\utF(\jtype,\jpolBG(\jtype))\Biggr],\\
 & = & \argmax_{\polBG i}\Biggl[\max_{\polBG{\neq i}}\sum_{\typeA{i}}\sum_{\typeA{\neq i}}\Pr(\typeA{\neq i}|\typeA i)\underbrace{\Biggl[\sum_{\typeA{\neq i}}\Pr(\left\langle \typeA i,\typeA{\neq i}\right\rangle )\Biggr]}_{\Pr(\typeA i)}\utF(\jtype,\jpolBG(\jtype))\Biggr],\\
 & = & \argmax_{\polBG i}\Biggl[\max_{\polBG{\neq i}}\sum_{\typeA{i}}\Pr(\typeA i)\sum_{\typeA{\neq i}}\Pr(\typeA{\neq i}|\typeA i)\utF(\jtype,\jpolBG(\jtype))\Biggr],\\
 & = & \argmax_{\polBG i}\sum_{\typeA{i}}\Pr(\typeA i)\sum_{\typeA{\neq i}}\Pr(\typeA{\neq i}|\typeA i)\utF(\left\langle \typeA i,\typeA{\neq i}\right\rangle ,\left\langle \polBG{i}(\typeA{i}),\polBG{\neq i}^{*}(\typeA{\neq i})\right\rangle ),\end{eqnarray*}
which means that $\polBG i^{*}$ is a best response for $\polBG{\neq i}^{*}$.
Since no special assumptions were made on $i$, it follows that $\jpolBG^{*}$
is a Nash equilibrium.

\paragraph*{Pareto optimality proof.}

Let us write $V_{\typeA{i}}(\aA i,\polBG{\neq i})$ for the payoff
agent $i$ expects for $\typeA{i}$ when performing $\aA i$ when
the other agents use policy profile $\polBG{\neq i}$. We have that

\[
V_{\typeA{i}}(\aA i,\polBG{\neq i})=\sum_{\typeA{\neq i}}\Pr(\typeA{\neq i}|\typeA i)\utF(\left\langle \typeA i,\typeA{\neq i}\right\rangle ,\left\langle \aA i,\polBG{\neq i}(\typeA{\neq i})\right\rangle ).\]
Now, a joint policy $\jpolBG^{*}$ satisfying \eqref{eq:BG_solution}
is not Pareto optimal if and only if there is another Nash equilibrium
$\jpolBG'$ that attains at least the same payoff for all agents $i$
and for all types $\typeA i$ and strictly more for at least one agent
and type. Formally $\jpolBG^{*}$ is not Pareto optimal when $\exists\jpolBG'$
such that: \begin{equation}
\forall_{i}\forall_{\typeA{i}}\quad V_{\typeA{i}}(\polBG i^{*}(\typeA i),\polBG{\neq i}^{*})\leq V_{\typeA{i}}(\polBG{i}{}'(\typeA i),\polBG{\neq i}{}')\;\wedge\;\exists_{i}\exists_{\typeA{i}}V_{\typeA{i}}(\polBG i^{*}(\typeA i),\polBG{\neq i}^{*})<V_{\typeA{i}}(\polBG{i}{}'(\typeA i),\polBG{\neq i}{}').\label{eq:pareto_def}\end{equation}

We prove that no such $\jpolBG'$ can exist by contradiction. Suppose
that $\jpolBG{}'=\langle\polBG i{}',\polBG{\neq i}^{{\prime}}\rangle$
is a NE such that \eqref{eq:pareto_def} holds (and thus $\jpolBG^{*}$
is not Pareto optimal). Because $\jpolBG^{*}$ satisfies \eqref{eq:BG_solution}
we know that: \begin{equation}
\sum_{\jtype\in\jtypeS}\Pr(\jtype)\utF(\jtype,\jpolBG^{*}(\jtype))\geq\sum_{\jtype\in\jtypeS}\Pr(\jtype)\utF(\jtype,\jpolBG'(\jtype)),\label{eq:par_proof_1}\end{equation}
and therefore, for all agents $i$\begin{multline*}
\Pr(\typeA{i,1})V_{\typeA{i,1}}(\polBG i^{*}(\typeA{i,1}),\polBG{\neq i}^{*})+...+\Pr(\typeA{i,|\typeAS i|})V_{\typeA{i,|\typeAS i|}}(\polBG i^{*}(\typeA{i,|\typeAS i|}),\polBG{\neq i}^{*})\geq\\
\Pr(\typeA{i,1})V_{\typeA{i,1}}(\polBG i^{\prime}(\typeA{i,1}),\polBG{\neq i}^{\prime})+...+\Pr(\typeA{i,|\typeAS i|})V_{\typeA{i,|\typeAS i|}}(\polBG i^{\prime}(\typeA{i,|\typeAS i|}),\polBG{\neq i}^{\prime})\end{multline*}
holds. However, by assumption that $\jpolBG{}'$ satisfies \eqref{eq:pareto_def}
we get that \[
\exists_{j}\quad V_{\typeA{i,j}}(\polBG i^{*}(\typeA{i,j}),\polBG{\neq i}^{*})<V_{\typeA{i,j}}(\polBG i^{\prime}(\typeA{i,j}),\polBG{\neq i}^{\prime}).\]
Therefore it must be that \[
\sum_{k\neq j}\Pr(\typeA{i,k})V_{\typeA{i,k}}(\polBG i^{*}(\typeA{i,k}),\polBG{\neq i}^{*})>\sum_{k\neq j}\Pr(\typeA{i,k})V_{\typeA{i,k}}(\polBG i^{\prime}(\typeA{i,k}),\polBG{\neq i}^{{\prime}}),\]
and thus that \[
\exists_{k}\quad V_{\typeA{i,k}}(\polBG i^{*}(\typeA{i,k}),\polBG{\neq i}^{*})>V_{\typeA{i,k}}(\polBG i^{\prime}(\typeA{i,k}),\polBG{\neq i}^{{\prime}}),\]
contradicting the assumption that $\jpolBG{}'$ satisfies \eqref{eq:pareto_def}.
\end{proof}

\end{theorem}

\subsection{Modeling Dec-POMDPs with Series of Bayesian Games}

\label{sub:Modeling-Dec-POMDPs-with-BGs}

\begin{figure}
\begin{centering}
{\input{notation.DecPOMDP}
\providecommand{\size}{\normalsize \small}
\psfrag{t0}{\size $\ts=0$}
\psfrag{t1}{\size $\ts=1$}
\psfrag{BG0}{\size } 
\psfrag{BG1}{\size } 
\psfrag{BG1b}{\size }
\psfrag{act}{\size joint actions}
\psfrag{obs}{\size joint observations}
\psfrag{dec}{\size joint act.-obs. history}

\psfrag{a1a1}{\size $\langle \aoo,\ato \rangle$}
\psfrag{a2a1}[tl][tl]{\size $\langle \aot,\ato \rangle$}
\psfrag{a1a2}{\size $\langle \aoo,\att \rangle$}
\psfrag{a2a2}{\size $\langle \aot,\att \rangle$}

\psfrag{o1o1}[cc][br]{\size $\langle \ooo,\oto \rangle$}
\psfrag{o2o1}[cc][b]{\size $\langle \oot,\oto \rangle$}
\psfrag{o1o2}{\size $\langle \ooo,\ott \rangle$}
\psfrag{o2o2}{\size $\langle \oot,\ott \rangle$}\includegraphics[width=0.65\columnwidth,keepaspectratio]{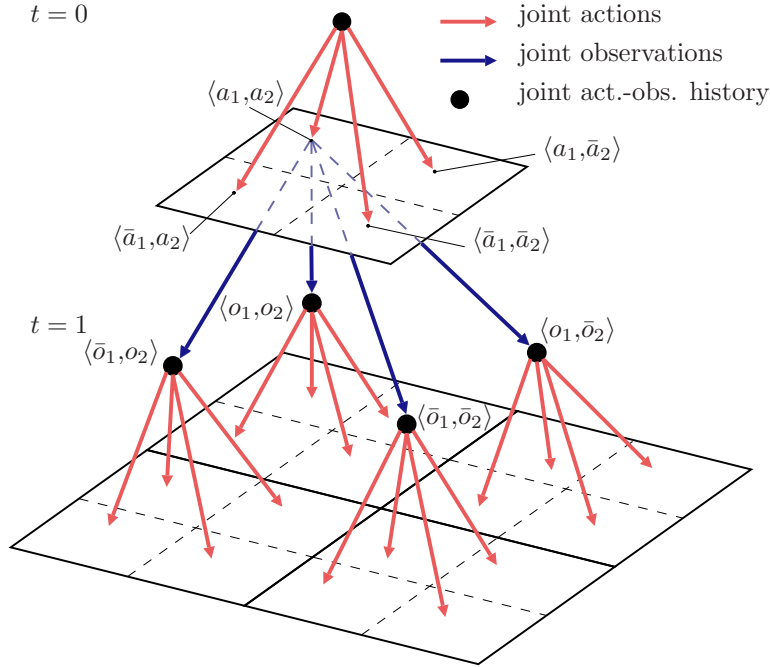}}
\par\end{centering}

\caption{A Dec-POMDP can be seen as a tree of joint actions and observations.
The indicated planes correspond with the Bayesian games for the first
two stages.}

\label{fig:A-Dec-POMDP-is-a-tree}
\end{figure}

Now we will discuss how Bayesian games can be used to model Dec-POMDPs.
Essentially, a Dec-POMDP can be seen as a tree where nodes are joint
action-observation histories and edges represent joint actions and
observations, as illustrated in \fig \ref{fig:A-Dec-POMDP-is-a-tree}.
At a specific stage~$\ts$ in a Dec-POMDP, the main difficulty in
coordinating action selection is presented by the fact that each agent
has its own individual action-observation history. That is, there
is no global signal that the agents can use to coordinate their actions.
This situation can be conveniently modeled by a Bayesian game as we
will now discuss. 

At a time step $\ts$, one can directly associate the primitives of
a Dec-POMDP with those of a BG with identical payoffs: the actions
of the agents are the same in both cases, and the types of agent $i$
correspond to its action-observation histories $\typeAS i\defas\oaHistATS i\ts$.
Figure \ref{tab:Bayesian-game} shows the Bayesian games for $\ts=0$
and $\ts=1$ for a fictitious Dec-POMDP with 2 agents. 

We denote the payoff function of the BG that models a stage of a Dec-POMDP
by $Q(\oaHistT\ts,\ja)$. This payoff function should be naturally
defined in accordance with the value function of the planning task.
For instance, \citet{Emery-Montemerlo04} define $Q(\oaHistT\ts,\ja)$
as the $\QMDP$-value of the underlying MDP. We will more extensively
discuss the payoff function in Section \ref{sub:The-optimal-Q-value}.

The probability $\Pr(\jtype)$ is equal to the probability of the
joint action-observation history to which $\jtype$ corresponds and
depends on the past joint policy $\pP^{\ts}=(\jdrT0,\dots,\jdrT{\ts-1})$
and the initial state distribution. It can be calculated as the marginal
of \eqref{eq:P(s,oaHist|jpol)}:\begin{equation}
\Pr(\jtype)=\Pr(\oaHistT{\ts}|\pP^{\ts},b^{0})=\sum_{s^{\ts}\in\mathcal{S}}\Pr(s^{\ts},\oaHistT{\ts}|\pP^{\ts},b^{0}).\label{eq:P(oaHist)}\end{equation}

When only considering pure joint policies $\pP^{\ts}$, the  action
probability component $\Pr_{\pP}(\ja|\oaHist)$ in \eqref{eq:P(s,oaHist|jpol)}
is $1$ for joint action-observation histories $\oaHistT\ts$ that
are `consistent' with the past joint policy $\pP^{\ts}$ and $0$
otherwise. We say that an action-observation $\oaHistA i$ history
is consistent with a pure policy $\polA i$ if it can occur when executing
$\polA i$, i.e., when the actions in $\oaHistA i$ would be selected
by $\polA i$. Let us more formally define this consistency as follows.

\begin{definition}[Consistency] \label{def:consistency} Let us write
$\oaHistAT i{\ts'}$ for the restriction of $\oaHistAT i\ts$ to stage
$0,\dots,\ts'$ (with $0\leq\ts'<\ts$). An action-observation history
$\oaHistAT i\ts$ of agent $i$ is \emph{consistent} with a pure policy
$\polA i$ if and only if at each time step $\ts'$ with $0\leq\ts'<\ts$\[
\polA i(\oaHistAT i{\ts'})=\polA i(\oHistAT i{\ts'})=\aAT i{\ts'}\]
is the $(\ts'+1)$-th action in $\oaHistAT i\ts$. A joint action-observation
history $\oaHistT\ts=\langle\oaHistAT1\ts,\dots,\oaHistAT\nrA\ts\rangle$
is consistent with a pure joint policy $\jpol=\langle\polA1,\dots,\polA\nrA\rangle$
if each individual $\oaHistAT i\ts$ is consistent with the corresponding
individual policy $\polA i$. $C$ is the indicator function for consistency.
For instance $C(\oaHistT{\ts},\jpol)$ `filters out' the action-observation
histories $\oaHistT\ts$ that are inconsistent with a joint pure policy
$\jpol$:

\begin{equation}
C(\oaHistT{\ts},\jpol)=\begin{cases}
1 & \textrm{, }\oaHistT{\ts}=\left(\jo^{0},\jpol(\jo^{0}),\jo^{1},\jpol(\jo^{0},\jo^{1}),...\right)\textrm{)}\\
0 & \textrm{, otherwise.}\end{cases}\label{eq:C_oaHist_pol}\end{equation}
We will also write $\CoaHistTS{\ts}{\jpol}\defas\{\oaHistT\ts\mid C(\oaHistT{\ts},\jpol)=1\}$
for the set of $\oaHistT\ts$ consistent with $\jpol$. \end{definition}

This definition allows us to write\begin{equation}
\Pr(\oaHistT\ts|\pP^{\ts},b^{0})=C(\oaHistT{\ts},\pP^{\ts})\sum_{s^{\ts}\in\mathcal{S}}\Pr(s^{\ts},\oaHistT{\ts}|b^{0})\label{eq:P(oaHist)_purePol}\end{equation}
with \begin{equation}
\Pr(s^{\ts},\oaHistT\ts|b^{0})=\sum_{s^{\ts-1}\in\mathcal{S}}\Pr(\jo^{\ts}|\ja^{\ts-1},s^{\ts})\Pr(s^{\ts}|s^{\ts-1},\ja^{\ts-1})\Pr(s^{\ts-1},\oaHistT{\ts-1}|b^{0}).\label{eq:P(s,oaHist)}\end{equation}
Figure \ref{tab:Bayesian-game} illustrates how the indicator function
`filters out' policies, when $\jpol^{\ts=0}(\oaHistT{\ts=0})=\left\langle \aoo,\ato\right\rangle $,
only the non-shaded part of the BG for $\ts=1$ `can be reached' (has
positive probability). %
\begin{figure}
\begin{centering}
{
\footnotesize

\noindent {

\renewcommand{\multirowsetup}{\centering}
\setlength\arrayrulewidth{\thickertableline}\arrayrulecolor{black}

\begin{tabular}{cc|cc}
&
$\oaHistAT2{\ts=0}$&
\multicolumn{2}{c}{$\left(\right)$}\tabularnewline
$\oaHistAT1{\ts=0}$&
&
$\ato$&
$\att$\tabularnewline
\hline
\multirow{2}{0.3cm}{ $\left(\right)$ }&
$\aoo$&
\cellcolor{tablemid} $+2.75$&
$-4.1$\tabularnewline
&
$\aot$&
$-0.9$&
$+0.3$\tabularnewline
\end{tabular}\\
~\\
~\\
\begin{tabular}{cc|cc|cc|c}
&
$\oaHistAT2{\ts=1}$&
\multicolumn{2}{c|}{$\left(\ato,\oto\right)$}&
\multicolumn{2}{c|}{$\left(\ato,\ott\right)$}&
...\tabularnewline
$\oaHistAT1{\ts=1}$&
&
$\ato$&
$\att$&
$\ato$&
$\att$&
\tabularnewline
\hline
\multirow{2}{0.9cm}{$\left(\aoo,\ooo\right)$}&
$\aoo$&
$-0.3$&
$+0.6$&
$-0.6$&
$+4.0$&
\cellcolor{tabledark} ...\tabularnewline
&
$\aot$&
$-0.6$&
\cellcolor{tablemid} $+2.0$&
$-1.3$&
\cellcolor{tablemid} $+3.6$&
\cellcolor{tabledark} ...\tabularnewline
\hline
\multirow{2}{0.9cm}{$\left(\aoo,\oot\right)$}&
$\aoo$&
$+3.1$&
\cellcolor{tablemid} $+4.4$&
$-1.9$&
\cellcolor{tablemid} $+1.0$&
\cellcolor{tabledark} ...\tabularnewline
&
$\aot$&
$+1.1$&
$-2.9$&
$+2.0$&
$-0.4$&
\cellcolor{tabledark} \tabularnewline
\hline
\multirow{2}{0.9cm}{$\left(\aot,\ooo\right)$}&
$\aoo$&
\cellcolor{tabledark} $-0.4$&
\cellcolor{tabledark} $-0.9$&
\cellcolor{tabledark} $-0.5$&
\cellcolor{tabledark} $-1.0$&
\cellcolor{tabledark} ...\tabularnewline
&
$\aot$&
\cellcolor{tabledark} $-0.9$&
\cellcolor{tabledark} $-4.5$&
\cellcolor{tabledark} $-1.0$&
\cellcolor{tabledark} $+3.5$&
\cellcolor{tabledark} ...\tabularnewline
\hline
\multirow{1}{0.9cm}{$\left(\aot,\oot\right)$}&
...&
\cellcolor{tabledark} ...&
\cellcolor{tabledark} ...&
\cellcolor{tabledark} ...&
\cellcolor{tabledark} ...&
\cellcolor{tabledark} ...\tabularnewline
\end{tabular}

}

} 
\par\end{centering}

\caption{The Bayesian game for the first and second time step (top: $\ts=0$,
bottom: $\ts=1$). The entries $\oaHistT\ts$, $\ja^{\ts}$ are given
by the payoff function $Q(\oaHistT\ts,\ja^{\ts})$. Light shaded entries
indicate the solutions. Dark entries will not be realized given $\left\langle \aoo,\ato\right\rangle $
the solution of the BG for $\ts=0$. }

\label{tab:Bayesian-game}
\end{figure}

\subsection{The Q-value Function of an Optimal Joint Policy}

\label{sub:The-optimal-Q-value}

Given the perspective of a Dec-POMDP interpreted as a series of BGs
as outlined in the previous section, the solution of the BG for stage
$\ts$ is a joint decision rule $\jdrT\ts$. If the payoff function
for the BG is chosen well, the quality of $\jdrT\ts$ should be high.
\citet{Emery-Montemerlo04} try to find a good joint policy $\jpol=(\jdrT0,\dots,\jdrT{h-1})$
by a procedure we refer to as \emph{forward-sweep policy computation
(FSPC)}: in one sweep forward through time, the BG for each stage
$\ts=0,1,\dots,h-1$ is consecutively solved. As such, the payoff
function for the BGs constitute what we call a Q-value function for
the Dec-POMDP.

Here, we show that there is an optimal Q-value function $Q^{*}$:
when using this $Q^{*}$ as the payoff functions for the BGs, forward-sweep
policy computation will lead to an optimal joint policy $\jpol^{*}=(\jdrT{0,*},\dots,\jdrT{h-1,*})$.
We first give a derivation of this $Q^{*}$. Next, we will discuss
that $Q^{*}$ can indeed be used to calculate $\jpol^{*}$, but computing
$Q^{*}$ seems impractical without already knowing an optimal joint
policy $\jpol^{*}$. This issue will be further addressed in Section~\ref{sub:Sequential-rationality-for}.

\subsubsection{Existence of {\large
 $Q^{*}$ }}

\label{sub:Q*}

We now state a theorem identifying a normative description of $Q^{*}$
as the Q-value function for an optimal joint  policy.

\begin{theorem} The expected cumulative reward over stages $\ts,\dots,h-1$
induced by $\jpol^{*}$, an optimal joint policy for a Dec-POMDP,
is given by:\begin{equation}
V^{\ts}(\jpol^{*})=\sum_{\oaHistT{\ts}\in\CoaHistTS{\ts}{\jpol^{*}}}\Pr(\oaHistT{\ts}|b^{0})Q^{*}(\oaHistT{\ts},\jpol^{*}(\oaHistT{\ts})),\label{eq:Vt(jpol)}\end{equation}
where $\oaHistT\ts=\left\langle \oHistT\ts,\aHistT\ts\right\rangle $,
where $\jpol^{*}(\oaHistT{\ts})=\jpol^{*}(\oHistT{\ts})$ denotes
the joint action that pure joint policy $\jpol^{*}$ specifies for
$\oHistT{\ts}$,  and where\begin{equation}
Q^{*}(\oaHistT{\ts},\ja)=R(\oaHistT{\ts},\ja)+\sum_{\jo^{\ts+1}\in\mathcal{O}}\Pr(\jo^{\ts+1}|\oaHistT{\ts},\ja)Q^{*}(\oaHistT{\ts+1},\jpol^{*}(\oaHistT{\ts+1}))\label{eq:Qt(oaHist_ja)}\end{equation}
is the Q-value function for $\jpol^{*}$, which gives the expected
cumulative future reward when taking joint action $\ja$ at $\oaHistT{\ts}$
given that an optimal joint  policy $\jpol^{*}$ is followed hereafter.

\proofup 
 By filling out \eqref{eq:V(pol)} for an optimal pure joint policy
$\jpol^{*}$, we obtain its expected cumulative reward as the summation
of $E\left[R(s^{\ts},\ja^{\ts})\big|\jpol^{*}\right]$ the expected
rewards it yields for each time step:\begin{equation}
V(\jpol^{*})=\sum_{\ts=0}^{h-1}E\left[R(s^{\ts},\ja^{\ts})\big|\jpol^{*}\right]=\sum_{\ts=0}^{h-1}\sum_{\oaHistT\ts\in\CoaHistTS{\ts}{}}\Pr(\oaHistT\ts|\jpol^{*},b^{0})R(\oaHistT\ts,\jpol^{*}(\oaHistT\ts)).\label{eq:V(pure_pol*)}\end{equation}
In this equation, $\Pr(\oaHistT\ts|\jpol^{*},b^{0})$ is given by
\eqref{eq:P(oaHist)_purePol}. As a result, the influence of $\jpol^{*}$
on $\Pr(\oaHistT\ts|\jpol^{*},b^{0})$ is only through $C$. I.e.,
$\jpol^{*}$ is only used to `filter out' inconsistent histories.
Therefore we can write:\begin{equation}
E\left[R(s^{\ts},\ja^{\ts})\big|\jpol^{*}\right]=\sum_{\oaHistT{\ts}\in\CoaHistTS{\ts}{\jpol^{*}}}\Pr(\oaHistT\ts|b^{0})R(\oaHistT\ts,\jpol^{*}(\oaHistT\ts)),\label{eq:ERt_sum_st}\end{equation}
where $\Pr(\oaHistT\ts|b^{0})$ is given by directly taking the marginal
of \eqref{eq:P(s,oaHist)}. Now, let us define the value starting
from time step $\ts$:\begin{multline}
V^{\ts}(\jpol^{*})=E\left[R(s^{\ts},\ja^{\ts})\big|\jpol^{*}\right]+V^{\ts+1}(\jpol^{*})=\sum_{\oaHistT{\ts}\in\CoaHistTS{\ts}{\jpol^{*}}}\Pr(\oaHistT{\ts}|b^{0})R(\oaHistT\ts,\jpol^{*}(\oaHistT\ts))+V^{\ts+1}(\jpol^{*}).\end{multline}
For the last time step $h-1$ there is no expected future reward,
so we get:\begin{equation}
V^{h-1}(\jpol^{*})=\sum_{\oaHistT{h-1}\in\CoaHistTS{h-1}{\jpol^{*}}}\Pr(\oaHistT{h-1}|b^{0})\underbrace{R(\oaHistT{h-1},\jpol^{*}(\oaHistT{h-1}))}_{Q^{*}(\oaHistT{h-1},\jpol^{*}(\oaHistT{h-1}))}.\label{eq:Vh-1(pol*)}\end{equation}
For time step $h-2$ this becomes:\begin{multline}
V^{h-2}(\jpol^{*})\defas E\left[R(s^{h-2},\ja^{h-2})\big|\jpol^{*}\right]+V^{h-1}(\jpol^{*})=\\
\sum_{\oaHistT{h-2}\in\CoaHistTS{h-2}{\jpol^{*}}}\Pr(\oaHistT{h-2}|b^{0})R(\oaHistT{h-2},\jpol^{*}(\oaHistT{h-2}))+\sum_{\oaHistT{h-1}\in\CoaHistTS{h-1}{\jpol^{*}}}\Pr(\oaHistT{h-1}|b^{0})Q^{*}(\oaHistT{h-1},\jpol^{*}(\oaHistT{h-1})).\label{eq:Vh-2(pol*)}\end{multline}
 Because $\Pr(\oaHistT{h-1})=\Pr(\oaHistT{h-2})\Pr(\jo^{h-1}|\oaHistT{h-2},\jpol^{*}(\oaHistT{h-2}))$,
\eqref{eq:Vh-2(pol*)} can be rewritten to:\begin{equation}
V^{h-2}(\jpol^{*})=\sum_{\oaHistT{h-2}\in\CoaHistTS{h-2}{\jpol^{*}}}\Pr(\oaHistT{h-2}|b^{0})Q^{*}(\oaHistT{h-2},\jpol^{*}(\oaHistT{h-2})),\label{eq:Vh-2(pol*)_2}\end{equation}
 with\begin{multline}
Q^{*}(\oaHistT{h-2},\jpol^{*}(\oaHistT{h-2}))=R(\oaHistT{h-2},\jpol^{*}(\oaHistT{h-2}))+\\
\sum_{\jo^{h-1}}\Pr(\jo^{h-1}|\oaHistT{h-2},\jpol^{*}(\oaHistT{h-2}))Q^{*}(\oaHistT{h-1},\jpol^{*}(\oaHistT{h-1})).\vphantom{\sum_{\jo^{h-1}}}\label{eq:Qh-2(oahist,jpol*)}\end{multline}
Reasoning in the same way we see that \eqref{eq:Vt(jpol)} and \eqref{eq:Qt(oaHist_ja)}
constitute a generic expression for the expected cumulative future
reward starting from time step $\ts$. \end{proof}\end{theorem}

Note that in the above derivation, we explicitly included $b^{0}$
as one of the given arguments. In the rest of this text, we will always
assume $b^{0}$ is given and therefore omit it, unless necessary.

\subsubsection{Deriving an Optimal Joint Policy from {\large$Q^{*}$}}

\label{sub:Calculating-the-optimal-joint-pol-from-Q*}

At this point we have derived $Q^{*}$, a Q-value function for an
optimal joint policy. Now, we extend the results of \citet{Emery-Montemerlo04}
into the exact setting:

\begin{theorem} Applying forward-sweep policy computation using $Q^{*}$
as defined by \eqref{eq:Qt(oaHist_ja)} yields an optimal joint policy.

\proofup 
Note that, per definition, the optimal Dec-POMDP policy $\jpol^{*}$
maximizes the expected future reward $V^{\ts}(\jpol^{*})$ specified
by \eqref{eq:Vt(jpol)}. Therefore $\jdrT{\ts,*}$, the optimal decision
rule for stage~$\ts$, is identical to an optimal joint policy $\jpolBG^{\ts,*}$
for the Bayesian game for time step $\ts$, if the payoff function
of the BG is given by $Q^{*}$, that is:

\begin{equation}
\jdrT{\ts,*}\defas\jpolBG^{\ts,*}=\argmax_{\jpolBG^{\ts}}\sum_{\oaHistT\ts\in\CoaHistTS{\ts}{\jpol^{*}}}\Pr(\oaHistT{\ts})Q^{*}(\oaHistT{\ts},\jpolBG^{\ts}(\oaHistT{\ts})).\label{eq:BG_ts:jpol_is_argmax}\end{equation}
Equation \eqref{eq:BG_ts:jpol_is_argmax} tells us that $\jdrT{\ts,*}\defas\jpolBG^{\ts,*}$.
This means that it is possible to construct the complete optimal Dec-POMDP
policy $\jpol^{*}=(\jdrT{0,*},\dots,\jdrT{h-1,*})$, by computing
$\jdrT{\ts,*}$ for all $\ts$.\end{proof}\end{theorem}

A subtlety in the calculation of $\jpol^{*}$ is that \eqref{eq:BG_ts:jpol_is_argmax}
itself is dependent on an optimal joint policy, as the summation is
over all $\oaHistT{\ts}\in\CoaHistTS{\ts}{\jpol^{*}}\defas\{\oaHistT\ts\mid C(\oaHistT{\ts},\jpol^{*})=1\}$.
This is resolved by realizing that only the past actions influence
which action-observation histories can be reached at time step $\ts.$
Formally, let $\pP^{\ts}=(\jdrT{0,*},\dots,\jdrT{\ts-1,*})$ denote
the past joint policy, which is a partial joint policy $\jpol$ specified
for stages $0,...,\ts-1$. If we denote the optimal past joint policy
by $\pP^{\ts,*}$, we have that $\CoaHistTS{\ts}{\jpol^{*}}=\CoaHistTS{\ts}{\pP^{\ts,*}}$,
and therefore that:

\begin{equation}
\jpolBG^{\ts,*}=\argmax_{\jpolBG^{\ts}}\sum_{\oaHistT{\ts}\in\CoaHistTS{\ts}{\pP^{\ts,*}}}\Pr(\oaHistT{\ts})Q^{*}(\oaHistT{\ts},\jpolBG^{\ts}(\oaHistT{\ts})).\label{eq:forward_BG_Solving:jpol_is_argmax}\end{equation}
 This can be solved in a forward manner for time steps $\ts=0,1,2,...,h-1$,
because at every time step $\pP^{\ts,*}=(\jdrT{0,*},\dots,\jdrT{\ts-1,*})$
will be available: it is specified by $(\jpolBG^{0,*},...,\jpolBG^{\ts-1,*})$
the solutions of the previously solved BGs.

\subsubsection{Computing $Q^{*}$ }

\label{sub:Computing-Q*-nontrivial}

So far we discussed that $Q^{*}$ can be used to find an optimal joint
policy $\jpol^{*}$. Unfortunately, when an optimal joint policy $\jpol^{*}$
is not known, computing $Q^{*}$ itself is impractical, as we will
discuss here. This is in contrast with the (fully observable) single-agent
case where the optimal Q-values can be found relatively easily in
a single  sweep backward through time.

For MDPs and POMDPs we can compute the $Q$-values for time step $\ts$
from those for $\ts+1$ by applying a backup operator. This is possible
because there is a single agent that perceives a Markovian signal.
This allows the agent to (1) select the optimal action (policy) for
the next time step and (2) determine the expected future reward given
the optimal action (policy) found in step 1. For instance, the backup
operator for a POMDP is given by:\[
Q^{*}(b^{\ts},a)=R(b^{\ts},a)+\sum_{o}P(o|b^{\ts},a)\max_{a}Q^{*}(b^{\ts+1},a),\]
 which can be rewritten as a 2-step procedure:

\begin{enumerate}
\item $\jpol^{\ts+1,*}(b^{\ts+1})=\argmax_{a'}Q^{*}(b^{\ts+1},a')$ 
\item $Q^{*}(b^{\ts},a)=R(b^{\ts},a)+\sum_{o}P(o|b^{\ts},a)Q^{*}(b^{\ts+1},\jpol^{\ts+1,*}(b^{\ts+1})).$
\end{enumerate}
In the case of Dec-POMDPs, step 2 would correspond to calculating
$Q^{*}$ using \eqref{eq:Qt(oaHist_ja)} and thus depends on $\jpol^{\ts+1,*}$
an optimal joint policy at the next stage. However, step 1 that calculates
$\jpol^{\ts+1,*}$, corresponds to \eqref{eq:forward_BG_Solving:jpol_is_argmax}
and therefore is dependent on $\pP^{\ts+1,*}$ (an optimal joint policy
for time steps $0,...,\ts$). So to calculate the $Q^{\ts,*}$ the
optimal Q-value function as specified by \eqref{eq:Qt(oaHist_ja)}
for stage $\ts$, an optimal joint policy up to and including stage
$\ts$ is needed. Effectively, there is a dependence on both the future
and the past optimal policy, rather than only on the future optimal
policies as in the single agent case.   The only clear solution
seems to be evaluation for all possible past policies, as detailed
next. We conjecture that the problem encountered here is inherent
to all decentralized decision making with imperfect information. For
example, we can also observe this in exact point-based dynamic programming
for Dec-POMDPs, as described in Section~\ref{sub:DP_extensions},
where it is necessary to to generate all (multiagent belief points
generated by all) possible past policies.

\subsection{Sequential Rationality for Dec-POMDPs}

\label{sub:Sequential-rationality-for}

We conjectured that computing $Q^{*}$ as introduced in Section~\ref{sub:The-optimal-Q-value}
seems impractical without knowing $\jpol^{*}$. Here we will relate
this to concepts from game theory. In particular, we discuss a different
formulation of $Q^{*}$ based on the principle of sequential rationality,
i.e., also considering joint action-observation histories that are
not realized given an optimal joint policy. This formulation of $Q^{*}$
is computable without knowing an optimal joint policy in advance,
and we present a dynamic programming algorithm to perform this computation.

\subsubsection{Sub-game Perfect and Sequential Equilibria}

The problem we are facing is very much related to the notion of \emph{sub-game
perfect equilibria} from game theory. A sub-game perfect Nash equilibrium
$\jpol=\left\langle \polA1,\dots,\polA\nrA\right\rangle $ has the
characteristic that the contained policies $\polA i$ specify an optimal
action for \emph{all} possible situations---even situations that can
not occur when following $\jpol$. A commonly given rationale behind
this concept is that, by a mistake of one of the agents during execution,
situations that should not occur according to $\jpol$, can occur,
and also in these situations the agents should act optimally. A different
rationale is given by \citet{Binmore92}, who remarks that it is ``tempting
to shrug one's shoulders at these difficulties {[}because] rational
players will not stray from the equilibrium path'', but that would
clearly be a mistake, because the agents ``remain on the equilibrium
path because of what they anticipate \emph{would} happen if they \emph{were}
to deviate''. This implies that agents can decide upon a Nash equilibrium
by analyzing what the expected outcome would be by following other
policies: That is, when acting optimally from other situations. We
will perform a similar reasoning here for Dec-POMDPs, which---in a
similar fashion---will result in a description that allows to deduce
an optimal Q-value function and thus joint policy.

A Dec-POMDP can be modeled as an extensive form game of imperfect
information \citep{Oliehoek06_TR_POSGs_extensiveform}. For such games,
the notion of sub-game perfect equilibria is inadequate; because this
type of games often do not contain proper sub-games, every Nash equilibrium
is trivially sub-game perfect.%
\footnote{The extensive form of a Dec-POMDP indeed does not contain proper sub-games,
because agent can never discriminate between the other agents' observations.%
} To overcome this problem different refinements of the Nash equilibrium
concept have been defined, of which we will mention the \emph{assessment
equilibrium} \citep{Binmore92} and the closely related, but stronger
\emph{sequential equilibrium} \citep{OsborneRubinstein94}. Both these
equilibria are based on the concept of an assessment, which is a pair
$\left\langle \jpol,\mathbf{b}\right\rangle $ consisting of a joint
policy $\jpol$ and a \emph{belief system} $\mathbf{b}$. The belief
system maps each possible situation, or information set, of an agent---also
the ones that are not reachable given $\jpol$---to a probability
distribution over possible joint histories. Roughly speaking, an assessment
equilibrium requires \emph{sequential rationality} and \emph{belief
consistency.}%
\footnote{\citet{OsborneRubinstein94} refer to this second requirement as simply
`consistency'. In order to avoid any confusion with definition~\ref{def:consistency}
we will use the term `belief consistency'.%
} The former entails that the joint policy $\jpol$  specifies optimal
actions for each information set given $\mathbf{b}$. Belief consistency
means that all the beliefs that are assigned by $\mathbf{b}$ are
Bayes rational given the specified joint policy~$\jpol$. For instance,
in the context of Dec-POMDPs $\mathbf{b}$ would prescribe, for a
particular $\oaHistAT i\ts$ of agent $i$, a belief over joint histories
$\Pr(\oaHistT\ts|\oaHistAT i\ts)$. If all beliefs prescribed by belief
system~$\mathbf{b}$ are Bayes-rational (i.e., computed as the appropriate
conditionals of \eqref{eq:P(oaHist)}), $\mathbf{b}$ is called belief
consistent.%
\footnote{A sequential equilibrium includes a more technical part in the definition
of belief consistency that addresses what beliefs should be held for
information sets that are not reached according to $\jpol$. For more
information we refer to \citet{OsborneRubinstein94}.%
}

\subsubsection{Sequential Rationality and the Optimal Q-value Function}

The dependence of sequential rationality on a belief system $\mathbf{b}$
indicates that the optimal action at a particular point is dependent
on the probability distribution over histories. In Section \ref{sub:Computing-Q*-nontrivial}
we encountered a similar dependence on the history as specified by
$\pJPolT{\ts+1,*}$. Here we will make this dependence more exact.

At a particular stage $\ts$, a policy is optimal or, in game-theoretic
terms, rational if it maximizes the expected return from that point
on. In Section \ref{sub:Q*}, we were able to express this expected
return as $Q^{*}(\oaHistT\ts,\ja)$ \emph{assuming an optimal joint
policy $\jpol^{*}$ is followed up to the current stage} $\ts$. However,
when no such previous policy is assumed, the maximal expected return
is not defined.

\begin{proposition} For a pair $(\oaHistT{\ts},\ja^{\ts})$ with
$\ts<h-1$ the optimal value $Q^{*}(\oaHistT{\ts},\ja^{\ts})$ cannot
be defined without assuming some (possibly randomized) past policy
$\pJPolT{\ts+1}=\left(\jdrT0,\dots,\jdrT{\ts}\right)$. Only for the
last stage $\ts=h-1$ such expected reward is defined as\[
Q^{*}(\oaHistT{h-1},\ja^{h-1})\defas R(\oaHistT{h-1},\ja^{h-1})\]
without assuming a past policy.

\begin{proof} 
 Let us try to deduce $Q^{*}(\oaHistT{\ts},\ja^{\ts})$ the optimal
value for a particular $\oaHistT\ts$ assuming the $Q^{*}$-values
for the next time step $\ts+1$ are known. The $Q^{*}(\oaHistT{\ts},\ja^{\ts})$-values
for each of the possible joint actions can be evaluated as follows
\[
\forall_{\ja}\quad Q^{*}(\oaHistT{\ts},\ja^{\ts})=R(\oaHistT{\ts},\ja^{\ts})+\sum_{\jo^{\ts+1}}\Pr(\jo^{\ts+1}|\oaHistT{\ts},\ja^{\ts})Q^{*}(\oaHistT{\ts+1},\jdrT{\ts+1,*}(\oaHistT{\ts+1})).\]
where $\jdrT{\ts+1,*}$ is an optimal decision rule for the next stage.
But what should $\jdrT{\ts+1,*}$ be? If we assume that  up to stage
$\ts+1$ we followed a particular (possibly randomized) $\pJPolT{\ts+1}$,
\[
\jdrT{\ts+1,*}_{\pJPolT{}}=\argmax_{\jpolBG^{\ts+1}}\sum_{\oaHistT{\ts+1}\in\CoaHistTS{\ts+1}{}}\Pr(\oaHistT{\ts+1}|\pJPolT{\ts+1},b^{0})Q^{*}(\oaHistT{\ts+1},\jpolBG^{{\ts+1}}(\oaHistT{\ts+1})).\]
is optimal. However, there are many pure and infinite randomized past
policies $\pJPolT{\ts+1}$ that are consistent with $\oaHistT{\ts},\ja^{\ts}$,
leading to many $\jdrT{\ts+1,*}_{\pJPolT{}}$ that might be optimal.
The conclusion we can draw is that  $Q^{*}(\oaHistT{\ts},\ja^{\ts})$
is ill-defined without $\Pr(\oaHistT{\ts+1}|\pJPolT{\ts+1},b^{0})$,
the probability distribution (belief) over joint action-observation
histories, which is induced by $\pJPolT{\ts+1}$, the policy followed
for stages $0,\dots,\ts$. \end{proof}\end{proposition}

Let us illustrate this by reviewing the optimal Q-value function as
defined in Section~\ref{sub:Q*}. Consider $\jpol^{*}(\oaHistT{\ts+1})$
in \eqref{eq:Qt(oaHist_ja)}. This optimal policy is a mapping from
observation histories to actions $\jpol^{*}:\oHistS\rightarrow\mathcal{A}$
induced by the individual policies and observation histories. This
means that for two joint action-observation histories with the same
joint observation history $\jpol^{*}$ results in the same joint action.
That is $\forall\aHist,\oHist,\aHist'\;\jpol^{*}(\left\langle \aHist,\oHist\right\rangle )=\jpol^{*}(\langle\aHist',\oHist\rangle).$
Effectively this means that when we reach some $\oaHistT{\ts}\not\in\oaHistTS\ts_{\jpol^{*}}$
, say through a mistake%
\footnote{The question as to how the mistake of one agent should be detected
by another agent is a different matter altogether and beyond the scope
of this text.%
}, $\jpol^{*}$ continues to specify actions as if no mistake ever
happened: That is, \emph{still} assuming that $\jpol^{*}$ has been
followed up to this stage $\ts$. In fact, $\jpol^{*}(\oaHistT{\ts})$
\emph{might not even be optimal} if $\oaHistT{\ts}\not\in\oaHistTS\ts_{\jpol^{*}}$.
Which in turn means that $Q^{*}(\oaHistT{\ts-1},\ja)$, the Q-values
for predecessors of $\oaHistT{\ts}$, might not be the optimal expected
reward. 

We demonstrated that the optimal Q-value function for a Dec-POMDP
is not well-defined without assuming a past joint policy. We propose
a new definition of $Q^{*}$ that explicitly incorporates $\pJPolT{\ts+1}$. 

\begin{figure}
\begin{centering}
\newcommand{\thissize}{\normalsize}

{
\psfrag{a11}[cc][cb]{\thissize $\aoo$}
\psfrag{a12}[cc][cb]{\thissize $\aot$}
\psfrag{o11}[cc][cb]{\thissize $\ooo$}
\psfrag{o12}[cc][cb]{\thissize $\oot$}

\psfrag{a21}[cc][cb]{\thissize $\ato$}
\psfrag{a22}[cc][cb]{\thissize $\att$}
\psfrag{o21}[cc][cb]{\thissize $\oto$}
\psfrag{o22}[cc][cb]{\thissize $\ott$}

\psfrag{t0}[cc][cc]{\thissize $\ts=0$}
\psfrag{t1}[cc][cc]{\thissize $\ts=1$}
\psfrag{t2}[cc][cc]{\thissize $\ts=2$}
\psfrag{phi1}[cc][cc]{\thissize $\pPolAT{1}{2}$}
\psfrag{phi2}[cc][cc]{\thissize $\pPolAT{2}{2}$}
\psfrag{phi}[cc][cc]{\thissize $\pJPolT{2}$}
\psfrag{dr2}[cc][cc]{\thissize $\jdrT{2,*}_{\pJPolT{}}$}
\psfrag{t}{\thissize `new' component}\includegraphics[width=0.95\columnwidth]{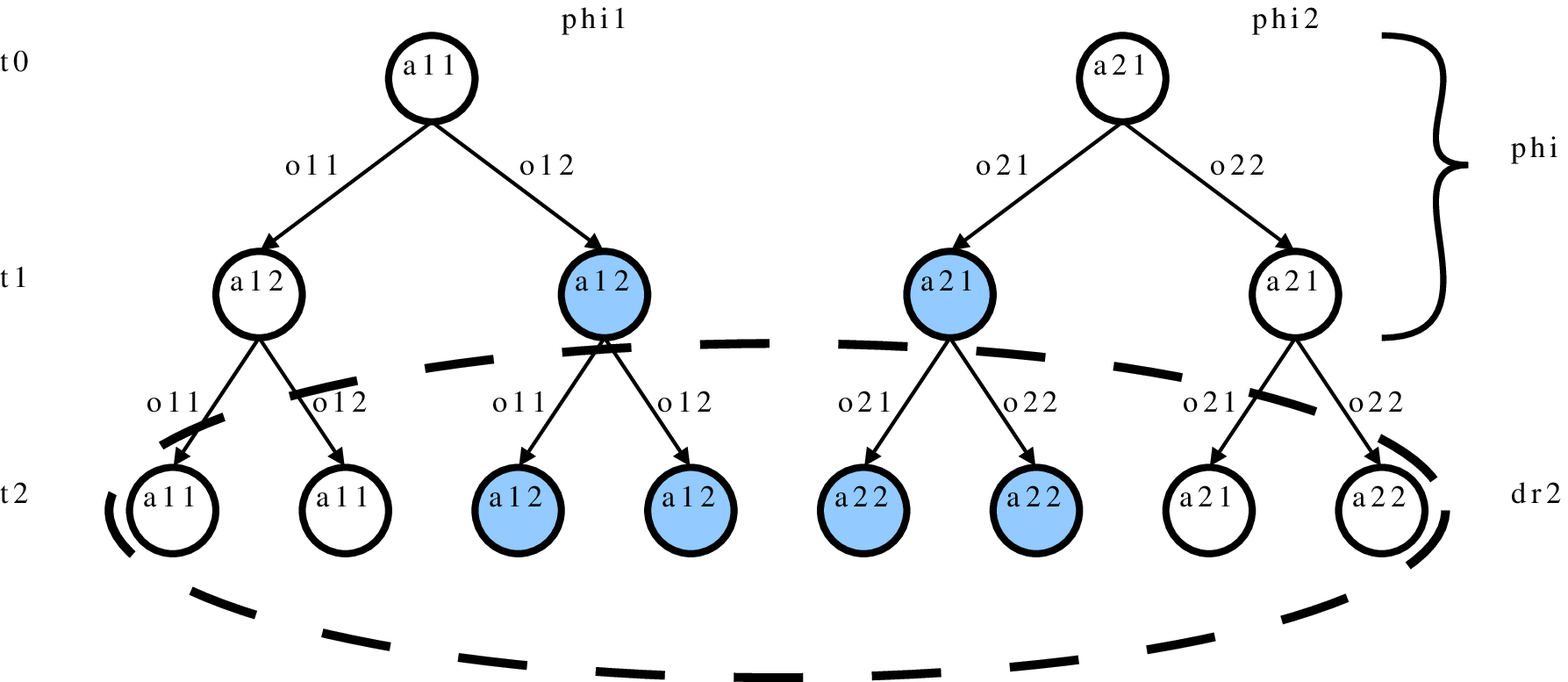}}
\par\end{centering}

\caption{Computation of sequential rational $Q^{*}$. $\jdrT{2,*}_{\pJPolT{}}$
is the optimal decision rule for stage $\ts=2$, given that $\pJPolT2$
is followed for the first two stages. $Q^{*}(\oaHistT{1},\pJPolT{2})$
entries are computed by propagating relevant $Q^{*}$-values of the
next stage. For instance, for the highlighted joint history $\oaHistT{1}=\left\langle (\aoo,\oot),(\ato,\oto)\right\rangle $,
the $Q^{*}$-value under $\pJPolT2$ is computed by propagating the
values of the four successor joint histories, as per \eqref{eq:Q*(oaH,pJPol)}.}

\label{fig:seq-rat-Q}
\end{figure}

\begin{theorem}[Sequentially rational $Q^*$]

The optimal Q-value function is properly defined as a function of
joint action-observation histories and past joint policies, $Q^{*}(\oaHistT\ts,\pJPolT{\ts+1})$.
This $Q^{*}$ specifies the optimal value given\emph{ }for all $(\oaHistT\ts,\pJPolT{\ts+1})$,
even for $\oaHistT\ts$ that are not reached by execution of an optimal
policy $\jpol^{*}$, and therefore is referred to as sequentially
rational.

\begin{proof}For all $\oaHistT{\ts},\pJPolT{\ts+1}$, the optimal
expected return is given by\begin{equation}
Q^{*}(\oaHistT{\ts},\pJPolT{\ts+1})=\begin{cases}
{\displaystyle R(\oaHistT{\ts},\pJPolT{\ts+1}(\oaHistT{\ts})),} & \ts=h-1\\
{\displaystyle R(\oaHistT{\ts},\pJPolT{\ts+1}(\oaHistT{\ts}))+\sum_{\jo^{\ts+1}}\Pr(\jo^{\ts+1}|\oaHistT{\ts},\pJPolT{\ts+1}(\oaHistT{\ts}))Q^{*}(\oaHistT{\ts+1},\pJPolT{\ts+2,*}),} & 0\leq\ts<h-1\end{cases}\label{eq:Q*(oaH,pJPol)}\end{equation}
where $\pJPolT{\ts+2,*}=\left(\pJPolT{\ts+1},\jdrT{\ts+1,*}_{\pJPolT{}}\right)$
and\begin{equation}
\jdrT{\ts+1,*}_{\pJPolT{}}=\argmax_{\jpolBG^{\ts+1}}\sum_{\oaHistT{\ts+1}\in\CoaHistTS{\ts+1}{}}\Pr(\oaHistT{\ts+1}|\pJPolT{\ts+1},b^{0})Q^{*}(\oaHistT{\ts+1},\left(\pJPolT{\ts+1},\jpolBG^{{\ts+1}}\right)).\label{eq:jdr*_given_pJPol}\end{equation}
which is well-defined.\end{proof}\end{theorem} 

The above equations constitute a dynamic program. When assuming that
only pure joint past policies $\pJPol$ can be used, \eqref{eq:jdr*_given_pJPol}
transforms to \begin{equation}
\jdrT{\ts+1,*}_{\pJPolT{}}=\argmax_{\jpolBG^{\ts+1}}\sum_{\oaHistT{\ts+1}\in\CoaHistTS{\ts+1}{\pJPol}}\Pr(\oaHistT{\ts+1})Q^{*}(\oaHistT{\ts+1},\left(\pJPolT{\ts+1},\jpolBG^{{\ts+1}}\right))\label{eq:jdr*_given_pure-pJPol}\end{equation}
and for all $(\oaHist,\pJPol)$ such that $\oaHist$ is consistent
with $\pJPol$ the dynamic program can be evaluated from the end ($\ts=h-1$)
to the begin $(\ts=0$). \fig\ref{fig:seq-rat-Q} illustrates the
computation of $Q^{*}$. When arriving at stage $0$, the $\pJPolT1$
reduce to joint actions and it is possible to select\[
\jdrT{0,*}=\argmax_{\ja}Q^{*}(\oaHistEmpty,\ja)=\argmax_{\pJPolT1}Q^{*}(\oaHistT0,\pJPolT1).\]
Then given $\pJPolT{1}=\jdrT{0,*}$ we can determine $\jdrT{1,*}=\jdrT{1,*}_{\pJPolT{}}$
using \eqref{eq:jdr*_given_pure-pJPol}, etc. This essentially is
the forward-sweep policy computation using the optimal Q-value function
$Q^{*}(\oaHistT{\ts},\pJPolT{\ts+1})$ as defined by \eqref{eq:Q*(oaH,pJPol)}. 

The computation of $Q^{*}$ is also closely related to point-based
dynamic programming for Dec-POMDPs as discussed in section \ref{sub:DP_extensions}.
Suppose that $\ts=2$ in \fig\ref{fig:seq-rat-Q} is the last stage
(i.e., $h=3$). When the $\jdrT{2,*}$ for all $\pJPolT2$ have been
computed, it is easy to construct the sets of non-dominated action
for each agent: every action $\aA i$ of agent $i$ that is specified
by some~$\drAT i{2,*}$ is non-dominated. Once we have computed the
values for all $(\oaHistT1,\pJPolT2)$ at $\ts=1$, each $\pJPolT2$
has an associated optimal future policy~$\jdrT{2,*}_{\pJPol}$. This
means that each individual history~$\oaHistAT i1$ has an associated
sub-tree policy $\stPolAT i2$ for each $\pJPolT2$ and as such each
$(\oaHistT1,\pJPolT2)$-pair has an associated joint sub-tree policy
(e.g, the shaded trees in \fig\ref{fig:seq-rat-Q}). Clearly, $Q^{*}(\oaHistT1,\pJPolT2)$
corresponds to expected value of this associated joint sub-tree policy.
Rather than keeping track of these sub-trees policies, however, the
algorithm presented here keeps track of the values.

The advantage of the description of $Q^{*}$ using \eqref{eq:jdr*_given_pJPol}
rather than \eqref{eq:Qt(oaHist_ja)} is twofold. First the description
treated here describes the way to actually compute the values which
can then be used to construct $\jpol^{*}$, while the latter only
gives a normative description and needs $\jpol^{*}$ in order to compute
the Q-values. 

Second, this $Q^{*}(\oaHistT{\ts},\pJPolT{\ts+1})$ describes sequential
rationality for Dec-POMDPs. For \emph{any} past policy (and corresponding
consistent belief system) the optimal future policy can be computed.
A variation of this might even be applied on-line. Suppose agent $i$
makes a mistake at stage $\ts$, executing an action not prescribed
by $\polA i^{*}$, assuming the other agents execute their policy
$\polA{\neq i}$ without mistakes, agent $i$ knows the actually executed
previous policy $\pJPolT{\ts+1}$. Therefore it can compute a new
individual policy by\[
\drAT{i,\pJPolT{\ts+1}}{\ts+1,*}=\argmax_{{\polBG i}^{\ts+1}}\sum_{\oaHistT{\ts+1}\in\CoaHistTS{\ts+1}{\pJPol}}\Pr(\oaHistT{\ts+1})Q^{*}(\oaHistT{\ts+1},\left(\pJPolT{\ts+1},\left\langle {\polBG i}^{{\ts+1}},\drAT{\neq i}{\ts+1}\right\rangle \right)).\]

\subsubsection{The Complexity of Computing a Sequentially Rational $Q^{*}$}

Although we have now found a way to compute $Q^{*}$, this computation
is intractable for all but the smallest problems, as we will now show.
At stage $\ts-1$ there are $\sum_{\ts'=0}^{\ts-1}{|\oAS i|}^{\ts'}=\frac{\left|\oAS i\right|^{\ts}-1}{\left|\oAS i\right|-1}$
observation histories for agent $i$, leading to \[
\left|\aAS*\right|^{\frac{\nrA\left(\left|\oAS*\right|^{\ts}-1\right)}{\left|\oAS*\right|-1}}\]
pure joint past policies $\pJPolT{\ts}$. For each of these there
are $|\oHistTS{\ts}|={|\mathcal{O}|}^{\ts-1}$ consistent joint action-observation
histories (for each observation history $\oHistT{\ts-1}$, $\pJPolT{\ts}$
specifies the actions forming $\oaHistT{\ts-1}$). This means that
for stage $h-2$ (for $h-1$, the Q-values are easily calculated),
the number of entries to be computed is the number of joint past policies
$\pJPolT{h-1}$ times the number of joint histories \[
O\left(\left|\aAS*\right|^{\frac{\nrA\left(\left|\oAS*\right|^{h-1}-1\right)}{\left|\oAS*\right|-1}}\cdot{|\mathcal{O}|}^{h-2}\right),\]
indicating that computation of this function is doubly exponential,
just as brute force policy evaluation. Also, for each joint past policy
$\pJPolT{h-1}$, we need to compute $\pJPolT{h}=(\pJPolT{h-1},\jdrT{h-1,*}_{\pJPolT{}})$
by solving the next-stage BG: \[
\jdrT{h-1,*}_{\pJPolT{}}=\argmax_{\jpolBG^{h-1}}\sum_{\oaHistT{h-1}\in\CoaHistTS{h-1}{\pJPol}}\Pr(\oaHistT{h-1})Q^{*}(\oaHistT{h-1},\left(\pJPolT{h-1},\jpolBG^{{h-1}}\right)).\]
To the authors' knowledge, the only method to optimally solve these
BGs is evaluation of all 
\[O\left(\left|\aAS*\right|^{\nrA\left|\oAS*\right|^{h-1}}\right)\]
joint BG-polices, which is also doubly exponential in the horizon.

\section{Approximate Q-value Functions}

\label{sec:Approximate-Q-value-functions}

As indicated in the previous section, although an optimal Q-value
function $Q^{*}$ exists, it is costly to compute and thus impractical.
In this section, we review some other Q-value functions, $\QH$, that
can be used as an approximation for $Q^{*}$. We will discuss underlying
assumptions, computation, computational complexity and other properties,
thereby providing a taxonomy of approximate Q-value functions for
Dec-POMDPs. In particular we will treat two well-known approximate
Q-value functions, $\QMDP$ and $\QPOMDP$, and $\QBG$ recently introduced
by \citet{Oliehoek07aamas}.

\subsection{$\QMDP$}

\label{sub:QMDP}

$\QMDP$ was originally proposed to approximately solve POMDPs by
\citet{Littman95ScalingUp}, but has also been applied to Dec-POMDPs
\citep{Emery-Montemerlo04,Szer05MAA}. The idea is that $Q^{*}$ can
be approximated using the state-action values $\QMDPQ(s,\ja)$ found
when solving the `underlying MDP' of a Dec-POMDP. This `underlying
MDP' is the horizon-$h$ MDP defined by a single agent that takes
joint actions $\ja\in\mathcal{A}$ and observes the nominal state
$s$ that has the same transition model $T$ and reward model $R$
as the original Dec-POMDP. Solving this underlying MDP can be efficiently
done using dynamic programming techniques \citep{Puterman94}, resulting
in the optimal non-stationary MDP Q-value function:\begin{equation}
\QMDPQ^{\ts,*}(s^{\ts},\ja)=R(s^{\ts},\ja)+\sum_{s^{\ts+1}\in\mathcal{S}}\Pr(s^{\ts+1}|s^{\ts},\ja)\max_{\ja}\QMDPQ^{\ts+1,*}(s^{\ts+1},\ja).\label{eq:QMDP}\end{equation}
In this equation, the maximization is an implicit selection of $\jpolMDP^{\ts+1,*},$
the optimal MDP policy at the next time step, as explained in Section
\ref{sub:Computing-Q*-nontrivial}. Note that $\QMDPQ^{\ts,*}$ also
is an optimal Q-value function, but in the MDP setting. In this article
$Q^{*}$ will always denote the optimal value function for the (original)
Dec-POMDP. In order to transform the $\QMDPQ^{\ts,*}(s^{\ts},\ja)$-values
to approximate $\QMDPQH(\oaHistT\ts,\ja)$-values to be used the original
Dec-POMDP, we compute:

\begin{equation}
\QMDPQH(\oaHistT\ts,\ja)=\sum_{s\in\mathcal{S}}\QMDPQ^{\ts,*}(s,\ja)\Pr(s|\oaHistT\ts),\label{eq:QMDPH_1}\end{equation}
 where $\Pr(s|\oaHistT\ts)$ can be computed from \eqref{eq:P(s,oaHist)}.
Combining \eqref{eq:QMDP} and \eqref{eq:QMDPH_1} and making the
selection of $\jpolMDP^{\ts+1,*}$ explicit we get:\begin{equation}
\QMDPQH(\oaHistT\ts,\ja)=R(\oaHistT\ts,\ja)+\sum_{s^{\ts+1}\in\mathcal{S}}\Pr(s^{\ts+1}|\oaHistT\ts,\ja)\max_{\jpolMDP^{\ts+1}(s^{\ts+1})}\QMDPQ^{\ts+1,*}(s^{\ts+1},\jpolMDP^{\ts+1}(s^{\ts+1})),\label{eq:QMDPQH_2}\end{equation}
which defines the approximate Q-value function that can be used as
payoff function for the various BGs of the Dec-POMDP. Note that $\QMDPQH$
is consistent with the established definition of Q-value functions
since it is defined as the expected immediate reward of performing
(joint) action $\ja$ plus the value of following an optimal joint
policy (in this case the optimal MDP-policy) thereafter.

Because calculation of the $\QMDPQ^{\ts}(s,\ja)$-values by dynamic
programming (which has a cost of $O(|\mathcal{S}|\times h)$ can be
performed in a separate phase, the cost of computation of $\QMDP$
is only dependent on the cost of evaluation of \eqref{eq:QMDPQH_2},
which is $O(|\mathcal{S}|)$. When we want to evaluate $\QMDP$ for
all $\sum_{\ts=0}^{h-1}\left(\left|\mathcal{A}\right|\left|\mathcal{O}\right|\right)^{\ts}=\frac{\left(\left|\mathcal{A}\right|\left|\mathcal{O}\right|\right)^{h}-1}{\left(\left|\mathcal{A}\right|\left|\mathcal{O}\right|\right)-1}$
joint action-observation histories is, the total computational cost
becomes:\begin{equation}
O\left(\frac{\left(\left|\mathcal{A}\right|\left|\mathcal{O}\right|\right)^{h}-1}{\left(\left|\mathcal{A}\right|\left|\mathcal{O}\right|\right)-1}|\mathcal{A}||\mathcal{S}|\right).\label{eq:complex:QMPD_complete}\end{equation}
However, when applying $\QMDP$ in forward-sweep policy computation,
we do not have to consider \emph{all} action-observation histories,
but only those that are consistent with the policy found for earlier
stages. Effectively we only have to evaluate \eqref{eq:QMDPQH_2}
for all observation histories and joint actions, leading to: \begin{equation}
O\left(\frac{\left(\left|\mathcal{O}\right|\right)^{h}-1}{\left(\left|\mathcal{O}\right|\right)-1}|\mathcal{A}||\mathcal{S}|\right).\label{eq:complex:QMPD_FSPC}\end{equation}

When used in the context of Dec-POMDPs, $\QMDP$ solutions are known
to undervalue actions that gain information \citep{Fernandez06}.
This is explained by realizing that the $\QMDP$ solution assumes
that the state will be fully observable in the next time step. Therefore
actions that provide information about the state, and thus can lead
to a high future reward (but might have a low immediate reward), will
be undervalued. When applying $\QMDP$ in the Dec-POMDP setting, this
effect can also be expected. Another consequence of the simplifying
assumption is that the $\QMDP$-value function is an upper bound to
the optimal value function when used to approximate a POMDP \citep{Hauskrecht00},
as a consequence it is also an upper bound to the optimal value function
of a Dec-POMDP. This is intuitively clear, as a Dec-POMDP is a POMDP
but with the additional difficulty of decentralization. A formal argument
will be presented in Section~\ref{sub:Generalized-QBG-and_bounds}.

\subsection{$\QPOMDP$}

\label{sub:QPOMDP}

Similar to the `underlying MDP', one can define the `underlying POMDP'
of a Dec-POMDP as the POMDP with the same $T$, $O$ and $R$, but
in which there is only a single agent that takes joint actions $\ja\in\mathcal{A}$
and receives joint observations $\jo\in\mathcal{O}$. $\QPOMDP$ approximates
$Q^{*}$ using the solution of the underlying POMDP \citep{Szer05MAA,Roth05AAMAS}. 

In particular, the optimal $\QPOMDP$ value function for an underlying
POMDP satisfies:

\begin{equation}
\QPOMDPQ^{*}(b^{\oaHistT{\ts}},\ja)=R(b^{\oaHistT{\ts}},\ja)+\sum_{\jo^{\ts+1}\in\mathcal{O}}P(\jo^{\ts+1}|b^{\oaHistT{\ts}},\ja)\max_{\jpolPOMDP^{\ts+1}(b^{\oaHistT{\ts+1}})}\QPOMDPQ^{*}(b^{\oaHistT{\ts+1}},\jpolPOMDP^{\ts+1}(b^{\oaHistT{\ts+1}})),\label{eq:QPOMDP}\end{equation}
where $b^{\oaHistT\ts}$ is the \emph{joint belief} of the single
agent that selects joint actions and receives joint observations at
time step $\ts$, where\begin{equation}
R(b^{\oaHistT{\ts}},\ja)=\sum_{s\in\mathcal{S}}R(s,\ja)b^{\oaHistT{\ts}}(s)\label{eq:R(b,a)}\end{equation}
is the immediate reward, and where $b^{\oaHistT{\ts+1}}$ is the joint
belief resulting from $b^{\oaHistT\ts}$ by action $\ja$ and joint
observation $\jo^{\ts+1}$, calculated by\begin{equation}
\forall_{s'}\quad b^{\oaHistT{\ts+1}}(s')=\frac{\Pr(\jo|\ja,s')\sum_{s\in\mathcal{S}}\Pr(s'|s,\ja)b^{\oaHistT{\ts}}(s)}{\sum_{s'\in\mathcal{S}}\Pr(\jo|\ja,s')\sum_{s\in\mathcal{S}}\Pr(s'|s,\ja)b^{\oaHistT{\ts}}(s)}.\label{eq:beliefupdate}\end{equation}
For each $\oaHistT{\ts}$ there is one joint belief $b^{\oaHistT{\ts}}$,
which corresponds to $\Pr(s|\oaHistT\ts)$ as can be derived from
\eqref{eq:P(s,oaHist)}. Therefore it is possible to directly use
the computed $\QPOMDP$ values as payoffs for the BGs of the Dec-POMDP,
that is, we define: \begin{equation}
\QPOMDPQH(\oaHistT{\ts},\ja)\defas\QPOMDPQ^{*}(b^{\oaHistT{\ts}},\ja).\label{eq:QPOMDPQH}\end{equation}

The maximization in \eqref{eq:QPOMDP} is stated in its explicit form:
a maximization over time step $\ts+1$ POMDP policies. However, it
should be clear that this maximization effectively is one over joint
actions, as it is conditional on the received joint observation $\jo^{\ts+1}$
and thus the resulting belief $b^{\oaHistT{\ts+1}}$.

For a finite horizon, $\QPOMDPQ^{*}$ can be computed by generating
all possible joint beliefs and solving the `belief MDP'. Generating
all possible beliefs is easy: starting with $b^{0}$ corresponding
to the empty joint action-observation history $\oaHistT{\ts=0}$,
for each $\ja$ and $\jo$ we calculate the resulting $\oaHistT{\ts=1}$
and corresponding belief $b^{\oaHistT1}$ and continue recursively.
Solving the belief MDP amounts to recursively applying \eqref{eq:QPOMDP}.

In the computation of $\QMDP$ we could restrict our attention to
only those $(\oaHistT\ts,\ja)$-pairs that were specified by forward-sweep
policy computation, because the $\QMDPQH(\oaHistT\ts,\ja)$-values
do not depend on the values of successor-histories $\QMDPQH(\oaHistT{\ts+1},\ja)$.
For $\QPOMDP$, however, there is such a dependence, meaning that
it is necessary to evaluate for all $\oaHistT\ts,\ja$. In particular,
the cost of calculating $\QPOMDP$ can be divided in the cost of calculating
the expected immediate reward for all $\oaHistT\ts,\ja$, and the
cost of evaluating future reward for all $\oaHistT\ts,\ja$, with
$\ts=0,...,h-2$. The former operation is given by \eqref{eq:R(b,a)}
and has cost $O(\left|\mathcal{S}\right|)$ per $\oaHistT\ts,\ja$
 and thus a total cost equal to \eqref{eq:complex:QMPD_complete}.
The latter requires selecting the maximizing joint action for each
joint observation for all $\oaHistT\ts,\ja$ with $\ts=0,...,h-2$,
leading to\begin{equation}
O\left(\frac{\left(\left|\mathcal{A}\right|\left|\mathcal{O}\right|\right)^{h-1}-1}{\left(\left|\mathcal{A}\right|\left|\mathcal{O}\right|\right)-1}|\mathcal{A}|\left(\left|\mathcal{A}\right|\left|\mathcal{O}\right|\right)\right).\label{eq:QPOMDPcomplex_fut_reward}\end{equation}
Therefore the total complexity of computing $\QPOMDP$ becomes\begin{equation}
O\left(\frac{\left(\left|\mathcal{A}\right|\left|\mathcal{O}\right|\right)^{h-1}-1}{\left(\left|\mathcal{A}\right|\left|\mathcal{O}\right|\right)-1}|\mathcal{A}|\left(\left|\mathcal{A}\right|\left|\mathcal{O}\right|\right)+\frac{\left(\left|\mathcal{A}\right|\left|\mathcal{O}\right|\right)^{h}-1}{\left(\left|\mathcal{A}\right|\left|\mathcal{O}\right|\right)-1}|\mathcal{A}||\mathcal{S}|\right).\label{eq:comp_QPOMDP}\end{equation}

\begin{figure}
\begin{centering}
{ 
\footnotesize

\noindent {

\renewcommand{\multirowsetup}{\centering}
\setlength\arrayrulewidth{\thickertableline}\arrayrulecolor{black}

\begin{tabular}{cc|cc}
&
$\oaHistAT2{\ts=0}$&
\multicolumn{2}{c}{$\left(\right)$}\tabularnewline
$\oaHistAT1{\ts=0}$&
&
$\ato$&
$\att$\tabularnewline
\hline
\multirow{2}{0.3cm}{ $\left(\right)$ }&
$\aoo$&
\cellcolor{tablemid} $+3.1$&
$-4.1$\tabularnewline
&
$\aot$&
$-0.9$&
$+0.3$\tabularnewline
\end{tabular}~~~\begin{tabular}{cc|cc|cc|c}
&
$\oaHistAT2{\ts=1}$&
\multicolumn{2}{c|}{$\left(\ato,\oto\right)$}&
\multicolumn{2}{c|}{$\left(\ato,\ott\right)$}&
...\tabularnewline
$\oaHistAT1{\ts=1}$&
&
$\ato$&
$\att$&
$\ato$&
$\att$&
\tabularnewline
\hline
\multirow{2}{0.9cm}{$\left(\aoo,\ooo\right)$}&
$\aoo$&
$-0.3$&
$+0.6$&
$-0.6$&
\cellcolor{tablemid} $+4.0$&
\cellcolor{tabledark} ...\tabularnewline
&
$\aot$&
$-0.6$&
\cellcolor{tablemid} $+2.0$&
$-1.3$&
$+3.6$&
\cellcolor{tabledark} ...\tabularnewline
\hline
\multirow{2}{0.9cm}{$\left(\aoo,\oot\right)$}&
$\aoo$&
$+3.1$&
\cellcolor{tablemid} $+4.4$&
$-1.9$&
$+1.0$&
\cellcolor{tabledark} ...\tabularnewline
&
$\aot$&
$+1.1$&
$-2.9$&
\cellcolor{tablemid} $+2.0$&
$-0.4$&
\cellcolor{tabledark} \tabularnewline
\hline
\multirow{2}{0.9cm}{$\left(\aot,\ooo\right)$}&
$\aoo$&
\cellcolor{tabledark} $-0.4$&
\cellcolor{tabledark} $-0.9$&
\cellcolor{tabledark} $-0.5$&
\cellcolor{tabledark} $-1.0$&
\cellcolor{tabledark} ...\tabularnewline
&
$\aot$&
\cellcolor{tabledark} $-0.9$&
\cellcolor{tabledark} $-4.5$&
\cellcolor{tabledark} $-1.0$&
\cellcolor{tabledark} $+3.5$&
\cellcolor{tabledark} ...\tabularnewline
\hline
\multirow{1}{0.9cm}{$\left(\aot,\oot\right)$}&
...&
\cellcolor{tabledark} ...&
\cellcolor{tabledark} ...&
\cellcolor{tabledark} ...&
\cellcolor{tabledark} ...&
\cellcolor{tabledark} ...\tabularnewline
\end{tabular}

}

}
\par\end{centering}

\caption{Backward calculation of $\QPOMDP$-values. Note that the solutions
(the highlighted entries) are different from those in Figure \ref{tab:Bayesian-game}:
$\QPOMDP$ assumes that the actions can be conditioned on the joint
action-observation history. The highlighted `$+3.1$' entry for the
Bayesian game for $\ts=0$ is calculated as the expected immediate
reward $(=0)$ plus a weighted sum of the maximizing entry (joint
action) per next joint observation history. When assuming a uniform
distribution over joint observations given $\left\langle \aoo,\ato\right\rangle $
the future reward is given by: $+3.1=0+0.25\times2.0+0.25\times4.0+0.25\times4.4+0.25\times2.0$. }

\label{tab:Bayesian-game_QPOMDP}
\end{figure}

Evaluating \eqref{eq:QPOMDP} for all joint action-observation histories
$\oaHistT\ts\in\oaHistTS\ts$ can be done in a single backward sweep
through time, as we mentioned in Section \ref{sub:Computing-Q*-nontrivial}.
This can also be visualized in Bayesian games as illustrated in Figure
\ref{tab:Bayesian-game_QPOMDP}; the expected future reward is calculated
as a maximizing weighted sum of the entries of the next time step
BG. 

Nevertheless, solving a POMDP optimally is also known as an intractable
problem. As a result, POMDP research in the last decade has focused
on approximate solutions for POMDPs. In particular, it is known that
the value function of a POMDP is \emph{piecewise-linear and convex
(PWLC)} over the (joint) belief space \citep{Sondik71}. This property
is exploited by many approximate POMDP solution methods \citep{Pineau03PBVI,Spaan05jair}.
Clearly such methods can also be used to calculate an approximate
$\QPOMDP$-value function for use with Dec-POMDPs.

It is intuitively clear that $\QPOMDP$ is also an admissible heuristic
for Dec-POMDPs, as it still assumes that more information is available
than actually is the case (again a formal proof will be given in  Section
\ref{sub:Generalized-QBG-and_bounds}). Also it should be clear that,
as fewer assumptions are made, $\QPOMDP$ should yield less of an
over-estimation than $\QMDP.$ I.e., the $\QPOMDP$-values should
lie between the $\QMDP$ and optimal $Q^{*}$-values.

In contrast to $\QMDP,$ $\QPOMDP$ does not assume full observability
of nominal states. As a result the latter does not share the drawback
of undervaluing actions that will gain information regarding the nominal
state. When applied in a Dec-POMDP setting, however, $\QPOMDP$ does
share the assumption of centralized control. This assumption might
also cause a relative undervaluation: there might be situations where
some action might gain information regarding the joint (i.e., each
other's) observation history. Under $\QPOMDP$ this will be considered
redundant, while in decentralized execution this might be very beneficial,
as it allows for better coordination.

\subsection{$\QBG$}

\label{sub:QBG}

$\QMDP$ approximates $Q^{*}$ by assuming that the state becomes
fully observable in the next time step, while $\QPOMDP$ assumes that
at every time step $\ts$ the agents know the joint action-observation
history $\oaHistT\ts$. Here we present a new approximate Q-value
function, called $\QBG$, that relaxes the assumptions further: it
assumes that the agents know $\oaHistT{\ts-1}$, the joint action-observation
history up to time step $\ts-1$, and the joint action $\ja^{\ts-1}$
that was taken at the previous time step. This means that the agents
are uncertain regarding each other's last observation, which effectively
defines a BG for each $\oaHistT{\ts-1},\ja$. Note, that these BGs
are different from the BGs used in Section \ref{sub:Modeling-Dec-POMDPs-with-BGs}:
the BGs here have types that correspond to single observations, whereas
the BGs in \ref{sub:Modeling-Dec-POMDPs-with-BGs} have types that
correspond to complete action-observation histories. Hence, the BGs
of $\QBG$ are much smaller in size and thus easier to solve. Formally
$\QBG$ is defined as:

\begin{equation}
\QBGQ^{*}(\oaHistT{\ts},\ja)=R(\oaHistT{\ts},\ja)+\max_{\jpolBG}\sum_{\jo^{\ts+1}\in\mathcal{O}}\Pr(\jo^{\ts+1}|\oaHistT{\ts},\ja)\QBGQ^{*}(\oaHistT{\ts+1},\jpolBG(\jo^{\ts+1})),\label{eq:QBG}\end{equation}
 where $\jpolBG=\langle\polBG{1}(\oAT1{\ts+1}),...,\polBG{\nrA}(\oAT\nrA{\ts+1})\rangle$
is a tuple of individual policies $\polBG{i}:\oAS{i}\rightarrow\aAS i$
for the BG constructed for $\oaHistT{\ts},\ja$.

Note that the only difference between \eqref{eq:QBG} and \eqref{eq:QPOMDP}
is the position and argument of the maximization operator: \eqref{eq:QBG}
maximizes over a (conditional) BG-policy, while the maximization in
\eqref{eq:QPOMDP} is effectively over unconditional joint actions.

The BG representation of the fictitious Dec-POMDP in \fig  \ref{tab:Bayesian-game}
illustrates the computation of $\QBG$.%
\footnote{Because the BG representing $\ts=1$ of a Dec-POMDP also involves
observation histories of length 1, the illustration of such a BG corresponds
to the BGs as considered in $\QBG$. For other stages this is not
the case.%
} The probability distribution $\Pr(\CoaHistTS1{\left\langle \aoo,\ato\right\rangle })$
over joint action-observation histories that can be reached given
$\left\langle \aoo,\ato\right\rangle $ at $\ts=0$ is uniform and
the immediate reward for $\left\langle \aoo,\ato\right\rangle $ is
0. Therefore, we have that $2.75=0.25\cdot2.0+0.25\cdot3.6+0.25\cdot4.4+0.25\cdot1.0$. 

The cost of computing $\QBG$ for all $\oaHistT\ts,\ja$ can be split
up in the cost of computing the immediate reward (see \eqref{eq:complex:QMPD_complete})
and the cost of computing the future reward (solving a BG over the
last received observation), which is\[
O\left(\frac{\left(\left|\mathcal{A}\right|\left|\mathcal{O}\right|\right)^{h-1}-1}{\left(\left|\mathcal{A}\right|\left|\mathcal{O}\right|\right)-1}|\mathcal{A}|\cdot\left|\mathcal{A}_{*}\right|^{\nrA\left|\mathcal{O}_{*}\right|}\right),\]
leading to a total cost of: \textbf{\begin{equation}
O\left(\frac{\left(\left|\mathcal{A}\right|\left|\mathcal{O}\right|\right)^{h-1}-1}{\left(\left|\mathcal{A}\right|\left|\mathcal{O}\right|\right)-1}|\mathcal{A}|\cdot\left|\mathcal{A}_{*}\right|^{\nrA\left|\mathcal{O}_{*}\right|}+\frac{\left(\left|\mathcal{A}\right|\left|\mathcal{O}\right|\right)^{h}-1}{\left(\left|\mathcal{A}\right|\left|\mathcal{O}\right|\right)-1}|\mathcal{A}||\mathcal{S}|\right).\label{eq:comp_QBG}\end{equation}
}Comparing to the cost of computing $\QPOMDP$, this contains an additional
exponential term, but this term does not depend on the horizon of
the problem.

As mentioned in Section \ref{sub:QPOMDP}, $\QPOMDP$ can be approximated
by exploiting the PWLC-property of the value function. It turns out
that the $\QBG$-value function corresponds to an optimal value function
for the situation where the agents can communicate freely with a one-step
delay \citep{Oliehoek07msdm}. \citet{Hsu82} showed how a complex
dynamic program can be constructed for such settings and that the
resulting value function also preserves the PWLC property. Not surprisingly,
the $\QBG$-value function also is piecewise-linear and convex over
the joint belief space and, as a result, approximation methods for
POMDPs can be transferred to the computation of $\QBG$ \citep{Oliehoek07msdm}.

\subsection{Generalized $\QBG$ and Bounds}

\label{sub:Generalized-QBG-and_bounds}

We can think of an extension of the $\QBG$-value function framework
to the case of $k$-steps delayed communication, where each agent
perceives the joint action-observation history with $k$ stages delay.
That is, at stage $\ts$, each agent $i$ knows $\oaHistT{\ts-k}$
the joint action-observation history of $k$ stages before in addition
to its own current action-observation history $\oaHistAT i{\ts}$.
Similar $k$-step delayed observation models for decentralized control
have been previously proposed by \citet{Aicardi87} and \citet{Ooi96}.
In particular \citeauthor{Aicardi87} consider the Dec-MDP setting in which
agent $i$'s observations are local states~$s_{i}$ and where a joint
observation identifies the state $s=\left\langle s_{1},\dots,s_{\nrA}\right\rangle $.
\citeauthor{Ooi96} examine the decentralized control of a broadcast channel
over an infinite horizon, where they allow the local observations
to be arbitrary, but still require the joint state to be observed
with a $k$-steps delay. Our assumption is less strong, as we only
require observation of $\oaHistT{\ts-k}$ and because we assume the
general Dec-POMDP (not Dec-MDP) setting.

Such a $k$-step delayed communication model for the Dec-POMDP setting
allows expressing the different Q-value functions defined in this
article as optimal value functions of appropriate $k$-step delay
models. More importantly, by resorting to such a $k$-step delay model
we can prove a hierarchy of bounds that hold over the various Q-functions
defined in this article:

\begin{theorem}[Hierarchy of upper bounds] \label{thm:upperbounds} 

The approximate Q-value functions $\QBG$ and $\QPOMDP$ correspond
to the optimal Q-value functions of appropriately defined $k$-step
delayed communication models. Moreover these Q-value functions form
a hierarchy of upper bounds to the optimal $Q^{*}$ of the Dec-POMDP:\begin{equation}
Q^{*}\leq\QBG\leq\QPOMDP\leq\QMDP.\label{eq:Q-hierarchy}\end{equation}

\proofup   See appendix. \end{proof} \end{theorem}

The idea is that a POMDP corresponds to a system with no ($0$-steps)
delayed communication, while the $\QBG$-setting corresponds to a
$1$-step delayed communication system. The appendix shows that the
Q-value function of a system with $k$ steps delay forms an upper
bound to that of a decentralized system with $k+1$ steps delay. We
note that the last inequality of \eqref{eq:Q-hierarchy} is a well-known
result \citep{Hauskrecht00}.

\section{Generalized Value-Based Policy Search}

\label{sec:generaliz-value-based-policy-search}

The hierarchy of approximate Q-value functions implies that all of
these Q-value functions can be used as \emph{admissible heuristics}
in $\MAA$ policy search, treated in Section \ref{sub:MAA*}. In this
section we will present a more general heuristic policy search framework
which we will call Generalized $\MAA$ ($\GMAA$), and show how it
unifies some of the solution methods proposed for Dec-POMDPs. 

$\GMAA$ generalizes $\MAA$ \citep{Szer05MAA} by making explicit
different procedures that are implicit in $\MAA$: (1) iterating over
a pool of partial joint policies, pruning this pool whenever possible,
(2) selecting a partial joint policy from the policy pool, and (3)
finding some new partial and/or full joint policies given the selected
policy. The first procedure is the core of $\GMAA$ and is fixed,
while the other two procedures can be performed in many ways. 

The second procedure, $\selectO$, chooses which policy to process
next and thus determines the type of search (e.g., depth-first, breadth-first,
A{*}-like) \citep{RussellNorvig2003,Bersekas05_vol1}. The third procedure,
which we will refer to as $\PSO$, determines how the set of next
(partial) joint policies are constructed, given a previous partial
joint policy. The original $\MAA$ can be seen as an instance of the
generalized case with a particular $\PSO$-operator, namely that shown
in algorithm \ref{alg:PSO-MAA}.

\subsection{The $\GMAA$ Algorithm}

In $\GMAA$ we refer to a `policy pool' $\polPool$ rather than an
open list, as it is a more neutral word which does not imply any ordering.
This policy pool $\polPool$ is initialized with a completely unspecified
joint policy $\partJPolT0=\left(\right)$ and the maximum lower bound
(found so far) $\maxlb$ is set to~$-\infty$. $\maxjpol$ denotes
the best joint policy found so far. %
\begin{algorithm}
\begin{algorithmic}[1]

\STATE $\maxlb\assign-\infty$ 

\STATE $\polPool\assign\{\partJPolT0=()\}$

\REPEAT

\STATE  $\partJPolT\ts\assign\selectO(\polPool)$$ $

\STATE  $\PSOset\assign\PSO(\partJPolT\ts)$

\IF{ \label{alg:GMAA:line:if_found_lowerbounds}$\PSOset$ contains
a subset of full policies $\jpolS_{\PSO}\subseteq\PSOset$}

\STATE  $\jpol'\assign\argmax_{\jpol\in\jpolS_{\PSO}}V(\jpol)$

\IF{$\V(\jpol') > \maxlb $ }

\STATE  $\maxlb\assign V(\jpol')$

\STATE $\maxjpol\assign\jpol'$

\STATE  $\polPool\assign\left\{ \partJPol\in\polPool\mid\VH(\partJPol)>\maxlb\right\} $\COMMENT{prune the policy pool}

\ENDIF

\STATE  $\PSOset\assign\PSOset\setminus\jpolS_{\PSO}$ \COMMENT{remove full policies}

\ENDIF 

\STATE  $\polPool\assign(\polPool\setminus\partJPolT\ts)\cup\left\{ \partJPol\in\PSOset\mid\VH(\partJPol)>\maxlb\right\} $\COMMENT{remove processed/add new partial policies}

\UNTIL{$\polPool$ is empty}

\end{algorithmic}

\caption{\label{alg:GMAA} $\GMAA$}

\end{algorithm}

At this point $\GMAA$ starts. First, the selection operator, $\selectO$,
selects a partial joint policy $\partJPol$ from $\polPool$. We will
assume that, in accordance with $\MAA$, the partial policy with the
highest heuristic value is selected. In general, however, any kind
of selection algorithm may be used. Next, the selected policy is processed
by the policy search operator $\PSO$, which returns a set of (partial)
joint policies $\PSOset$ and their heuristic values. When $\PSO$
returns one or more full policies $\jpol\in\PSOset$, the provided
values $\VH(\jpol)=V(\jpol)$ are a lower bound for an optimal joint
policy, which can be used to prune the search space. Any found partial
joint policies $\partJPol\in\PSOset$ with a heuristic value $\VH(\partJPol)>\maxlb$
are added to $\polPool$. The process is repeated until the policy
pool is empty.

\subsection{The $\PSO$ Operator}

Here we describe some different choices for the $\PSO$-operator and
how they correspond to existing Dec-POMDP solution methods.

\subsubsection{$\MAA$}

$\GMAA$ reduces to standard $\MAA$ by using the $\PSO$-operator
described by Algorithm~\ref{alg:PSO-MAA}. Line \ref{alg:CV-MAA:line_expand}
expands $\partJPolT\ts$ forming $\partJPolS^{\ts+1}$ the set of
partial joint policies for one extra stage. Line \ref{alg:CV-MAA:line_valuate}
valuates all these child policies, where\[
V^{0\dots\ts}(\partJPolT{\ts+1})=V^{0...\ts-1}(\partJPolT{\ts})+E\left[R(s^{\ts},\ja)\big|\partJPolT{\ts+1}\right]\]
gives the true expected reward over the first $\ts+1$ stages. $\VH^{(\ts+1)...h}(\partJPolT{\ts+1})$
is the heuristic value over stages $(\ts+1)...h$ given that $\partJPolT{\ts+1}$
has been followed the first $\ts+1$ stages.%
\begin{algorithm}
\begin{algorithmic}[1]

\STATE \label{alg:CV-MAA:line_expand} $\partJPolS^{\ts+1}\assign\left\{ \partJPolT{\ts+1}=\left\langle \partPolAT{1}{\ts+1},...,\partPolAT{\nrA}{\ts+1}\right\rangle \mid\partPolAT{i}{\ts+1}=\left(\partPolAT i\ts,\drAT i\ts\right),\,\drAT i\ts:\oHistATS i{\ts}\rightarrow\aAS i\right\} $

\STATE \label{alg:CV-MAA:line_valuate} $\forall_{\partJPolT{\ts+1}\in\partJPolS^{\ts+1}}\quad\VH(\partJPolT{\ts+1})\assign V^{0...\ts-1}(\partJPolT{\ts})+E\left[R(s^{\ts},\ja)\big|\partJPolT{\ts+1}\right]+\VH^{(\ts+1)...h}(\partJPolT{\ts+1})$

\RETURN  $\partJPolS^{\ts+1}$

\end{algorithmic}

\caption{$\PSO(\partJPolT\ts)$ --- $\MAA$}

\label{alg:PSO-MAA}
\end{algorithm}

When using an admissible heuristic, $\GMAA$ will never prune a partial
policy that can be expanded into an optimal policy. When combining
this with the fact that the $\MAA$-$\PSO$ operator returns all possible
$\partJPolT{\ts+1}$ for a $\partJPolT{\ts}$, it is clear that when
$\polPool$ becomes empty an optimal policy has been found.

\subsubsection{Forward-Sweep Policy Computation }

Forward-sweep policy computation, as introduced in Section \ref{sub:Q*},
is described by algorithms \ref{alg:GMAA} and \ref{alg:PSO-FSPC}
jointly. Given a partial joint policy $\partJPolT{\ts}$, the $\PSO$
operator now constructs and solves a BG for time step $\ts$. Because
$\PSO$ in algorithm \ref{alg:PSO-FSPC} only returns the best-ranked
policy, $\polPool$ will never contain more than 1 joint policy and
the whole search process reduces to solving BGs for time steps $0,...,h-1$.

The approach of \citet{Emery-Montemerlo04} is identical to forward-sweep
policy computation, except that 1) smaller BGs are created by discarding
or clustering low probability action-observation histories, and 2)
the BGs are approximately solved by alternating maximization. Therefore
this approach can also be incorporated in the $\GMAA$ policy search
framework by making the appropriate modifications in Algorithm~\ref{alg:PSO-FSPC}.

\begin{algorithm}
\begin{algorithmic}[1]

\STATE  $BG\assign\left\langle \mathcal{A},\CoaHistTS{\ts}{\partJPolT\ts},\Pr(\CoaHistTS{\ts}{\partJPolT\ts}),\QH^{\ts}\right\rangle $

\FORALL{$\jpolBG=\left\langle \polBG{1},...,\polBG\nrA\right\rangle \textrm{ s.t. }\polBG i:\oHistATS i{\ts}\rightarrow\aAS i$} 

\STATE $\VH^{\ts}(\jpolBG)\assign\sum_{\oaHistT{\ts}\in\CoaHistTS{\ts}{\partJPol{\ts}}}\Pr(\oaHistT{\ts})\QH^{\ts}(\oaHistT{\ts},\jpolBG(\oaHistT{\ts}))$ 

\STATE  $\partJPolT{\ts+1}\assign\left(\partJPolT{\ts},\jpolBG\right)$

\STATE \label{alg:CV-FSPC:line_valuate}$\VH(\partJPolT{\ts+1})\assign V^{0...\ts-1}(\partJPolT{\ts})+\VH^{\ts}(\jpolBG)$ 

\ENDFOR 

\RETURN  $\argmax_{\partJPolT{\ts+1}}\VH(\partJPolT{\ts+1})$

\end{algorithmic}

\caption{{$\PSO(\partJPolT{\ts})$ --- Forward-sweep policy computation }}

\label{alg:PSO-FSPC}
\end{algorithm}

\subsubsection{Unification}

Here we will give a unified perspective of the $\MAA$ and forward-sweep
policy computation by examining the relation between the corresponding
$\PSO$-operators. In particular we show that, when using any of the
approximate $Q$-value functions described in Section \ref{sec:Approximate-Q-value-functions}
as a heuristic, the sole difference between the two is that FSPC returns
only the joint policy with the highest heuristic value.

\begin{proposition}If a heuristic $\QH$ has the following form\begin{equation}
\QH^{\ts}(\oaHistT{\ts},\ja)=R(\oaHistT{\ts},\ja)+\sum_{\jo^{\ts+1}}\Pr(\jo^{\ts+1}|\oaHistT{\ts},\ja)\VH^{\ts+1}(\oaHistT{\ts+1}),\label{eq:propMAAisFSPC:qh_requirement}\end{equation}
 then for a partial policy $\partJPolT{\ts+1}=\left(\partJPolT{\ts},\jpolBG^{\ts}\right)$
\begin{equation}
\sum_{\oaHistT{\ts}\in\CoaHistTS{\ts}{\partJPol}}\Pr(\oaHistT{\ts})\QH^{\ts}(\oaHistT{\ts},\jpolBG(\oaHistT{\ts}))=E\left[R(s^{\ts},\ja)\big|\partJPolT{\ts+1}\right]+\VH^{(\ts+1)\dots h}(\partJPolT{\ts+1})\label{eq:propMAAisFSPC:line_is_line}\end{equation}
holds.

\proofup  The expectation of $R^{\ts}$ given $\partJPolT{\ts+1}$
can be written as \[
E\left[R(s^{\ts},\ja)\big|\partJPolT{\ts+1}\right]=\sum_{\oaHistT\ts\in\CoaHistTS{\ts}{\partJPol}}\Pr(\oaHistT\ts)\sum_{s\in\mathcal{S}}R(s,\partJPolT{\ts+1}(\oaHistT{\ts}))\Pr(s|\oaHistT\ts)=\sum_{\oaHistT\ts\in\CoaHistTS{\ts}{\partJPol}}\Pr(\oaHistT\ts)R(\oaHistT\ts,\partJPolT{\ts+1}(\oaHistT{\ts})).\]
Also, we can rewrite $\VH^{(\ts+1)\dots h}(\partJPolT{\ts+1})$ as\[
\VH^{(\ts+1)\dots h}(\partJPolT{\ts+1})=\sum_{\oaHistT\ts\in\CoaHistTS{\ts}{\partJPol}}\Pr(\oaHistT\ts)\sum_{\jo^{\ts+1}}\Pr(\jo^{\ts+1}|\oaHistT\ts,\partJPolT{\ts+1}(\oaHistT{\ts}))\VH^{(\ts+1)...h}(\oaHistT{\ts+1}),\]
such that\begin{multline}
E\left[R(s^{\ts},\ja)\big|\partJPolT{\ts+1}\right]+\VH^{(\ts+1)\dots h}(\partJPolT{\ts+1})=\sum_{\oaHistT\ts\in\CoaHistTS{\ts}{\partJPol}}\Pr(\oaHistT\ts)\\
\left[R(\oaHistT\ts,\partJPolT{\ts+1}(\oaHistT{\ts}))+\sum_{\jo^{\ts+1}}\Pr(\jo^{\ts+1}|\oaHistT\ts,\partJPolT{\ts+1}(\oaHistT{\ts}))\VH^{(\ts+1)...h}(\oaHistT{\ts+1})\right]\label{eq:Rt+1_+_VHt+2}\end{multline}
Therefore, assuming \eqref{eq:propMAAisFSPC:qh_requirement} yields
\eqref{eq:propMAAisFSPC:line_is_line}.\end{proof}\end{proposition}

This means that if a heuristic satisfies \eqref{eq:propMAAisFSPC:qh_requirement},
which is the case for all the Q-value functions we discussed in this
paper, the $\PSO$ operators of algorithms \ref{alg:PSO-MAA} and
\ref{alg:PSO-FSPC} evaluate the expanded policies the same. I.e.,
algorithms \ref{alg:PSO-MAA} and \ref{alg:PSO-FSPC} calculate identical
heuristic values for the same next time step joint policies. Also
the expanded policies $\partJPolT{\ts+1}$ are formed in the same
way: by considering all possible $\jdrT\ts$ respectively $\jpolBG^{\ts}$
to extend $\partJPolT{\ts}$. Therefore, the sole difference in this
case is that the latter returns only the joint policy with the highest
heuristic value.

Clearly there is a computation time/quality trade-off between $\MAA$
and FSPC: $\MAA$ is guaranteed to find an optimal policy (given an
admissible heuristic), while FSPC is guaranteed to finish in one forward
sweep. We propose a generalization, that returns the $k$-best ranked
policies. We refer to this as the `$k$-best joint BG policies' $\GMAA$
variant, or $\kGMAA$. In this way, $\kGMAA$ reduces to forward-sweep
policy computation for $k=1$ and to $\MAA$ for $k=\infty$.

\section{Experiments}

\label{sec:Experiments}

In order to compare the different approximate Q-value functions discussed
in this work, as well as to show the flexibility of the $\GMAA$ algorithm,
we have performed several experiments. We use $\QMDP$, $\QPOMDP$
and $\QBG$ as heuristic estimates of $Q^{*}$. We will provide some
qualitative insight in the different Q-value functions we considered,
as well as results on computing optimal policies using $\MAA$, and
on the performance of forward-sweep policy computation. First we will
describe our problem domains, some of which are standard test problems,
while others are introduced in this work.

\subsection{Problem Domains}

In Section \ref{sub:The-decentralized-tiger} we discussed the decentralized
tiger (\<dectig>) problem as introduced by \citet{Nair03_JESP}.
Apart from the standard \<dectig> domain, we consider a modified
version, called \<sdectig>, in which the start distribution is not
uniform. Instead, initially the tiger is located on the left with
probability $0.8$. We also include results from the BroadcastChannel
problem, introduced by \citet{Hansen04}, which models two nodes that
have to cooperate to maximize the throughput of a shared communication
channel. Furthermore, a test problem called {}``Meeting on a Grid''
is provided by \citet{Bernstein05}, in which two robots navigate
on a two-by-two grid. We consider the version with 2 observations
per agent \citep{Amato06msdm}.

\begin{figure}
\begin{centering}

\small

\parbox{2.1in}{ 

\input{results/fireFighting_2_3_3/h3policyAgent0} 

}\qquad\qquad\parbox{2.1in}{ 

\input{results/fireFighting_2_3_3/h3policyAgent1} 

}

\end{centering}

\caption{Optimal policy for FireFighting $\langle n_{h}=3,n_{f}=3\rangle$,
horizon 3. On the left the policy for the first agent, on the right
the second agent's policy. \label{fig:FF233-h3-Policies}}

\end{figure}

We introduce a new benchmark problem, which models a team of $\nrA$
fire fighters that have to extinguish fires in a row of $n_{h}$ houses.
Each house is characterized by an integer parameter~$f$, or fire
level. It indicates to what degree a house is burning, and it can
have $n_{f}$ different values, $0\leq f<n_{f}$. Its minimum value
is 0, indicating the house is not burning. At every time step, the
agents receive a reward of $-f$ for each house and each agent can
choose to move to any of the houses to fight fires at that location.
If a house is burning ($f>0$) and no fire fighting agent is present,
its fire level will increase by one point with probability $0.8$
if any of its neighboring houses are burning, and with probability
$0.4$ if none of its neighbors are on fire. A house that is not burning
can only catch fire with probability $0.8$ if one of its neighbors
is on fire. When two agents are in the same house, they will extinguish
any present fire completely, setting the house's fire level to 0.
A single agent present at a house will lower the fire level by one
point with probability 1 if no neighbors are burning, and with probability
$0.6$ otherwise. Each agent can only observe whether there are flames
or not at its location. Flames are observed with probability $0.2$
if $f=0$, with probability $0.5$ if $f=1$, and with probability
$0.8$ otherwise. Initially, the agents start outside any of the houses,
and the fire level $f$ of each house is drawn from a uniform distribution. 

We will test different variations of this problems, where the number
of agents is always 2, but which differ in the number of houses and
fire levels. In particular, we will consider $\langle n_{h}=3,n_{f}=3\rangle$
and $\langle n_{h}=4,n_{f}=3\rangle$. \fig\ref{fig:FF233-h3-Policies}
shows an optimal joint policy for horizon 3 of the former variation.
One agent initially moves to the middle house to fight fires there,
which helps prevent fire from spreading to its two neighbors. The
other agent moves to house 3, and stays there if it observes fire,
and moves to house 1 if it does not observe flames. As well as being
optimal, such a joint policy makes sense intuitively speaking.

\subsection{Comparing Q-value Functions}

\providecommand{\Sl}{$s_l$}
\providecommand{\Sr}{$s_r$}%
\begin{figure}[t]
\begin{centering}
{
\psfrag{Qmax - max_aQ(b,a)}[cc][b]{Qmax = }
\includegraphics[width=0.5\columnwidth]{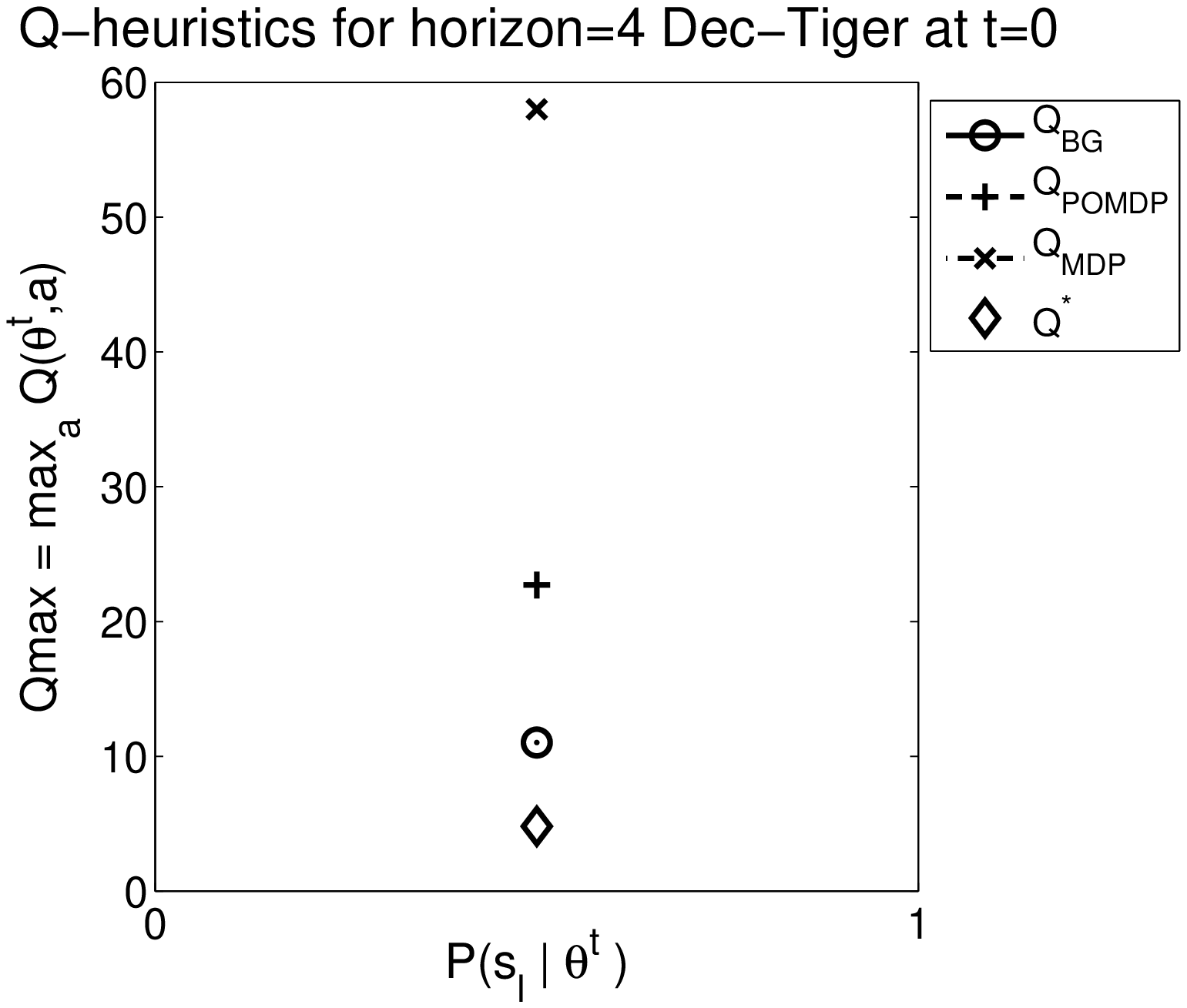}\includegraphics[width=0.5\columnwidth]{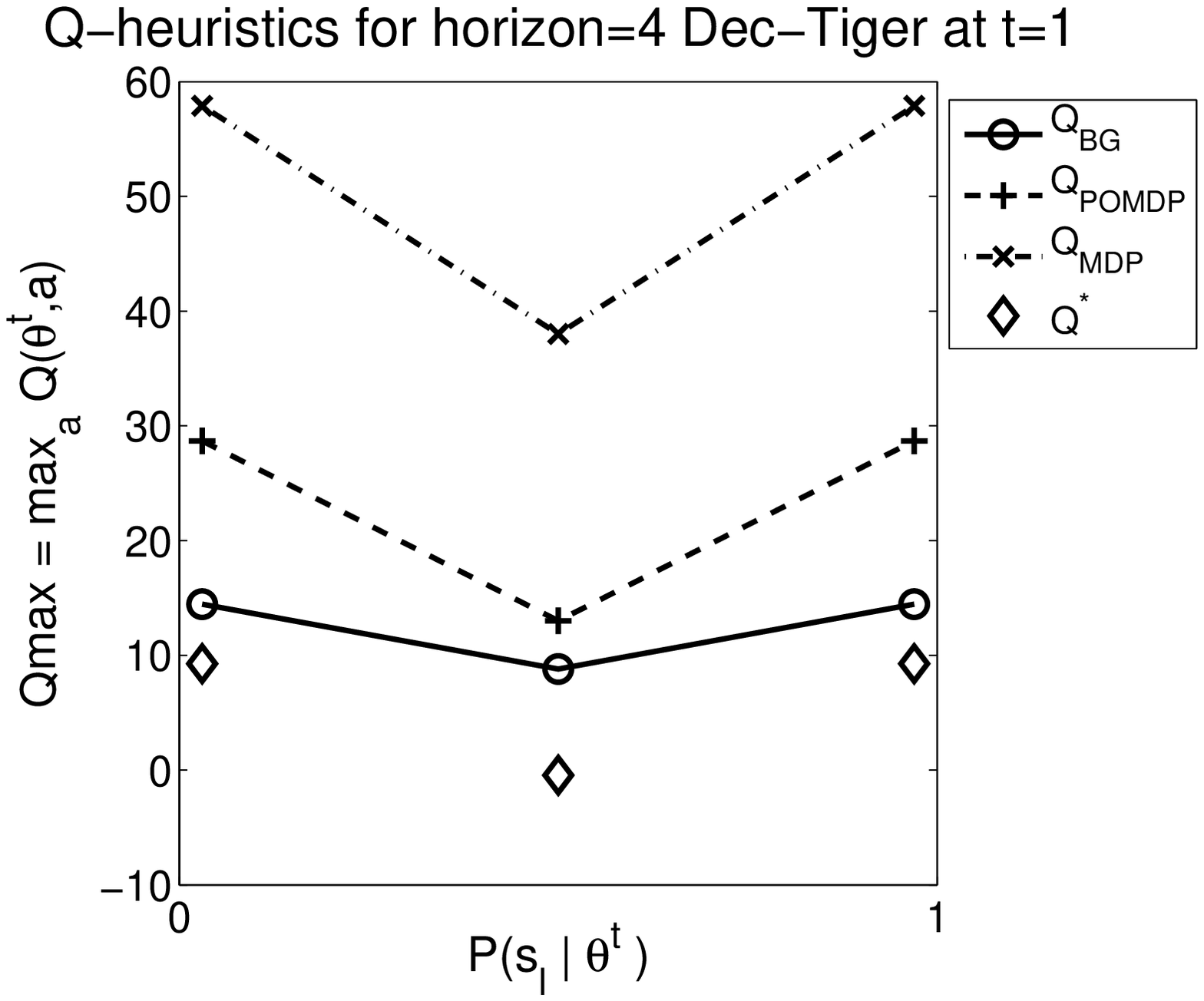}\\
\includegraphics[width=0.5\columnwidth]{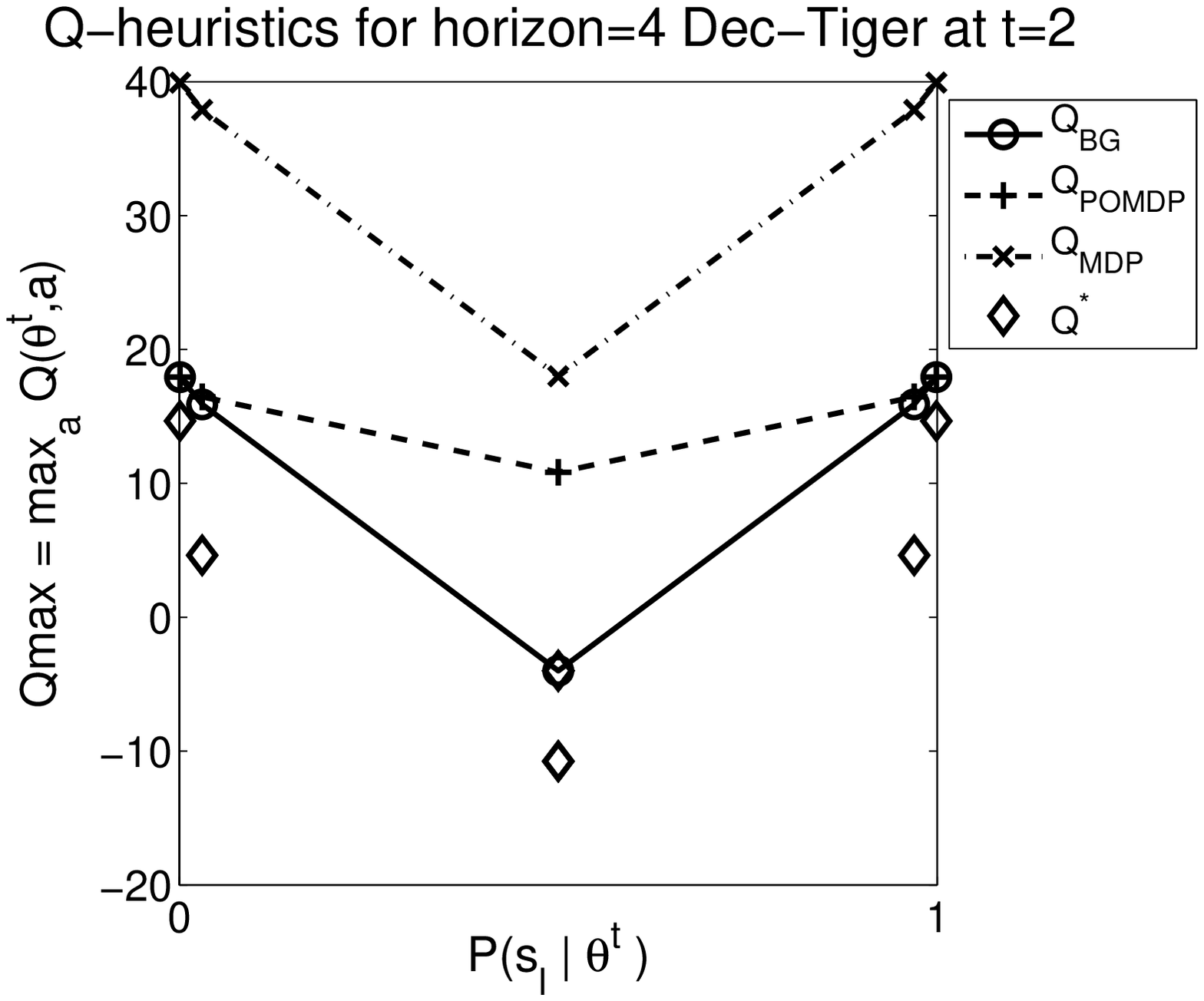}\includegraphics[width=0.5\columnwidth]{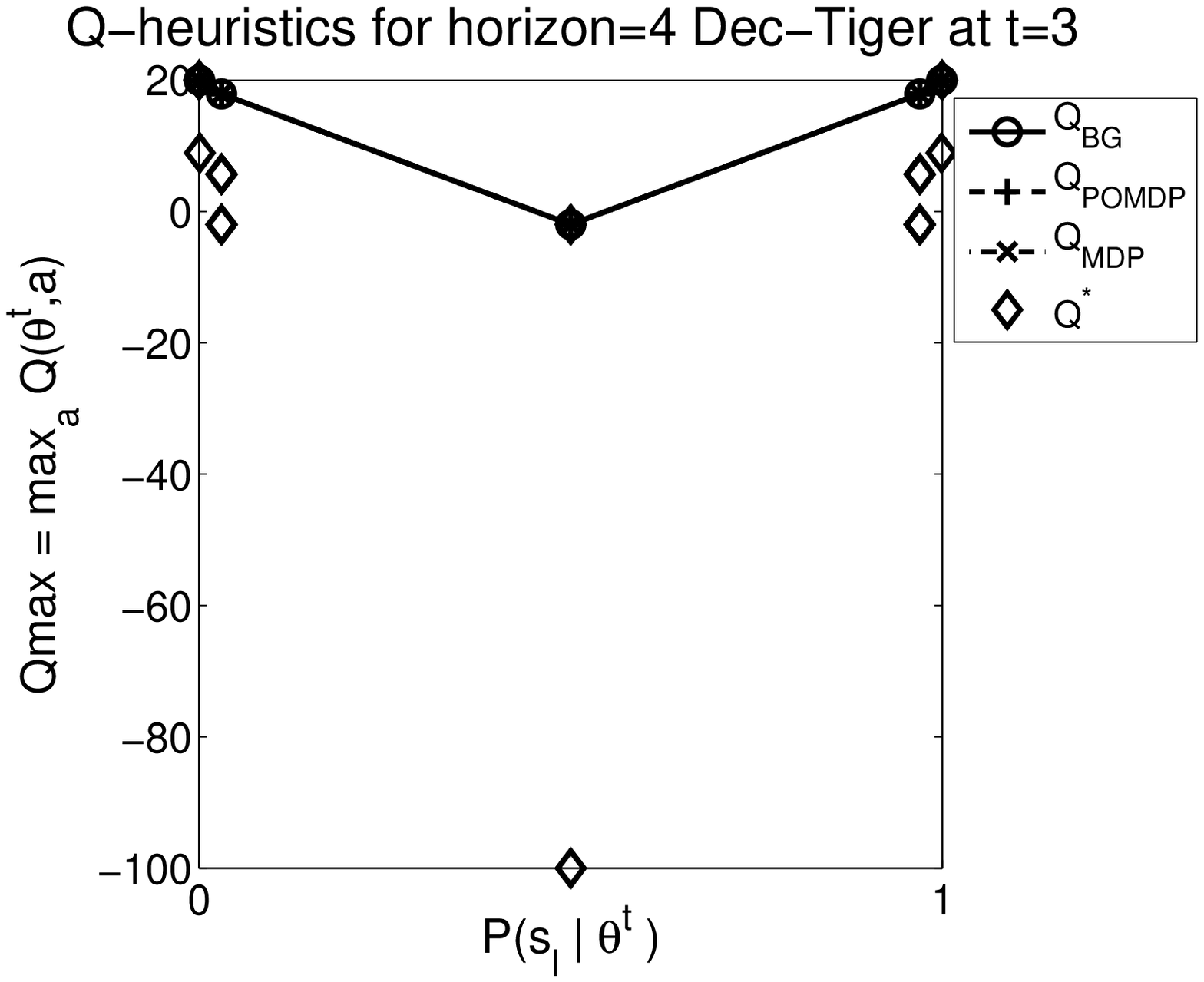}}
\par\end{centering}

\caption{\label{fig:Q-vals_plot}Q-values for horizon 4 \<dectig>. For each
$\oaHistT\ts$, corresponding to some $\Pr(\textrm{\Sl}|\oaHistT\ts)$,
the maximal $Q(\oaHistT\ts,\ja)$-value is plotted.}

\end{figure}
\begin{figure}
\begin{centering}
\includegraphics[width=0.65\columnwidth]{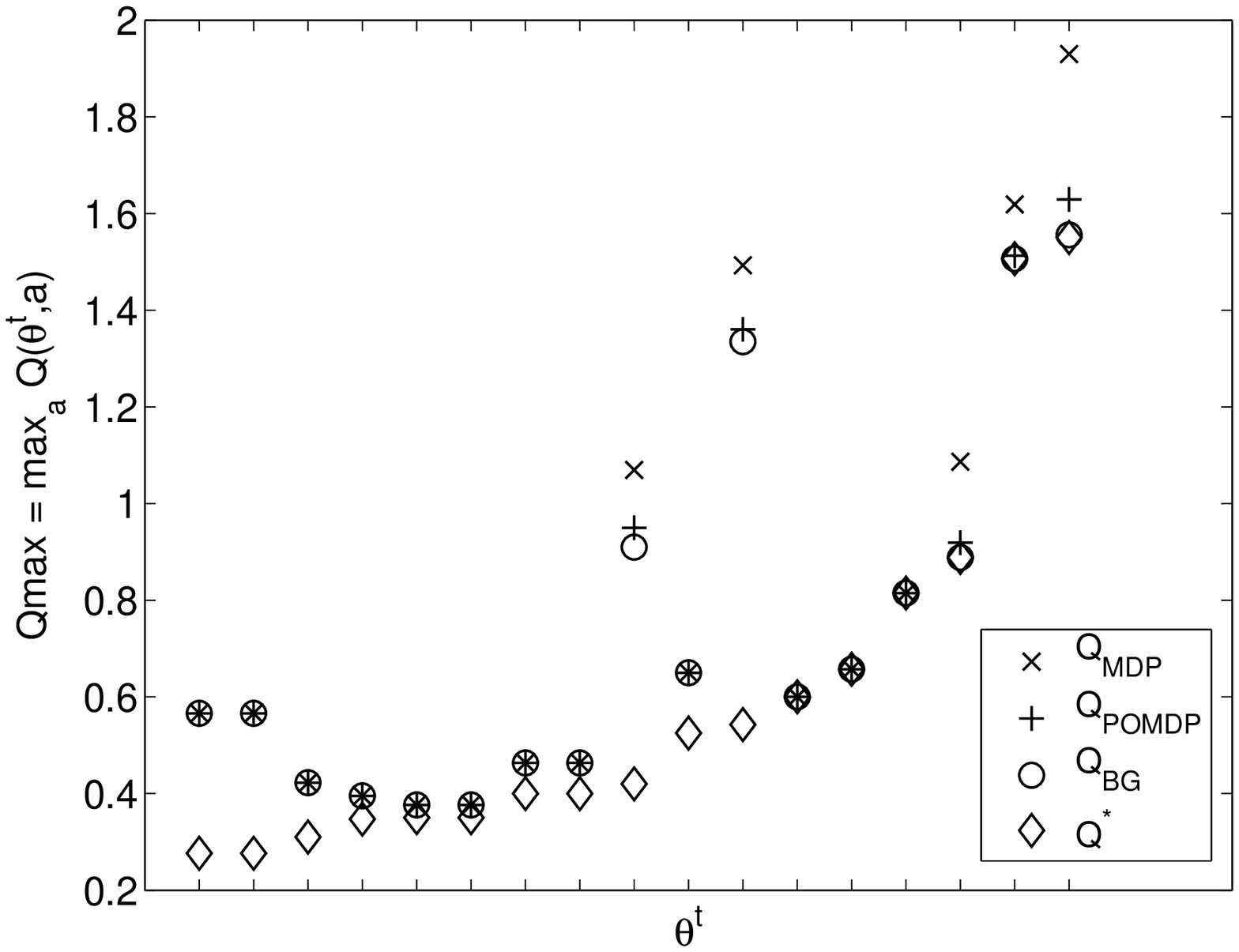}
\par\end{centering}

\caption{Comparison of maximal $Q(\oaHistT\ts,\ja)$-values for Meeting on
a Grid. We plot the value of all $\theta^{t}$ that can be reached
by an optimal policy, ordered according their optimal value.\label{fig:GridSmallQComparison}}

\end{figure}

Before providing a  comparison of performance of some of the approximate
Q-value functions described in this work, we will first give some
more insights in their actual values. For the $h=4$ \<dectig> problem,
we generated all possible $\oaHistT\ts$ and the corresponding $\Pr(\textrm{\Sl}|\oaHistT\ts)$,
according to \eqref{eq:P(s,oaHist)}. For each of these, the maximal
$Q(\oaHistT\ts,\ja)$-value is plotted in Figure~\ref{fig:Q-vals_plot}.
Apart from the three approximate Q-value functions, we also plotted
the optimal value for each joint action-observation history $\oaHistT\ts$
that can be realized when using $\jpol^{*}$. Note that different
$\oaHistT\ts$ can have different optimal values, but induce the same
$\Pr(\textrm{\Sl}|\oaHistT\ts)$, as demonstrated in the figure: there
are multiple $Q^{*}$-values plotted for some $\Pr(\textrm{\Sl}|\oaHistT\ts)$.
For the horizon 3 Meeting on a Grid problem we also collected all
$\oaHistT\ts$ that can be visited by the optimal policy, and in \fig\ref{fig:GridSmallQComparison}
we again plotted maximal $Q(\oaHistT\ts,\ja)$-values. Because this
problem has many states, a representation as in \fig\ref{fig:Q-vals_plot}
is not possible. Instead, we ordered the $\oaHist$ according to their
optimal value. We can see that the bounds are tight for some $\oaHist$,
while for others they can be quite loose. However, when used in the
$\GMAA$ framework, their actual performance as a heuristic also depends
on their valuation of $\oaHist\in\oaHistS$ not shown by \fig\ref{fig:GridSmallQComparison},
namely those that will \emph{not} be visited by an optimal policy:
especially when these are overestimated, $\GMAA$ will first examine
a sub-optimal branch of the search tree. A tighter upper bound can
speed up computation to a very large extent, as it allows the algorithm
to prune the policy pool more, reducing the number of Bayesian games
that need to be solved. Both figures clearly illustrate the main property
of the upper bounds we discussed, namely that $Q^{*}\leq\QBG\leq\QPOMDP\leq\QMDP$
(see Theorem \ref{thm:upperbounds}).

\subsection{Computing Optimal Policies}

As shown above, the hierarchy of upper bounds $Q^{*}\leq\QBG\leq\QPOMDP\leq\QMDP$
is not just a theoretical construct, but the differences in value
specified can be significant for particular problems. In order to
evaluate what the impact is of the differences between the approximate
Q-value functions, we performed several experiments. Here we describe
our evaluation of $\MAA$ on a number of test problems using $\QBG$,
$\QPOMDP$ and $\QMDP$ as heuristic. All timing results in this paper
are CPU times with a resolution of $0.01$s, and were obtained on
$3.4$GHz Intel Xeon processors.

\newcommand{\zeros}{0} 

\begin{table}
  \newcommand{\hhhhline}{\hhline{|~~|----|}}
  \centering
  \begin{tabular}{|ccrrrr|}
    \hline
    $h$&$V^*$&&$n_\partJPol$
    &$T_\GMAA$&$T_Q$\\
    \hline
    \hline
    \multirow{3}{*}{3}&\multirow{3}{*}{
      $5.1908$ 
    }
     &$\QMDP$&
    $105,228$
    & $0.31$ s& $\zeros$ s\\
    \hhhhline
    &&$\QPOMDP$&
    $6,651$ 
    & $0.02$ s& $\zeros$ s\\
    \hhhhline
    &&$\QBG$&
    $6,651$
    & $0.02$ s& $0.02$ s\\
    \hline
    \hline
    \multirow{3}{*}{4}&\multirow{3}{*}{
      $4.8028$ 
    }
     &$\QMDP$&
    $37,536,938,118$ 
    & $431,776$ s& $\zeros$ s\\
    \hhhhline
    &&$\QPOMDP$&
    $559,653,390$
    & $5,961$ s& $0.13$ s\\
    \hhhhline
    &&$\QBG$&
    $301,333,698$
    & $3,208$ s& $0.94$ s\\
    \hline

  \end{tabular}
  \caption{$\MAA$ results for Dec-Tiger.}
  \label{tab:MAAstardectiger}
\end{table}

\begin{table}
  \newcommand{\hhhhline}{\hhline{|~~|----|}}
  \centering
  \begin{tabular}{|ccrrrr|}
    \hline
    $h$&$V^*$&&$n_\partJPol$
    &$T_\GMAA$&$T_Q$\\
    \hline
    \hline
    \multirow{3}{*}{3}&\multirow{3}{*}{
      $5.8402$ 
    }
     &$\QMDP$&
    $151,236$
    & $0.46$ s& $\zeros$ s\\
    \hhhhline
    &&$\QPOMDP$&
    $19,854$ 
    & $0.06$ s& $0.01$ s\\
    \hhhhline
    &&$\QBG$&
    $13,212$ 
    & $0.04$ s& $0.03$ s\\
    \hline
    \hline
    \multirow{3}{*}{4}&\multirow{3}{*}{
      $11.1908$ 
    }
     &$\QMDP$&
    $33,921,256,149$ 
    & $388,894$ s& $\zeros$ s\\
    \hhhhline
    &&$\QPOMDP$&
    $774,880,515$ 
    & $8,908$ s& $0.13$ s\\
    \hhhhline
    &&$\QBG$&
    $86,106,735$
    & $919$ s& $0.92$ s\\
    \hline

  \end{tabular}
  \caption{$\MAA$ results for Skewed Dec-Tiger.}
  \label{tab:MAAstardectiger_skewed}
\end{table}

\tab\ref{tab:MAAstardectiger} shows the results $\MAA$ obtained
on the original \<dectig> problem for horizon 3 and 4. It shows for
each heuristic the number of partial joint policies evaluated $n_{\partJPol}$,
CPU time spent on the $\GMAA$ phase $T_{\GMAA}$, and CPU time spent
on calculating the heuristic $T_{Q}$. As $\QBG$, $\QPOMDP$ and
$\QMDP$ are upper bounds to $Q^{*}$, $\MAA$ is guaranteed to find
the optimal policy when using them as heuristic, however the timing
results may differ. 

For $h=3$ we see that using $\QPOMDP$ and $\QBG$ only a fraction
of the number of policies are evaluated when compared to $\QMDP$
which reflects proportionally in the time spent on $\GMAA$. For this
horizon $\QPOMDP$ and $\QBG$ perform the same, but the time needed
to compute the $\QBG$ heuristic is as long as the $\GMAA$-phase,
therefore $\QPOMDP$ outperforms $\QBG$ here. For $h=4$, the impact
of using tighter heuristics becomes even more pronounced. In this
case the computation time of the heuristic is negligible, and $\QBG$
outperforms both, as it is able to prune much more partial joint policies
from the policy pool. \tab\ref{tab:MAAstardectiger_skewed} shows
results for \<sdectig>. For this problem the $\QMDP$ and $\QBG$
results are roughly the same as the original \<dectig> problem; for
$h=3$ the timings are a bit slower, and for $h=4$ they are faster.
For $\QPOMDP,$ however, we see that for $h=4$ the results are slower
as well and that $\QBG$ outperforms $\QPOMDP$ by an order of magnitude.

\begin{table}
  \newcommand{\hhhhline}{\hhline{|~~|----|}}
  \centering
  \begin{tabular}{|ccrrrr|}
    \hline
    $h$&$V^*$&&$n_\partJPol$
    &$T_\GMAA$&$T_Q$\\
    \hline
    \multirow{3}{*}{4}&\multirow{3}{*}{
      $3.8900$ 
    }
     &$\QMDP$&
    $328,212$ 
    & $3.54$ s& $\zeros$ s\\
    \hhhhline
    &&$\QPOMDP$&
    $531$ 
    & $\zeros$ s& $0.01$ s\\
    \hhhhline
    &&$\QBG$&
    $531$ 
    & $\zeros$ s& $0.03$ s\\
    \hline
    \hline
    \multirow{3}{*}{5}&\multirow{3}{*}{
      $4.7900$ 
    }
     &$\QMDP$&
    $$ 
    N/A& $> 4.32\textrm{e}5$ s& $\zeros$ s\\
    \hhhhline
    &&$\QPOMDP$&
    $196,883$ 
    & $5.30$ s& $0.20$ s\\
    \hhhhline
    &&$\QBG$&
    $196,883$ 
    & $5.15$ s& $0.53$ s\\
    \hline

  \end{tabular}
  \caption{$\MAA$ results for BroadcastChannel.}
  \label{tab:MAAstarbroadcastChannel}
\end{table}

\begin{table}
  \newcommand{\hhhhline}{\hhline{|~~|----|}}
  \centering
  \begin{tabular}{|ccrrrr|}
    \hline
    $h$&$V^*$&&$n_\partJPol$
    &$T_\GMAA$&$T_Q$\\
    \hline
    \hline
    \multirow{3}{*}{2}&\multirow{3}{*}{
      $0.9100$ 
    }
     &$\QMDP$&
    $1,275$ 
    & $\zeros$ s& $\zeros$ s\\
    \hhhhline
    &&$\QPOMDP$&
    $1,275$ 
    & $\zeros$ s& $\zeros$ s\\
    \hhhhline
    &&$\QBG$&
    $194$ 
    & $\zeros$ s& $\zeros$ s\\
    \hline
    \hline
    \multirow{3}{*}{3}&\multirow{3}{*}{
      $1.5504$ 
    }
     &$\QMDP$&
    $29,688,775$
    & $81.93$ s& $\zeros$ s\\
    \hhhhline
    &&$\QPOMDP$&
    $3,907,525$ 
    & $10.80$ s& $0.15$ s\\
    \hhhhline
    &&$\QBG$&
    $1,563,775$ 
    & $4.44$ s& $1.37$ s\\
    \hline

  \end{tabular}
  \caption{$\MAA$ results for Meeting on a Grid.}
  \label{tab:MAAstarGridSmall}
\end{table}

\begin{table}
  \newcommand{\hhhhline}{\hhline{|~~|----|}}
  \centering
  \begin{tabular}{|ccrrrr|}
    \hline
    $h$&$V^*$&&$n_\partJPol$
    &$T_\GMAA$&$T_Q$\\
    \hline
    \hline
    \multirow{3}{*}{3}&\multirow{3}{*}{
      $-5.7370$ 
    }
     &$\QMDP$&
    $446,724$
    & $1.58$ s& $0.56$ s\\
    \hhhhline
    &&$\QPOMDP$&
    $26,577$ 
    & $0.08$ s& $0.21$ s\\
    \hhhhline
    &&$\QBG$&
    $26,577$
    & $0.08$ s& $0.33$ s\\
    \hline
    \hline
    \multirow{3}{*}{4}&\multirow{3}{*}{
      $-6.5788$ 
    }
     &$\QMDP$&
     $25,656,607,368$ 
     & $309,235$ s& $0.85$ s\\
     \hhhhline
     &&$\QPOMDP$&
     $516,587,229$ 
     & $5,730$ s& $7.22$ s\\
    \hhhhline
    &&$\QBG$&
    $516,587,229$ 
    & $5,499$ s& $11.72$ s\\
    \hline

  \end{tabular}
  \caption{$\MAA$ results for Fire Fighting $\langle n_h=3, n_f=3 \rangle$.}
  \label{tab:MAAstarFF233}
\end{table}

Results for the Broadcast Channel (\tab\ref{tab:MAAstarbroadcastChannel}),
Meeting on a Grid (\tab\ref{tab:MAAstarGridSmall}) and a Fire fighting
problem (\tab\ref{tab:MAAstarFF233}) are similar. The N/A entry in
\tab\ref{tab:MAAstarbroadcastChannel}  indicates the $\QMDP$ was
not able to compute a solution within 5 days. For these problems we
also see that the performance of $\QPOMDP$ and $\QBG$ is roughly
equal. For the Meeting on a Grid problem, $\QBG$ yields a significant
speedup over $\QPOMDP$.

\subsection{Forward-Sweep Policy Computation}

The $\MAA$ results described above indicate that the use of a tighter
heuristic can yield substantial time savings. In this section, the
approximate Q-value functions are used in forward-sweep policy computation.
We would expect that when using a Q-value function that more closely
resembles $Q^{*}$, the quality of the resulting policy will be higher.
We also tested whether $\kGMAA$ with $k>1$ improved the quality
of the computed policies. In particular, we tested $k=1,2,\dots,5$.

\begin{figure}
\begin{centering}

\small

\parbox{1.3in}{ 

\input{results/dectiger/normal_dectiger-h4-QMDP-policy_agent0} 

}\qquad\qquad\qquad\parbox{1.3in}{ 

\input{results/dectiger/normal_dectiger-h4-QPOMDP+QBG-policy_agent0} 

}

\end{centering}

\caption{Policies found using forward-sweep policy computation (i.e., $k=1$)
for the $h=4$ \<dectig> problem. Left: the policy resulting from
$\QMDP$. Right: the optimal policy as calculated by $\QPOMDP$ and
$\QBG$. The framed entries highlight the crucial differences.\label{fig:DecTiger-h3-Policies}}

\end{figure}

\begin{figure}[tb]
\psfrag{V}{\scriptsize $V$}
\psfrag{k}{\scriptsize $k$}
\centering
\hspace{0.03\columnwidth}\subfigure[\<sdectig>, $h=3$.]{
\includegraphics[height=3cm]{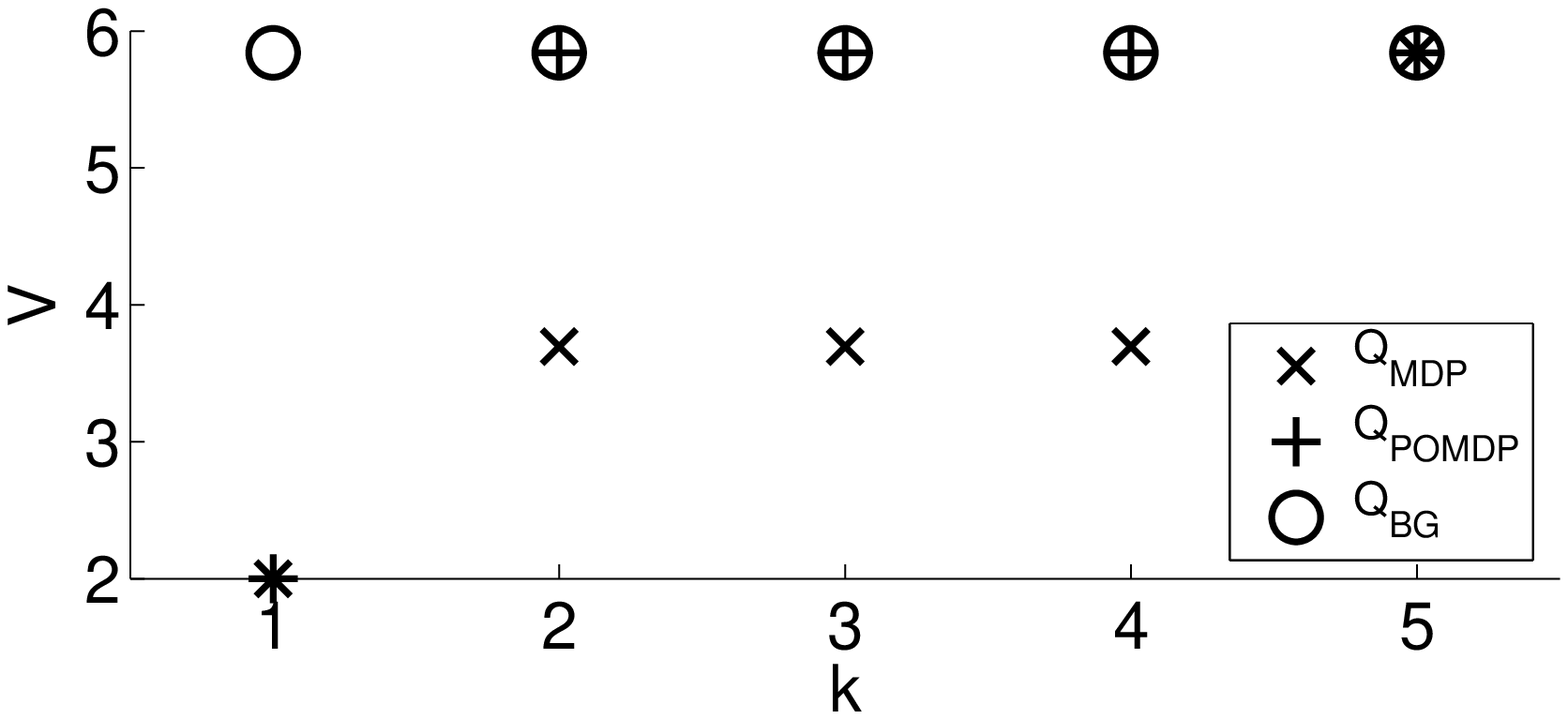}\label{fig:kGMAA-skewedDecTiger3}}\hspace{0.07\columnwidth}
\subfigure[\<sdectig>, $h=4$.]{
\includegraphics[height=3cm]{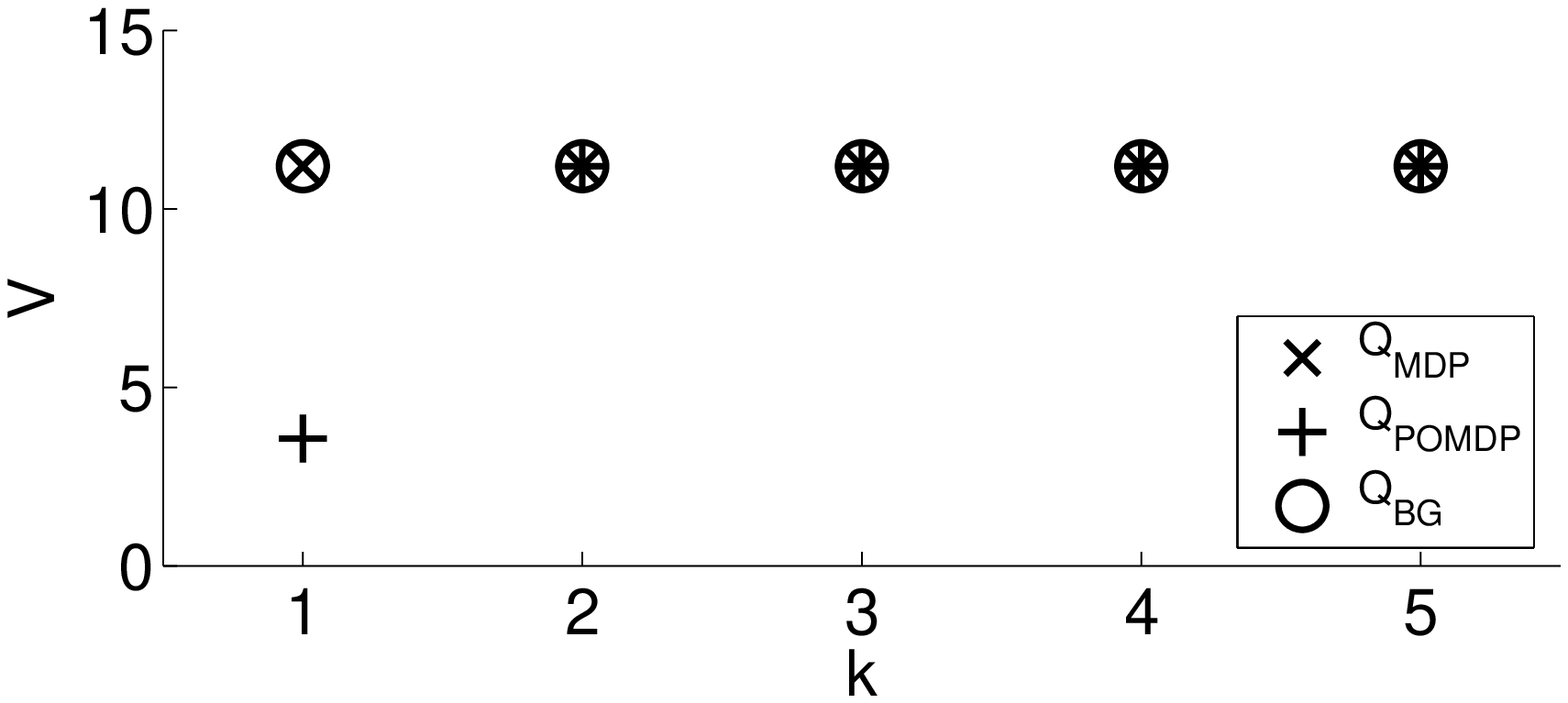}\label{fig:kGMAA-skewedDecTiger4}}

\subfigure[GridSmall, $h=3$.]{
\includegraphics[height=3cm]{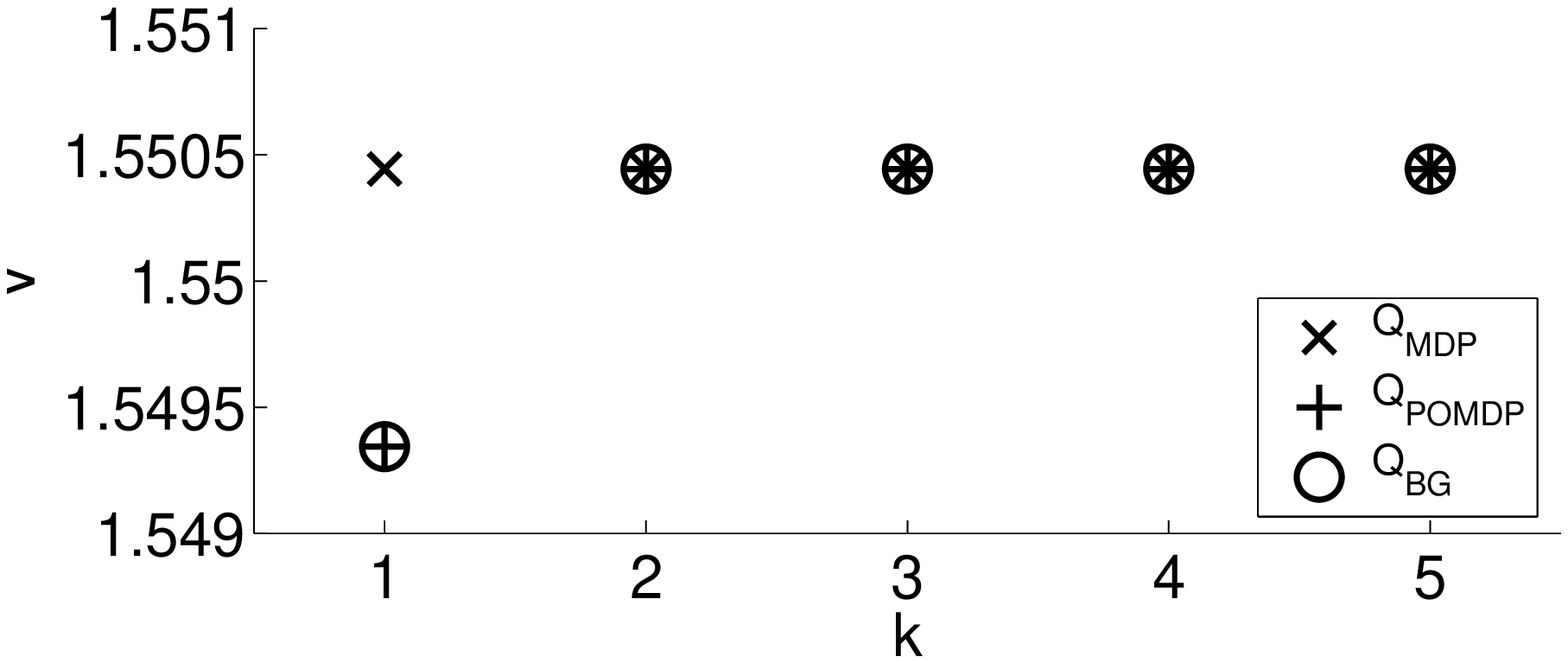}\label{fig:kGMAA-GridSmall}}\hspace{0.05\columnwidth}
\subfigure[FireFighting with $\langle
n_{h}=3,n_{f}=3\rangle$, $h=4$.]{
\includegraphics[height=3cm]{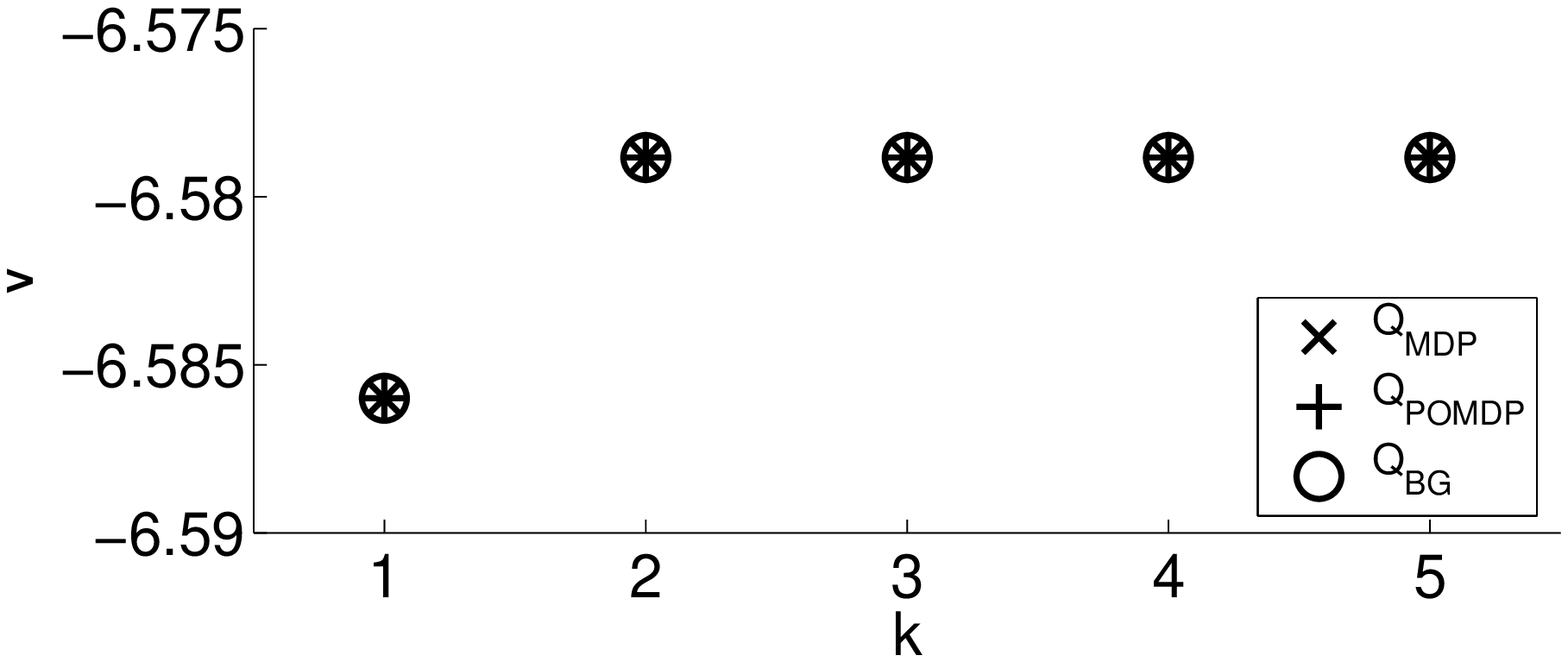}\label{fig:kGMAA-FF233}}

\subfigure[FireFighting with $\langle
n_{h}=4,n_{f}=3\rangle$, $h=3$.]{
\includegraphics[height=3cm]{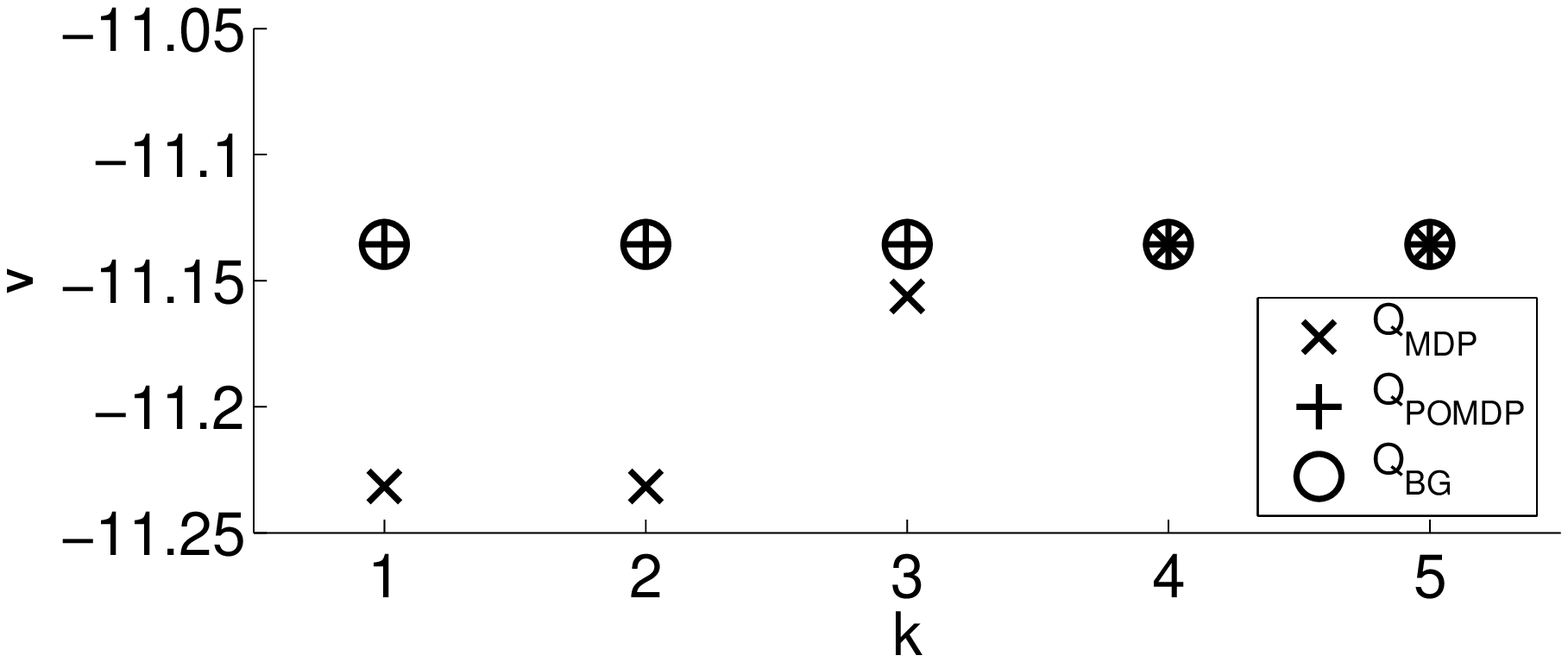}\label{fig:kGMAA-FF243h3}}\hspace{0.05\columnwidth}
\subfigure[FireFighting with $\langle
n_{h}=4,n_{f}=3\rangle$, $h=4$.]{
\includegraphics[height=3cm]{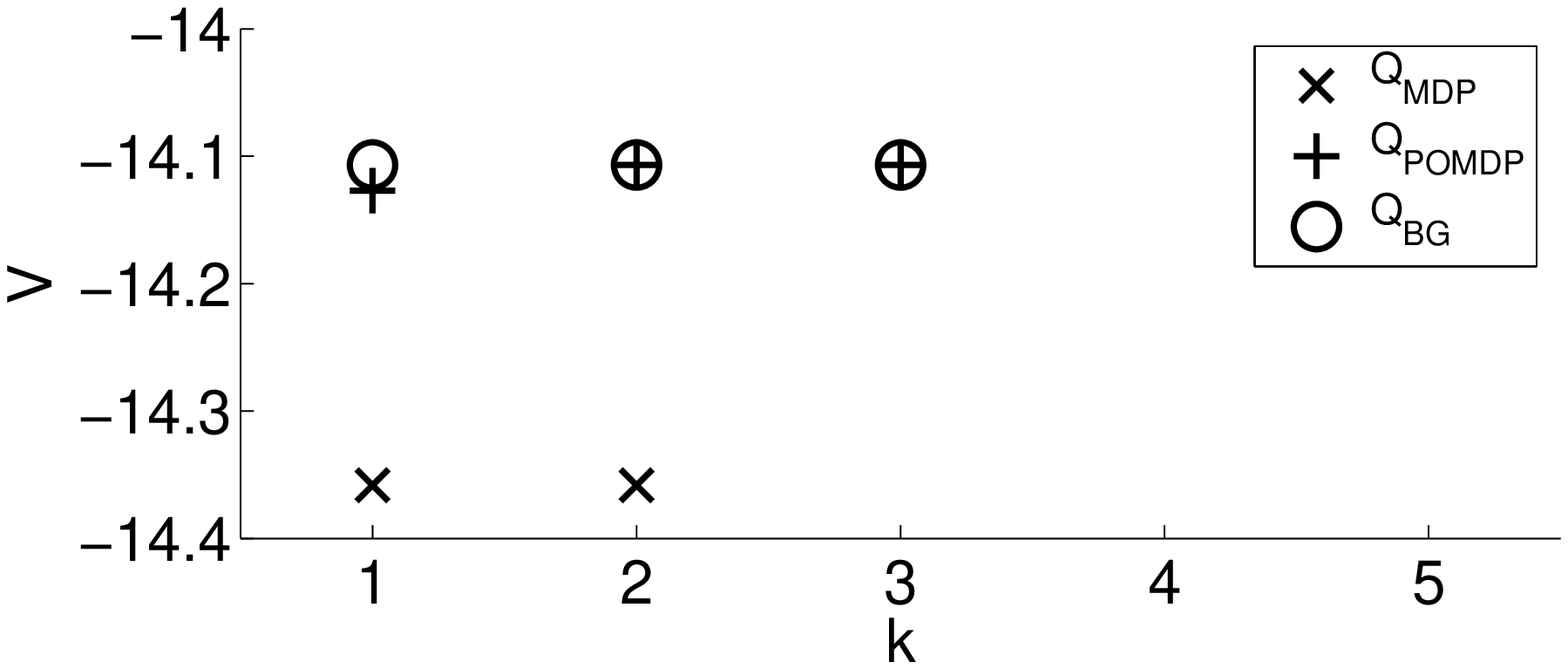}\label{fig:kGMAA-FF243h4}}

\caption{$\kGMAA$ results for different problems and horizons. The
  $y$-axis indicates value of the initial joint belief, while the
  $x$-axis denotes $k$.}
\vspace{-1em}  
\end{figure}

For the \<dectig> problem, $\kGMAA$ with $k=1$ (and thus also $2\leq k\leq5$)
found the optimal policy (with $V(\jpol^{*})=5.19$) for horizon $3$
using all approximate Q-value functions. For horizon $h=4$, also
all different values of $k$ produced the same result for each approximate
Q-value function. In this case, however, $\QMDP$ found a policy with
expected return of $3.19$. $\QPOMDP$ and $\QBG$ did find the optimal
policy ($V(\jpol^{*})=4.80)$. \fig\ref{fig:DecTiger-h3-Policies}
illustrates the optimal policy (right) and the one found by $\QMDP$
(left). It shows that $\QMDP$ overestimates the value for opening
the door in stage $\ts=2$.

For the \<sdectig> problem, different values of $k$ did produce
different results. In particular, for $h=3$ only $\QBG$ finds the
optimal policy (and thus attains the optimal value) for all values
of $k$, as shown in \fig\ref{fig:kGMAA-skewedDecTiger3}. $\QPOMDP$
does find it starting from $k=2$, and $\QMDP$ only from $k=5$.
\fig\ref{fig:kGMAA-skewedDecTiger4} shows a somewhat unexpected
result for $h=4$: here for $k=1$ $\QMDP$ and $\QBG$ find the optimal
policy, but $\QPOMDP$ doesn't. This clearly illustrates that a tighter
approximate Q-value function is not a guarantee for a better joint
policy, which is also illustrated by the results for GridSmall in
\fig\ref{fig:kGMAA-GridSmall}.

We also performed the same experiment for two settings of the FireFighting
problem. For $\langle n_{h}=3,n_{f}=3\rangle$ and $h=3$ all Q-value
functions found the optimal policy (with value $-5.7370$) for all
$k$, and horizon $4$ is shown in \fig\ref{fig:kGMAA-FF233}. \figs\ref{fig:kGMAA-FF243h3}
and \ref{fig:kGMAA-FF243h4} show the results for $\langle n_{h}=4,n_{f}=3\rangle$.
For $h=4$, $\QMDP$ did not finish for $k\geq3$ within 5 days. 

It is encouraging that for all experiments $\kGMAA$ using $\QBG$
and $\QPOMDP$ with $k\leq2$ found the optimal policy. Using $\QMDP$
the optimal policy was also always found with $k\leq5$, except in
horizon $4$ \<dectig> and the $\langle n_{h}=4,n_{f}=3\rangle$
FireFighting problem. These results seem to indicate that this type
of approximation might be likely to produce (near-) optimal results
for other domains as well.

\section{Conclusions}

\label{sec:Conclusion-and-discussion}

A large body of work in single-agent decision-theoretic planning is
based on value functions, but such theory has been lacking thus far
for Dec-POMDPs. Given the large impact of value functions on single-agent
planning under uncertainty, we expect that a thorough study of value
functions for Dec-POMDPs can greatly benefit multiagent planning under
certainty. In this work, we presented a framework of Q-value functions
for Dec-POMDPs, providing a significant contribution to fill this
gap in Dec-POMDP theory. Our theoretical contributions have lead to
new insights, which we applied to improve and extend solution methods.

We have shown how an optimal joint policy $\jpol^{*}$ induces an
optimal Q-value function $Q^{*}(\oaHistT\ts,\ja)$, and how it is
possible to construct the optimal policy $\jpol^{*}$ using forward-sweep
policy computation. This entails solving Bayesian games for time steps
$\ts=0\,,...,\, h-1$ which use $Q^{*}(\oaHistT\ts,\ja)$ as the payoff
function. Because there is no clear way to compute $Q^{*}(\oaHistT\ts,\ja)$,
we introduced a different description of the optimal Q-value function
$Q^{*}(\oaHistT{\ts},\pJPolT{\ts+1})$ that is based on sequential
rationality. This new description of $Q^{*}$ can be computed using
dynamic programming and can then be used to construct $\jpol^{*}$.

Because calculating $Q^{*}$ is computationally expensive, we examined
approximate Q-value functions that can be calculated more efficiently
and we discussed how they relate to $Q^{*}$. We covered $\QMDP$,
$\QPOMDP$, and $\QBG$, a recently proposed approximate Q-value function.
 Also, we established that decreasing communication delays in decentralized
systems cannot decrease the expected value and thus that $Q^{*}\leq\QBG\leq\QPOMDP\leq\QMDP$.
Experimental evaluation indicated that these upper bounds are not
just of theoretical interest, but that significant differences exist
in the tightness of the various approximate Q-value functions.

Additionally we showed how the approximate Q-value functions can be
used as heuristics in a generalized policy search method $\GMAA$,
thereby unifying forward-sweep policy computation and the recent Dec-POMDP
solution techniques of \citet{Emery-Montemerlo04} and \citet{Szer05MAA}.
Finally, we performed an empirical evaluation of $\GMAA$ showing
significant reductions in computation time when using tighter heuristics
to calculate optimal policies. Also $\QBG$ generally found  better
approximate solutions in forward-sweep policy computation and the
`$k$-best joint BG policies' $\GMAA$ variant, or $\kGMAA$.

There are quite a few directions for future research. One is to try
to extend the results of this paper to partially observable stochastic
games (POSGs) \citep{Hansen04}, which are Dec-POMDPs with an individual
reward function for each agent. Since the dynamics of the POSG model
are identical to those of a Dec-POMDP, a similar modeling via Bayesian
games is possible. An interesting question is whether also in this
case, an optimal (i.e., rational) joint policy can be found by forward-sweep
policy computation. 

Staying within the context of Dec-POMDPs, a research direction could
be to further generalize $\GMAA$, by defining other $\PSO$ or $\selectO$
operators, with the hope that the resulting algorithms will be able
to scale to larger problems. Also it is important to establish bounds
on the performance and learning curves of $\GMAA$ in combination
with different $\PSO$ operators and heuristics. A different direction
is to experimentally evaluate the use of even tighter heuristics such
as Q-value functions for the case of observations delayed by multiple
time steps. This research should be paired with methods to efficiently
find such Q-value functions. Finally, future research should further
examine Bayesian games. In particular, the work of \citet{Emery-Montemerlo05}
could be used as a starting point for further research to approximately
modeling Dec-POMDPs using BGs. Finally, there is a need for efficient
approximate methods for solving the Bayesian games.

\section*{Acknowledgments}

We thank the anonymous reviewers for their useful comments. The research
reported here is part of the Interactive Collaborative Information
Systems (ICIS) project, supported by the Dutch Ministry of Economic
Affairs, grant nr: BSIK03024. This work was supported by   Funda\c{c}\~{a}o para a Ci\^encia e a Tecnologia (ISR/IST   pluriannual funding) through the POS\_Conhecimento Program that   includes FEDER funds, and through grant PTDC/EEA-ACR/73266/2006.

\appendix

\section{Proofs}

\subsection{There is At Least One Optimal Pure Joint Policy}

\label{sub:1optimalPureJP}

\begin{propositionRep}[\ref{prop:1optimalPureJP}] A Dec-POMDP has
at least one optimal pure joint policy.

\begin{proof} 
 This proof follows a proof by \citet{Schoute78}. It is possible
to convert a Dec-POMDP to an extensive game and thus to a strategic
game, in which the actions are pure policies for the Dec-POMDP \citep{Oliehoek06_TR_POSGs_extensiveform}.
In this strategic game, there is at least one maximizing entry corresponding
to a pure joint policy which we denote $\jppol_{\textrm{max}}$. Now,
assume that there is a joint stochastic policy $\jspol=\left\langle \spol1,\dots,\spol\nrA\right\rangle $
that attains a higher payoff. \citet{Kuhn53} showed that for each
stochastic $\spol i$ policy, there is a corresponding mixed policy~$\mpol i$.
Therefore $\jspol$ corresponds to a joint mixed policy $\jmpol=\left\langle \mpol1,\dots,\mpol\nrA\right\rangle $.
Let us write $\polS{i,\mpol i}$ for the support of $\mpol i$. $\jmpol$
now induces a probability distribution $\Pr_{\jmpol}$ over the set
of joint policies $\jpolS_{\jmpol}=\polS{1,\mpol1}\times\dots\times\polS{\nrA,\mpol\nrA}\subseteq\jpolS$
which is a subset of the set of all joint policies. The expected payoff
can now be written as\[
V(\jspol)=E_{\Pr_{\jmpol}}(V(\jpol)|\jpol\in\jpolS_{\jmpol})\leq\max_{\jpol\in\jpolS}V(\jpol)=V(\jppol_{\textrm{max}}),\]
contradicting that $\jspol$ is a joint stochastic policy that attains
a higher payoff.\end{proof}\end{propositionRep}

\subsection{Hierarchy of Q-value Functions}

\label{sub:Hierarchy-of-Q-value}

This section lists the proof of theorem \ref{thm:upperbounds}. It
is ordered as follows. First, Section \ref{sub:k-steps-delayed-com}
presents a model and resulting value functions for Dec-POMDPs with
$k$-steps delayed communication. Next, Section \ref{sub:Relation-k-steps-delayed-systems-and-Qvalue-functions}
shows that $\QPOMDP,$ $\QBG$ and $Q^{*}$ correspond with the case
that $k$ is respectively $0$, 1 and $h$. Finally, Section \ref{sub:Shorter-communication-delays}
shows that when the communication delay $k$ increases, the optimal
expected return cannot decrease, thereby proving theorem~\ref{thm:upperbounds}.

\subsubsection{Modeling Dec-POMDPs with $k$-Steps Delayed Communication}

\label{sub:k-steps-delayed-com}

\begin{figure}[t]
\newcommand{\thissize}{\small}{
\psfrag{a11}[cc][b]{\thissize $\aoo$}
\psfrag{a12}[cc][b]{\thissize $\aot$}
\psfrag{a21}[cc][b]{\thissize $\ato$}
\psfrag{a22}[cc][b]{\thissize $\att$}
\psfrag{o11}[cc][b]{\thissize $\ooo$}
\psfrag{o12}[cc][b]{\thissize $\oot$}
\psfrag{o21}[cc][b]{\thissize $\oto$}
\psfrag{o22}[cc][b]{\thissize $\ott$}
\psfrag{pol1}[cc][c]{\thissize {  $\qAT 1{\ts}$ } }
\psfrag{pol2}[cc][c]{\thissize {  $\qAT 2{\ts}$ } }
\psfrag{polBG}[cc][c]{\thissize $\jnT{\ts+k}$}
\psfrag{o12o21}[lc][b]{\thissize $\joT{\ts+1} = \langle\oot,\oto\rangle$}
\psfrag{pol1b}[cc][lc]{\thissize { $\qAT 1{\ts+1}$ }  }
\psfrag{pol2b}[cc][lc]{\thissize { $\qAT 2{\ts+1}$ } }
\psfrag{t-k}[cc][c]{\thissize $\ts$}
\psfrag{t-k+1}[cc][c]{\thissize $\ts+1$}
\psfrag{t-1}[cc][c]{\thissize $\ts+k-1$}
\psfrag{t}[cc][c]{\thissize $\ts+k$}

\noindent \begin{centering}
\includegraphics[width=0.8\columnwidth]{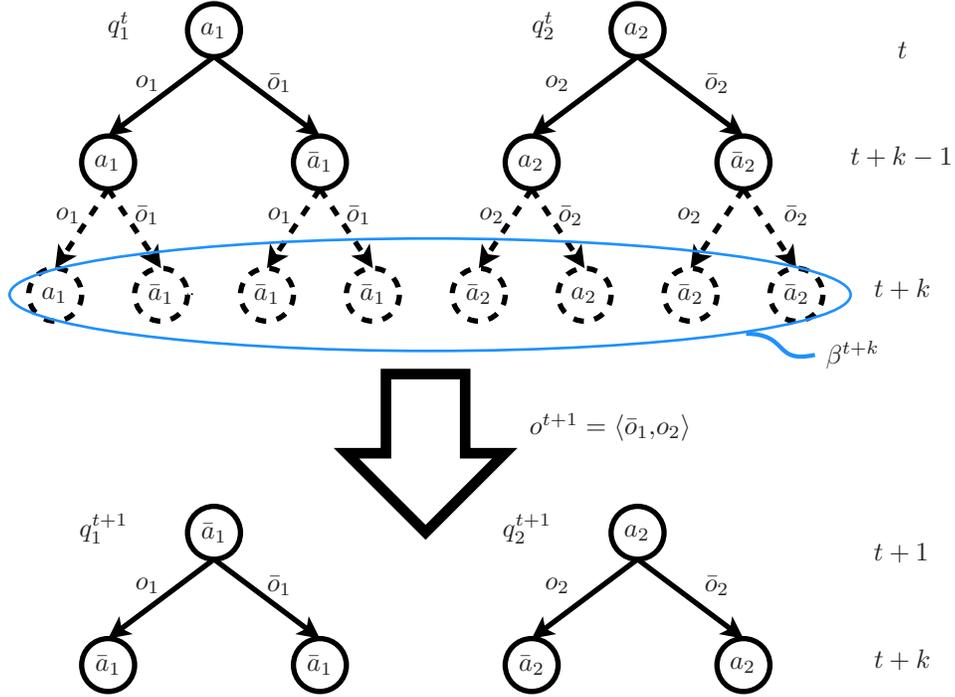}\\
~\\

\par\end{centering}

}

\caption{Policies specified by the states of the augmented MDP for $k=2$.
Top: policies for $\hat{s}^{\ts}$. The policy extended by augmented
MDP action $\hat{a}^{\ts}=\jpolBG$ is shown dashed. Bottom: The resulting
policies after for joint observation $\left\langle \oot,\oto\right\rangle $.}

\label{fig:Ooi_next_actions}
\end{figure}

Here we present an augmented MDP that can be used to find the optimal
solution for Dec-POMDPs with $k$ steps delayed communication. This
is a reformulation of the work by \citet{Aicardi87} and \citet{Ooi96},
extended to the Dec-POMDP setting. We define this augmented MDP as
$\hat{M}=\langle\hat{\mathcal{S}},\hat{\mathcal{A}},\hat{T},\hat{R}\rangle$,
where the augmented MDP stages are indicated $\ats$.

The state space is $\hat{\mathcal{S}}=(\hat{\mathcal{S}}^{\ats=0},\dots,\hat{\mathcal{S}}^{\ats=h-1})$.
An augmented state is composed of a joint action-observation history,
and a joint policy tree $\jqT\ts$.\[
\hat{s}^{\ats=\ts}=\begin{cases}
\langle\oaHistT{\ts},\jqT{\ts}\rangle & ,\,0\leq\ts\leq h-k-1\\
\langle\oaHistT{\ts},\jqTTG{h-\ts,\ts}\rangle & ,\, h-k\leq\ts\leq h-1\end{cases}.\]
The contained $\jqT\ts$ is a joint depth-$k$ (specifying actions
for $k$ stages) joint policy tree $\jqT{\ts}=\left\langle \qAT{1}{\ts},...,\qAT{\nrA}{\ts}\right\rangle $,
to be used starting at stage $\ts$. For the last $k$ stages, the
contained joint policy $\jqTTG{h-\ts,\ts}$ specifies $\ttg=h-\ts\leq k$
stages.

$\hat{\mathcal{A}}$ is the set of augmented actions. For $0\leq\ats\leq h-k-1$,
an action $\hat{a}^{\ats}\in\hat{\mathcal{A}}$ is a joint policy
$\hat{a}^{\ats=\ts}=\jnT{\ts+k}=\langle\nAT1{\ts+k}\dots\nAT\nrA{\ts+k}\rangle$
implicitly mapping length-$k$ observation histories to joint actions
to be taken at stage $\ts+k$. I.e., $\nAT i{\ts+k}:\oHistATS ik\rightarrow\aAS i^{{\ts+k}}$.
For the last $k$ stages $h-k\leq\ats\leq h-1$ there only is one
empty action $\aEmpty$ that has no influence whatsoever. 

The augmented actions are used to expand the joint policy trees. When
`appending' a policy $\jnT{\ts+k}$ to $\jqT{\ts}$ we form a depth
$k+1$ policy, which we denote $\jqTTG{k+1,\ts}=\cat{\jqT{\ts}}{\jnT{\ts+k}}$.
After execution of its initial joint action $\jqTTG{k+1,\ts}(\oHistEmpty)$
and receiving a particular joint observation~$\jo$, a $\jqTTG{k+1,\ts}$
reduces to its depth $k$ sub-tree policy for that particular joint
observation, denoted $\jqT{\ts-k+1}=\jqTTG{k+1,\ts}(\jo)=\cat{\jqT{\ts-k}}{\jnT{\ts}}(\jo)$.
This is illustrated in Figure \ref{fig:Ooi_next_actions}.%
\begin{figure}
\newcommand{\thissize}{\small}{\psfrag{a11}[cc][b]{\thissize $\aoo$}
\psfrag{a12}[cc][b]{\thissize $\aot$}
\psfrag{a21}[cc][b]{\thissize $\ato$}
\psfrag{a22}[cc][b]{\thissize $\att$}
\psfrag{o11}[cc][b]{\thissize $\ooo$}
\psfrag{o12}[cc][b]{\thissize $\oot$}
\psfrag{o21}[cc][b]{\thissize $\oto$}
\psfrag{o22}[cc][b]{\thissize $\ott$}

\psfrag{hata}[bl][bl]{\thissize {$\joT{\ts+1}=\langle\ooo,\oto\rangle$} } 
\psfrag{s^t-k}[cl][bl]{\thissize $\oaHistT{t},$}
\psfrag{s^t+1-k}[cl][bl]{\thissize $\oaHistT{t+1},$}
\psfrag{hst}[cc][c]{\thissize $\hat{s}^{\ts}$}
\psfrag{hst+}[cc][c]{\thissize $\hat{s}^{\ts+1}$}

\psfrag{t-k}[cc][b]{\thissize $\ts$}
\psfrag{t-1}[cc][b]{\thissize $\ts+1$}
\psfrag{(t+1)-k}[cc][b]{\thissize $\ts+1$}
\psfrag{(t+1)-1}[cc][b]{\thissize $\ts+k$}

\noindent \begin{centering}
\includegraphics[scale=0.72]{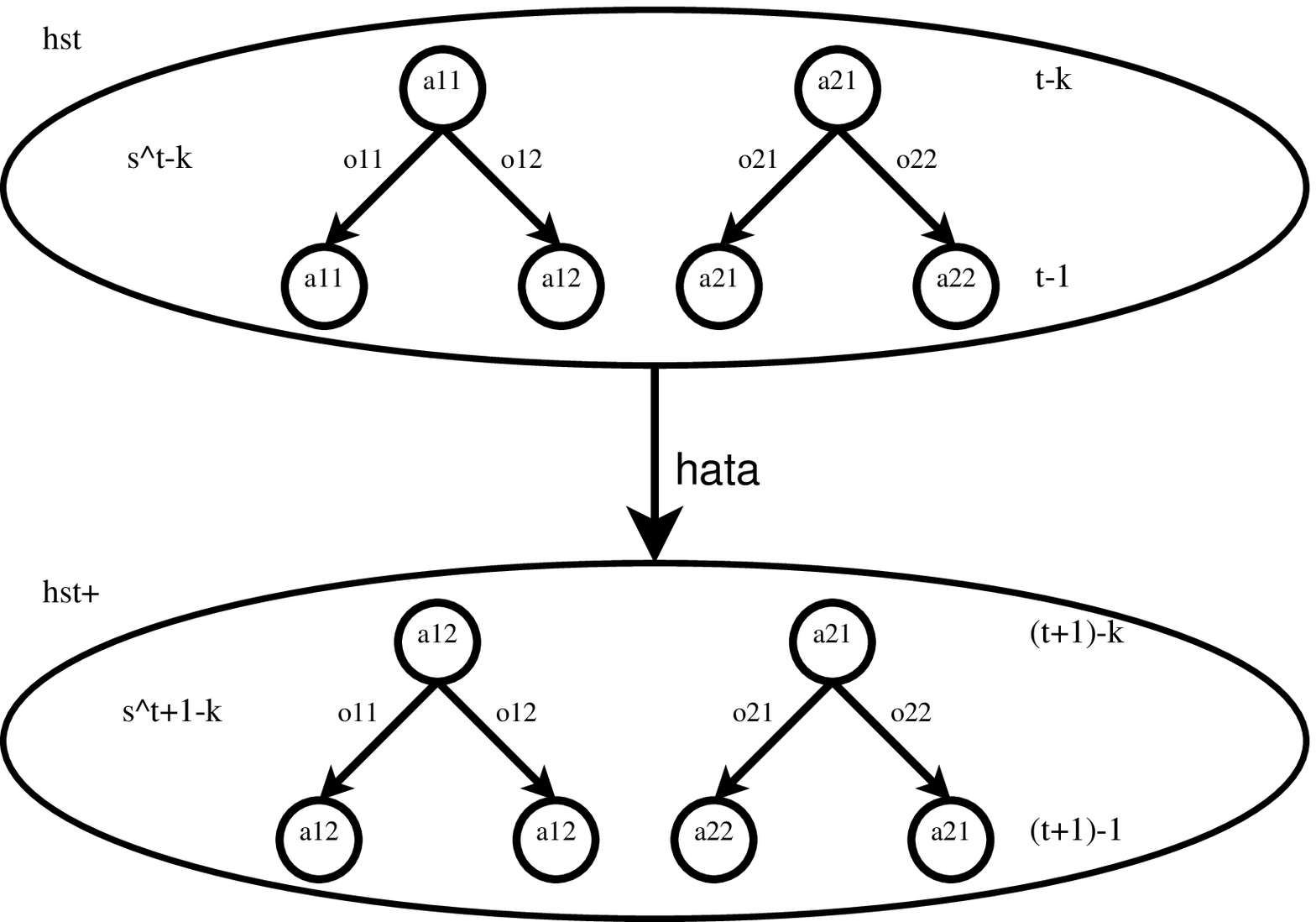}
\par\end{centering}

}

\caption{An illustration of the augmented MDP with $k=2$, showing a transition
from $\hat{s}^{\ts}$ to $\hat{s}^{\ts+1}$ by action $\hat{a}=\jnT\ts$.
In this example $\oaHistT{\ts-k+1}=(\oaHistT{\ts-k+1},\left\langle \aoo,\ato\right\rangle ,\left\langle \oot,\oto\right\rangle )$.
The actions specified for stage $\ts$ are given by $\jnT\ts(\left\langle \oot,\oto\right\rangle )$
as depicted in Figure~\ref{fig:Ooi_next_actions}.}

\label{fig:Ooi_augmented_MDP}
\end{figure}

$\hat{T}$ is the transition model. A probability $\Pr(\hat{s}^{\ats+1}|\hat{s}^{\ats},\hat{a}^{\ats})$
for stage $\ats=\ts$ translates as follows for $0\leq\ts\leq h-k-1$\begin{equation}
\Pr(\langle\oaHistT{\ts+1},\jqT{\ts+1}\rangle|\langle\oaHistT{\ts},\jqT{\ts}\rangle,\jnT{\ts+k})=\begin{cases}
\Pr(\joT{\ts+1}|\oaHistT\ts,\jqT{\ts}(\oHistEmpty)) & \textrm{if conditions hold,}\\
0 & \textrm{otherwise,}\end{cases}\label{eq:augMDP:T}\end{equation}
where the conditions are: 1) $\jqT{\ts+1}=\cat{\jqT{\ts}}{\jnT{\ts+k}}(\joT{\ts+1})$,
and 2) $\oaHistT{\ts+1}=(\oaHistT{\ts},\ja^{\ts},\jo^{\ts+1})$. For
$h-k\leq\ts\leq h-1$, $\jnT{\ts+k}$ in \eqref{eq:augMDP:T} reduces
to $\aEmpty$. The probabilities are unaffected, but the first condition
changes to $\jqTTG{h-\ts-1,\ts+1}=\jqTTG{h-\ts,\ts}(\joT{\ts+1})$.

Finally, $\hat{R}$ is the reward model, which is specified as follows:
\begin{equation}
\forall_{0\leq\ts\leq h-1}\quad\hat{R}(\hat{s}^{\ats=\ts})=\hat{R}(\langle\oaHistT{\ts},\jqT{\ts}\rangle)=R(\oaHistT{\ts},\jqT{\ts}(\oHistEmpty)),\label{eq:kQBG:def:2:R}\end{equation}
where $\jqT{\ts}(\oHistEmpty)$ is the initial joint action specified
by $\jqT{\ts}$. $R(\oaHistT\ts,\ja)$ is defined as before in \eqref{eq:exp_imm_R_R(oaHist,ja)}.

The resulting optimality equations $\hat{Q}^{\ats}(\hat{s},\hat{a})$
for the augmented MDP are as follows. We will write $\kQ$ for the
optimal Q-value function for a $k$-steps delayed communication system.
We will also refer to this as the $\kQBG$ value function. \begin{equation}
\forall_{0\leq\ts\leq h-k-1}\quad\kQ(\oaHistT{\ts},\jqT{\ts},\jnT{\ts+k})=R(\oaHistT{\ts},\jqT{\ts}(\oHistEmpty))+\sum_{\jo^{\ts+1}\in\mathcal{O}}\Pr(\jo^{\ts+1}|\oaHistT{\ts},\jqT{\ts}(\oHistEmpty))\kQ^{*}(\oaHistT{\ts+1},\jqT{\ts+1}),\label{eq:oaH-delay:Q^ts}\end{equation}
with $\jqT{\ts+1}=\cat{\jqT{\ts}}{\jnT{\ts+k}}(\jo^{\ts+1})$ and
where\begin{equation}
\kQ^{*}(\oaHistT{\ts},\jqT{\ts})\defas\max_{\jnT{\ts+k}}\kQ(\oaHistT{\ts},\jqT{\ts},\jnT{\ts+k}).\label{eq:oaH-delay:Q*^t_ismax_Q^t}\end{equation}
For the last $k$ stages, $h-k\leq\ts\leq h-1$, there are $\ttg'=h-\ts$
stages to go and we get\begin{equation}
\kQ^{*}(\oaHistT{\ts},\jqTTG{\ttg',\ts})=R(\oaHistT{\ts},\jqTTG{\ttg',\ts}(\oHistEmpty))+\sum_{\jo^{\ts+1}}\Pr(\jo^{\ts+1}|\oaHistT{\ts},\jqTTG{\ttg',\ts}(\oHistEmpty))\kQ^{*}(\oaHistT{\ts+1},\jqTTG{\ttg'-1,\ts+1}).\label{eq:Q^(ts_geq_(h-k))}\end{equation}
Note that \eqref{eq:Q^(ts_geq_(h-k))} does not include any augmented
actions $\hat{a}^{\ats=\ts}=\jnT{\ts+k}$. Therefore, the last $k$
stages should be interpreted as a Markov chain. Standard dynamic programming
can be applied to calculate all $Q^{*}(\oaHistT{\ts},\jqT{\ts})$-values.

\subsubsection{Relation of $\kQBG$ with Other Approximate Q-value Functions}

\label{sub:Relation-k-steps-delayed-systems-and-Qvalue-functions}

Here we briefly show how $\kQBG$ in fact reduces to some of the cases
treated earlier.

For $k=0$, $\kQBG$ \eqref{eq:oaH-delay:Q^ts} reduces to $\QPOMDP$.
In the $k=0$ case, $\jqT{\ts-k}$ becomes a depth-0, i.e.\ empty,
policy. Also, $\jnT\ts$ becomes a mapping from length-0 observation
histories to actions, i.e., it becomes a joint action. Substitution
in \eqref{eq:oaH-delay:Q^ts} yields\[
\kQk0(\left\langle \oaHistT{\ts},\emptyset\right\rangle ,\jaT\ts)=R(\oaHistT{\ts},\jaT\ts)+\sum_{\jo^{\ts+1}}\Pr(\jo^{\ts+1}|\oaHistT{\ts},\jaT{\ts})\max_{\ja^{\ts+1}}\kQk0(\left\langle \oaHistT{\ts+1},\emptyset\right\rangle ,\ja^{\ts+1}).\]
Now, as $\QPOMDPQH(\oaHistT{\ts},\ja)\defas\QPOMDPQ^{*}(b^{\oaHistT{\ts}},\ja),$
this clearly corresponds to the $\QPOMDP$-value function \eqref{eq:QPOMDP}.

$\kQBGk1$ reduces to regular $\QBG$. Notice that for $k=1,$ $\stJPolT{k,\ts}$
reduces to $\ja^{\ts}$. Filling out yields:\[
\kQk1(\left\langle \oaHistT{\ts},\jaT{\ts}\right\rangle ,\jnT{\ts+1})=R(\oaHistT{\ts},\jaT{\ts})+\sum_{\jo^{\ts+1}}\Pr(\jo^{\ts+1}|\oaHistT{\ts},\jaT{\ts})\max_{\jnT{\ts+2}}\kQk1(\left\langle \oaHistT{\ts+1},\jnT{\ts+1}(\jo^{\ts+1})\right\rangle ,\jnT{\ts+2}).\]
Now using \eqref{eq:oaH-delay:Q*^t_ismax_Q^t} we obtain the $\QBG$-value
function \eqref{eq:QBG}. 

A Dec-POMDP is identical to an $h$-steps delayed communication system.
Augmented states have the form $\hat{s}^{\ats=0}=\langle\oaHistEmpty,\jqT0\rangle$,
where $\jqT{0}=\jpol$ specifies a full length $h$ joint policy.
The first stage $\ts=0$ in this augmented MDP, is also one of the
last $k$ $(=h)$ stages. Therefore, the applied $Q$-function is
\eqref{eq:Q^(ts_geq_(h-k))}, which means that the Markov chain evaluation
starts immediately. Effectively this boils down to evaluation of all
joint policies (corresponding to all augmented start states). The
maximizing one specifies the value function of an optimal joint policy
$Q^{*}.$

\subsubsection{Shorter Communication Delays cannot Decrease the Value}

\label{sub:Shorter-communication-delays}

\providecommand{\jnTL}[2]{\jnT{#1}_{|#2|}}

First, we introduce some notation. Let us write $\Po$ for all the
observation probabilities given $\oaHistT{\ts},\jqT{\ts}$ and the
sequence of `intermediate observations' $(\jo^{\ts+1},\dots,\jo^{\ts+l-1})$
\begin{equation}
\Po(\jo^{\ts+l})\defas\Pr\left[\jo^{\ts+l}|(\oaHistT{\ts},\jqT{\ts}(\oHistEmpty),\jo^{\ts+1},\jqT{\ts}(\jo^{\ts+1}),\dots,\jo^{\ts+l-1}),\jqT{\ts}(\jo^{\ts+l-1})\right],\qquad\forall l\leq k.\label{eq:def:Po}\end{equation}

In order to avoid confusion, we write $\jnTL{\ts}{k}$ for a policy
that implicitly maps $k$-length observation histories to actions,
and $\jnTL{\ts}{k+1}$ for one that is a mapping from length $(k+1)$
observation-histories to actions. 

Now we give a reformulation of $\kQ$. $\kQ^{\ts,*}(\oaHistT\ts,\jqT{\ts})$
specifies the expected return for $\oaHistT\ts,\jqT{\ts}$ over stages
$\ts,\ts+1,\dots,h-1$. Here, we will split this\begin{equation}
\kQ^{*}(\oaHistT\ts,\jqT{\ts})=\kQK(\oaHistT\ts,\jqT{\ts})+\kQF^{\ts,*}(\oaHistT\ts,\jqT{\ts})\label{eq:Q_is_QK+QF}\end{equation}
in $\kQK(\oaHistT\ts,\jqT{\ts})$, the \emph{expected $k$-step reward},
i.e., the expected return over for stages $\ts,\dots,\ts+k-1$ and
$\kQF^{\ts,*}(\oaHistT\ts,\jqT{\ts})$, the expected return over stages
$\ts+k,\ts+k+1,\dots,h-1$, referred to as the `in $k$-steps' expected
return.

The former is defined as

\begin{equation}
\kQK(\oaHistT\ts,\jqT{\ts})\defas E\left[\sum_{\ts'=\ts}^{\ts+k-1}R(\oaHistT{\ts'},\jaT{\ts'})\,\Big{|}\,\oaHistT\ts,\jqT{\ts}\right].\label{eq:kQK:def}\end{equation}
Let us define $\QK^{\ttg=i}(\oaHistT\ts,\jqTTG{i,\ts})$ as the expected
reward for the next $i$ stages, i.e., \begin{equation}
\kQK(\oaHistT\ts,\jqT{\ts})=\QK^{\ttg=k}(\oaHistT\ts,\jqT{\ts}).\label{eq:kQK_is_QK_TTGk}\end{equation}
We then have $\QK^{\ttg=1}(\oaHistT\ts,\ja^{\ts})=R(\oaHistT{\ts},\ja^{\ts})$
and\begin{multline}
\QK^{\ttg=i}(\oaHistT\ts,\jqTTG{i,\ts})=R(\oaHistT{\ts},\jqTTG{i,\ts}(\oHistEmpty))+\label{eq:QK_TTGi}\\
\sum_{\jo^{\ts+1}}\Pr(\jo^{\ts+1}|\oaHistT\ts,\jqTTG{i,\ts}(\oHistEmpty))\QK^{\ttg=i-1}(\oaHistT{\ts+1},\jqTTG{i-1,\ts+1}(\jo^{\ts+1})),\end{multline}
where $\jqTTG{i-1,\ts+1}(\jo^{\ts+1})$ is the depth-$(i-1)$ joint
policy that results from $\jqTTG{i,\ts}$ after observation of $\jo^{\ts+1}$.

\paragraph{}

\label{sub:kQF:The-in-k-steps-exp-return}

If we define $\kQF^{\ttg=i,\ts}(\oaHistT{\ts},\jqT{\ts},\jnTL{\ts+k}{k})$
to be the expected reward for stages $\ts+i,\ts+i+1,\dots,h-1$. That
is, the time-to-go $\ttg=i$ denotes how much time-to-go before we
start accumulating expected reward. The `in $k$-steps' expected
return is then given by\[
\kQF^{\ts}(\oaHistT{\ts},\jqT{\ts},\jnTL{\ts+k}{k})=\kQF^{\ttg=k,\ts}(\oaHistT{\ts},\jqT{\ts},\jnTL{\ts+k}{k}).\]
  The evaluation is then performed by\begin{eqnarray}
\kQF^{\ttg=0,\ts}(\oaHistT{\ts},\jqT{\ts},\jnTL{\ts+k}{k}) & = & \kQ(\oaHistT{\ts},\jqT{\ts},\jnTL{\ts+k}{k})\label{eq:QF^ttg0}\\
\kQF^{\ttg=i,\ts}(\oaHistT{\ts},\jqT{\ts},\jnTL{\ts+k}{k}) & = & \sum_{\joT{\ts+1}}\Po(\joT{\ts+1})\kQF^{\ttg=i-1,\ts+1,*}(\oaHistT{\ts+1},\jqT{\ts+1}),\label{eq:QF^ttgi}\end{eqnarray}
where $\jqT{\ts+1}=\cat{\jqT{\ts+1}}{\jnTL{\ts+k}{k}}(\joT{\ts+1})$,
and where \begin{equation}
\kQF^{\ttg=i,\ts,*}(\oaHistT{\ts},\jqT{\ts})=\max_{\jnTL{\ts+k}{k}}\kQF^{\ttg=i,\ts}(\oaHistT{\ts},\jqT{\ts},\jnTL{\ts+k}{k}).\label{eq:QF^ttgi*}\end{equation}

%
{}

\begin{theorem}[Shorter communication delays cannot decrease the value]The
optimal Q-value function $Q_{k}$ of a finite horizon Dec-POMDP with
$k$-steps delayed communication is an upper bound to $Q_{k+1}$,
that of a $k+1$-steps delayed communication system. That is\begin{equation}
\forall_{\ts}\forall_{\oaHistT\ts}\forall_{\jqTTG{k,\ts},\jnTL{\ts+k}k}\quad Q_{k}(\oaHistT{\ts},\jqTTG{k,\ts},\jnTL{\ts+k}k)\geq\max_{\jnTL{\ts+k+1}{k+1}}Q_{k+1}(\oaHistT{\ts},\cat{\jqTTG{k,\ts}}{\jnTL{\ts+k}k},\jnTL{\ts+k+1}{k+1}).\label{eq:UBproof:toProof}\end{equation}

\begin{proof}  The proof is by induction. The base case is that \eqref{eq:UBproof:toProof}
holds for stages $h-(k+1)\leq\ts\leq h-1$, as shown by lemma \ref{lem:UBproof:base-case}.
The induction hypothesis states that, assuming \eqref{eq:UBproof:toProof}
holds for some stage $\ts+k$, it also holds for stage~$\ts$. The
induction step is proven in lemma \ref{lem:UPproof:induction-step}.
\end{proof} \end{theorem}

%
{}

\begin{lemma}[Base case]  \label{lem:UBproof:base-case} For all
$h-k-1\leq\ts\leq h-1$, the expected cumulative future reward under
$k$ steps delay is equal to that under $k+1$ steps delay if the
same policies are followed from that point. That is,\begin{equation}
\forall_{h-k\leq\ts\leq h-1}\forall_{\oaHistT{\ts}}\forall_{\jqTTG{h-\ts,\ts}}\quad\kQ^{*}(\oaHistT{\ts},\jqTTG{h-\ts,\ts})=Q_{k+1}^{*}(\oaHistT{\ts},\jqTTG{h-\ts,\ts}),\label{eq:lem:UBproof:base-case:ToProof:Q^ts}\end{equation}
and $\forall_{\oaHistT{h-k-1}}\forall_{\jqTTG{k,h-k-1},\jnTL{h-1}k}$\begin{equation}
Q_{k}(\oaHistT{h-k-1},\jqTTG{k,h-k-1},\jnTL{h-1}k)=Q_{k+1}(\oaHistT{h-k-1},\cat{\jqTTG{k,h-k-1}}{\jnTL{h-1}k}).\label{eq:lem:UBproof:base-case:ToProof:Q^h-(k+1)}\end{equation}

\proofup For a particular stage $\ts=h-\ttg'$ with $h-k\leq\ts\leq h-1$
and an arbitrary $\oaHistT{\ts},\jqTTG{\ttg',\ts}$, we can write\[
Q_{k}^{*}(\oaHistT{\ts},\jqTTG{\ttg',\ts})=Q_{k+1}^{*}(\oaHistT{\ts},\jqTTG{\ttg',\ts}),\]
because both are given by the evaluation of \eqref{eq:Q^(ts_geq_(h-k))},
and this evaluation involves no actions: Basically \eqref{eq:Q^(ts_geq_(h-k))}
has reduced to a Markov chain, and this Markov chain is the same for
$Q_{k}^{*}$ and $Q_{k+1}^{*}$.  We can conclude that \[
\forall_{h-k\leq\ts\leq h-1}\forall_{\oaHistT{\ts},\jqTTG{\ttg',\ts}}\quad Q_{k}^{*}(\oaHistT{\ts},\jqTTG{\ttg',\ts})=Q_{k+1}^{*}(\oaHistT{\ts},\jqTTG{\ttg',\ts}).\]

Now we will prove \eqref{eq:lem:UBproof:base-case:ToProof:Q^h-(k+1)}.
The left side of \eqref{eq:lem:UBproof:base-case:ToProof:Q^h-(k+1)}
is given by application of \eqref{eq:oaH-delay:Q^ts} \begin{multline*}
Q_{k}(\oaHistT{h-k-1},\jqTTG{k,h-k-1},\jnTL{h-1}k)=R(\oaHistT{h-k-1},\jqTTG{k,h-k-1}(\oHistEmpty))+\\
\sum_{\jo^{h-k}}\Pr(\jo^{h-k}|\oaHistT{h-k-1},\jqTTG{k,h-k-1}(\oHistEmpty))Q_{k}^{*}(\oaHistT{h-k},\jqTTG{k,h-k}),\end{multline*}
with $\jqTTG{k,h-k}=\cat{\jqTTG{k,h-k-1}}{\jnTL{h-1}k}(\jo^{h-k})$.
The right side is given by application of \eqref{eq:Q^(ts_geq_(h-k))}\begin{multline*}
Q_{k+1}(\oaHistT{h-k-1},\cat{\jqTTG{k,h-k-1}}{\jnTL{h-1}k})=R(\oaHistT{h-k-1},\jqTTG{k,h-k-1}(\oHistEmpty))+\\
\sum_{\jo^{h-k}}\Pr(\jo^{h-k}|\oaHistT{h-k-1},\jqTTG{k,h-k-1}(\oHistEmpty))Q_{k+1}^{*}(\oaHistT{h-k},\jqTTG{k,h-k})\end{multline*}
with $\jqTTG{k,h-k}=\cat{\jqTTG{k,h-k-1}}{\jnTL{h-1}k}(\jo^{h-k})$.
Now, because the policies $\jqTTG{k,h-k}$ are the same, we get \[
Q_{k}^{*}(\oaHistT{h-k},\jqTTG{k,h-k})=Q_{k+1}^{*}(\oaHistT{h-k},\jqTTG{k,h-k})\]
and thus \eqref{eq:lem:UBproof:base-case:ToProof:Q^h-(k+1)} holds.\end{proof} \end{lemma}

%
{}

\begin{lemma}[Induction step] \label{lem:UPproof:induction-step}
Given that \begin{equation}
\forall_{\oaHistT{\ts'}}\forall_{\jqTTG{k,\ts'},\jnTL{\ts'+k}{k}}\quad Q_{k}(\oaHistT{\ts'},\jqTTG{k,\ts'},\jnTL{\ts'+k}{k})\geq\max_{\jnTL{\ts'+k+1}{k+1}}Q_{k+1}(\oaHistT{\ts'},{\cat{\jqTTG{k,\ts'}}{\jnTL{\ts'+k}{k}}},\jnTL{\ts'+k+1}{k+1})\label{eq:UBproof:IS:Assumption}\end{equation}
 holds for ${\ts'}=\ts+(k+1)$, then\begin{equation}
\forall_{\oaHistT\ts}\forall_{\jqTTG{k,\ts},\jnTL{\ts+k}{k}}\quad Q_{k}(\oaHistT{\ts},\jqTTG{k,\ts},\jnTL{\ts+k}{k})\geq\max_{\jnTL{\ts+k+1}{k+1}}Q_{k+1}(\oaHistT{\ts},{\cat{\jqTTG{k,\ts}}{\jnTL{\ts+k}{k}}},\jnTL{\ts+k+1}{k+1})\label{eq:UBproof:IS:toProof}\end{equation}
 holds for stage $\ts$. 

\proofup  For the $k$-steps delay Q-function, we can write\begin{multline}
\kQ(\oaHistT{\ts},\jqTTG{k,\ts},\jnTL{\ts+k}k)=R(\oaHistT{\ts},\jqTTG{k,\ts}(\oHistEmpty))+\\
\sum_{\jo^{\ts+1}}\Po(\jo^{\ts+1}|\oaHistT{\ts},\jqTTG{k,\ts}(\oHistEmpty))\max_{\jnTL{\ts+k+1}k}\left[\kQK(\oaHistT{\ts+1},\jqT{\ts+1})+\kQF^{\ts+1}(\oaHistT{\ts+1},\jqTTG{k,\ts+1},\jnTL{\ts+k+1}k)\right]\label{eq:UBproof:Q_k}\end{multline}
where $\jqTTG{k,\ts+1}=\cat{\jqTTG{k,\ts}}{\jnTL{\ts+k}k}(\joT{\ts+1})$.
Because $\kQK$ is independent of $\jnTL{\ts+k+1}k$, we can regroup
the terms to get\begin{multline}
\kQ(\oaHistT{\ts},\jqTTG{k,\ts},{\jnTL{\ts+k}k})=\left[R(\oaHistT{\ts},\jqTTG{k,\ts}(\oHistEmpty))+\sum_{\jo^{\ts+1}}\Po(\jo^{\ts+1})\kQK(\oaHistT{\ts+1},\jqTTG{k,\ts+1})\right]+\\
\left[\sum_{\jo^{\ts+1}}\Po(\jo^{\ts+1})\max_{\jnTL{\ts+k+1}k}\kQF^{\ts+1}(\oaHistT{\ts+1},\jqTTG{k,\ts+1},\jnTL{\ts+k+1}k)\right].\label{eq:UBproof:Q_k-regroup}\end{multline}

In the case of $k+1$-steps delay, we can write\begin{equation}
\kQk{k+1}(\oaHistT{\ts},\cat{\jqTTG{k,\ts}}{\jnTL{\ts+k}{k}},{\jnTL{\ts+k}{k+1}})=\kQKk{k+1}(\oaHistT{\ts},\cat{\jqTTG{k,\ts}}{\jnTL{\ts+k}{k}})+\kQFk{k+1}^{\ts}(\oaHistT{\ts},\cat{\jqTTG{k,\ts}}{\jnTL{\ts+k}{k+1}},{\jnTL{\ts+k+1}{k+1}})\label{eq:UBproof:Q_k+1}\end{equation}
where, per definition (by \eqref{eq:kQK_is_QK_TTGk} and \eqref{eq:QK_TTGi})
\begin{align}
\kQKk{k+1}(\oaHistT{\ts},\cat{\jqTTG{k,\ts}}{\jnTL{\ts+k}{k}}) & =R(\oaHistT{\ts},\jqTTG{k,\ts}(\oHistEmpty))+\sum_{\jo^{\ts+1}}\Po(\jo^{\ts+1})\QK^{\ttg=k}(\oaHistT{\ts+1},\jqTTG{k,\ts+1}),\nonumber \\
 & =R(\oaHistT{\ts},\jqTTG{k,\ts}(\oHistEmpty))+\sum_{\jo^{\ts+1}}\Po(\jo^{\ts+1})\kQK(\oaHistT{\ts+1},\jqTTG{k,\ts+1}),\label{eq:UBproof:IS:K_(k+1)}\end{align}
where $\jqTTG{k,\ts+1}=\cat{\jqTTG{k,\ts}}{\jnT{\ts+k}}(\joT{\ts+1})$. 

Equation \eqref{eq:UBproof:IS:K_(k+1)} is equal to the first part
in \eqref{eq:UBproof:Q_k-regroup}. Therefore, for an arbitrary $\oaHistT{\ts},\jqTTG{k,\ts}$
and $\jnTL{\ts+k}{k}$, we know that \eqref{eq:UBproof:IS:toProof}
holds if and only if\begin{equation}
\sum_{\jo^{\ts+1}}\Po(\jo^{\ts+1})\max_{{\jnTL{\ts+k+1}{k}}}\QF_{k}^{\ts+1}(\oaHistT{\ts+1},\jqTTG{k,\ts+1},{\jnTL{\ts+k+1}{k}})\geq\max_{\jnTL{\ts+k+1}{k+1}}\QF_{k+1}^{\ts}(\oaHistT{\ts},\cat{\jqTTG{k,\ts}}{\jnTL{\ts+k}{k}},\jnTL{\ts+k+1}{k+1})\label{eq:UBproof:IS:TP_1a}\end{equation}
where $\jqTTG{k,\ts+1}=\cat{\jqTTG{k,\ts}}{\jnTL{\ts+k}{k}}(\joT{\ts+1})$.
When filling this out and expanding $\QF_{k+1}^{\ts}$ using \eqref{eq:QF^ttgi}
we get\begin{multline}
\sum_{\jo^{\ts+1}}\Po(\jo^{\ts+1})\max_{{\jnTL{\ts+k+1}{k}}}\QF_{k}^{\ttg=k,\ts+1}(\oaHistT{\ts+1},{\cat{\jqTTG{k,\ts}}{\jnTL{\ts+k}{k}}(\joT{\ts+1})},{\jnTL{\ts+k+1}{k}})\geq\\
\max_{\jnTL{\ts+k+1}{k+1}}\sum_{\jo^{\ts+1}}\Po(\jo^{\ts+1})\QF_{k+1}^{\ttg=k,\ts+1,*}(\oaHistT{\ts+1},{\cat{\cat{\jqTTG{k,\ts}}{\jnTL{\ts+k}{k}}}{\jnTL{\ts+k+1}{k+1}}(\joT{\ts+1})}).\label{eq:UBproof:IS:TP_1}\end{multline}
This clearly holds if\begin{multline}
\sum_{\jo^{\ts+1}}\Po(\jo^{\ts+1})\max_{{\jnTL{\ts+k+1}{k}}}\QF_{k}^{\ttg=k,\ts+1}(\oaHistT{\ts+1},{\cat{\jqTTG{k,\ts}}{\jnTL{\ts+k}{k}}(\joT{\ts+1})},{\jnTL{\ts+k+1}{k}})\geq\\
\sum_{\jo^{\ts+1}}\Po(\jo^{\ts+1})\max_{\jnTL{\ts+k+1}{k}}\QF_{k+1}^{\ttg=k,\ts+1,*}(\oaHistT{\ts+1},{\cat{{\cat{\jqTTG{k,\ts}}{\jnTL{\ts+k}{k}}}(\joT{\ts+1})}{\jnTL{\ts+k+1}{k}}}),\label{eq:UBproof:IS:TP_2}\end{multline}
because the second part of \eqref{eq:UBproof:IS:TP_2} is an upper
bound to the second part of \eqref{eq:UBproof:IS:TP_1}. Therefore,
the induction step is proved if we can show that \begin{multline}
\forall_{\joT{\ts+1}}\forall_{{\jnTL{\ts+k+1}{k}}}\quad\QF_{k}^{\ttg=k,\ts+1}(\oaHistT{\ts+1},{\cat{\jqTTG{k,\ts}}{\jnTL{\ts+k}{k}}(\joT{\ts+1})},{\jnTL{\ts+k+1}{k}})\geq\\
\QF_{k+1}^{\ttg=k,\ts+1,*}(\oaHistT{\ts+1},{\cat{{\cat{\jqTTG{k,\ts}}{\jnTL{\ts+k}{k}}}(\joT{\ts+1})}{\jnTL{\ts+k+1}{k}}}).\label{eq:UBproof:IS:TP_3}\end{multline}
which through \eqref{eq:QF^ttgi*} and $\jqTTG{k,\ts+1}=\cat{\jqTTG{k,\ts}}{\jnTL{\ts+k}{k}}(\joT{\ts+1})$
transforms to \begin{multline}
{\forall_{\jqTTG{k,\ts+1}}\forall}_{{\jnTL{\ts+k+1}{k}}}\quad\QF_{k}^{\ttg=k,\ts+1}(\oaHistT{\ts+1},\jqTTG{k,\ts+1},{\jnTL{\ts+k+1}{k}})\geq\label{eq:UBproof:IS:TP_4}\\
\max_{\jnTL{\ts+k+2}{k+1}}\QF_{k+1}^{\ttg=k,\ts+1}(\oaHistT{\ts+1},{\cat{\jqTTG{k,\ts+1}}{\jnTL{\ts+k+1}{k}}},\jnTL{\ts+k+2}{k+1}).\end{multline}
Now, we apply \eqref{eq:QF^ttg0} to the induction hypothesis \eqref{eq:UBproof:IS:Assumption}
and yield\begin{equation}
\forall_{\oaHistT{\ts'}}\forall_{\jqTTG{k,\ts'},\jnTL{\ts'+k}{k}}\quad\QF_{k}^{\ttg=0,\ts'}(\oaHistT{\ts'},\jqTTG{k,\ts'},\jnTL{\ts'+k}{k})\geq\max_{\jnTL{\ts'+k+1}{k+1}}\QF_{k+1}^{\ttg=0,\ts'}(\oaHistT{\ts'},{\cat{\jqTTG{k,\ts'}}{\jnTL{\ts'+k}{k}}},\jnTL{\ts'+k+1}{k+1}).\label{eq:UPproof:IS:IH_transformed}\end{equation}
Application of lemma \ref{lem:QF_k_>_QF_k+1_TTG>i} to this transformed
induction hypothesis asserts \eqref{eq:UBproof:IS:TP_4} and thereby
proves the lemma. \end{proof} \end{lemma}

\paragraph{Auxiliary Lemmas.}

\begin{lemma} \label{lem:QF_k_>_QF_k+1_TTGi} If, at stage $\ts$,
the `in $i$-steps' expected return for a $k$-steps delayed system
is higher than a $(k+1)$-steps delayed system, then at $\ts-1$ the
`in $(i+1)$-steps' expected return for a $k$-steps delayed system
is higher than the $(k+1)$-steps delayed system. That is, if for
a particular ${\jqTTG{k,\ts}}={\cat{\jqTTG{k,\ts-1}}{\jnTL{\ts-1+k}k}(\joT{\ts})}$
\begin{multline}
\forall_{\jnTL{\ts+k}k}\forall_{\joT{\ts}}\quad\QF_{k}^{\ttg=i,\ts}(\oaHistT{\ts},{\jqTTG{k,\ts}},\jnTL{\ts+k}k)=\QF_{k}^{\ttg=i,\ts}(\oaHistT{\ts},{\cat{\jqTTG{k,\ts-1}}{\jnTL{\ts-1+k}k}(\joT{\ts})},\jnTL{\ts+k}k)\geq\\
\max_{\jnTL{\ts+k+1}{k+1}}\QF_{k+1}^{\ttg=i,\ts}(\oaHistT{\ts},{\cat{{\cat{\jqTTG{k,\ts-1}}{\jnTL{\ts+k}k}}(\joT{\ts})}{\jnTL{\ts+k}k}},\jnTL{\ts+k+1}{k+1})\label{eq:lem:QF_k_>_QF_k+1_TTGi:condition}\end{multline}
holds, then\begin{equation}
\QF_{k}^{\ttg=i+1,\ts-1}(\oaHistT{\ts-1},\jqTTG{k,\ts-1},\jnTL{\ts-1+k}k)\geq\max_{\jnTL{\ts+k}{k+1}}\QF_{k+1}^{\ttg=i+1,\ts-1}(\oaHistT{\ts-1},{\cat{\jqTTG{k,\ts-1}}{\jnTL{\ts-1+k}k}},\jnTL{\ts+k}{k+1}).\label{eq:lem:QF_k_>_QF_k+1_TTGi:consequence}\end{equation}
\proofup The following derivation \begin{align*}
 & \QF_{k}^{\ttg=i+1,\ts-1}(\oaHistT{\ts-1},\jqTTG{k,\ts-1},\jnTL{\ts-1+k}k)\\
= & \sum_{\joT{\ts}}\Po(\jo^{\ts})\max_{\jnTL{\ts+k}k}\left[\QF_{k}^{\ttg=i,\ts}(\oaHistT{\ts},{\cat{\jqTTG{k,\ts-1}}{\jnTL{\ts-1+k}k}(\joT{\ts})},\jnTL{\ts+k}k)\right]\\
\geq & \sum_{\joT{\ts}}\Po(\jo^{\ts})\max_{\jnTL{\ts+k}k}\left[\max_{\jnTL{\ts+k+1}{k+1}}\QF_{k+1}^{\ttg=i,\ts}(\oaHistT{\ts},{\cat{{\cat{\jqTTG{k,\ts-1}}{\jnTL{\ts-1+k}k}}(\joT{\ts})}{\jnTL{\ts+k}k}},\jnTL{\ts+k+1}{k+1})\right]\\
\geq & \max_{\jnTL{\ts+k}{k+1}}\sum_{\joT{\ts}}\Po(\jo^{\ts})\left[\max_{\jnTL{\ts+k+1}{k+1}}\QF_{k+1}^{\ttg=i,\ts}(\oaHistT{\ts},{\cat{\cat{\jqTTG{k,\ts-1}}{\jnTL{\ts-1+k}k}}{\jnTL{\ts+k}{k+1}}(\joT{\ts})},\jnTL{\ts+k+1}{k+1})\right]\\
= & \max_{\jnTL{\ts+k}{k+1}}\QF_{k+1}^{\ttg=i+1,\ts-1}(\oaHistT{\ts-1},{\cat{\jqTTG{k,\ts-1}}{\jnTL{\ts-1+k}k}},\jnTL{\ts+k}{k+1})\end{align*}
proves the lemma. \end{proof} \end{lemma}

\begin{lemma} \label{lem:QF_k_>_QF_k+1_TTG>i} If, for some stage
$\ts$\begin{equation}
\forall_{\oaHistT{\ts},\jqTTG{k,\ts},\jnTL{\ts+k}k}\quad\QF_{k}^{\ttg=0,\ts}(\oaHistT{\ts},\jqTTG{k,\ts},\jnTL{\ts+k}k)\geq\max_{\jnTL{\ts+k+1}{k+1}}\QF_{k+1}^{\ttg=0,\ts}(\oaHistT{\ts},{\cat{\jqTTG{k,\ts}}{\jnTL{\ts+k}k}},\jnTL{\ts+k+1}{k+1})\label{eq:lem:QF_k_>_QF_k+1_TTG>i:condition}\end{equation}
 holds, then $\forall_{i}\forall_{\oaHistT{\ts-i},\jqTTG{k,\ts-i},\jnTL{\ts-i+k}k}$

\begin{equation}
\QF_{k}^{\ttg=i,\ts-i}(\oaHistT{\ts-i},\jqTTG{k,\ts-i},\jnTL{\ts-i+k}k)\geq\max_{\jnTL{\ts-i+k+1}{k+1}}\QF_{k+1}^{\ttg=i,\ts-i}(\oaHistT{\ts-i},{\cat{\jqTTG{k,\ts-i}}{\jnTL{\ts-i+k}k}},\jnTL{\ts-i+k+1}{k+1}).\label{eq:lem:QF_k_>_QF_k+1_TTG>i:consequence}\end{equation}
\proofup \sloppy  If \eqref{eq:lem:QF_k_>_QF_k+1_TTG>i:condition}
holds for all $\oaHistT{\ts}$, $\jqTTG{k,\ts}$, $\jnTL{\ts+k}k$,
then eq. \eqref{eq:lem:QF_k_>_QF_k+1_TTGi:condition} is satisfied
for all $\oaHistT{\ts}$, $\jqTTG{k,\ts}$, $\jnTL{\ts+k}k$, and
lemma \eqref{lem:QF_k_>_QF_k+1_TTGi} yields $\forall_{\oaHistT{\ts-1},\jqTTG{k,\ts-1},\jnT{\ts-1+k}}$
\begin{equation}
\QF_{k}^{\ttg=1,\ts-1}(\oaHistT{\ts-1},\jqTTG{k,\ts-1},\jnTL{\ts-1+k}k)\geq\max_{\jnTL{\ts+k}{k+1}}\QF_{k+1}^{\ttg=1,\ts-1}(\oaHistT{\ts-1},\cat{\jqTTG{k,\ts-1}}{\jnT{\ts+k-1}},\jnTL{\ts+k}{k+1}).\label{eq:lem:QF_k_>_QF_k+1_TTG>i:eq1}\end{equation}
At this point we can apply the lemma again, etc. The $i$-th application
of the lemma yields \eqref{eq:lem:QF_k_>_QF_k+1_TTG>i:consequence}.
\end{proof} \end{lemma}

\newpage

\bibliographystyle{theapa}
\bibliography{jp_QV4Dec-POMDPs}

\end{document}